\documentclass[pdflatex,sn-mathphys-num]{sn-jnl}

\usepackage{graphicx}%
\usepackage{multirow}%
\usepackage{amsmath,amssymb,amsfonts}%
\usepackage{amsthm}%
\usepackage{mathrsfs}%
\usepackage[title]{appendix}%
\usepackage{xcolor}%
\usepackage{textcomp}%
\usepackage{manyfoot}%
\usepackage{booktabs}%
\usepackage{algorithm}%
\usepackage{algorithmicx}%
\usepackage{algpseudocode}%
\usepackage{listings}%

\theoremstyle{thmstyleone}%
\theoremstyle{thmstyletwo}%

\theoremstyle{thmstylethree}%

\begin{document}

\title[Contextual trajectory and incremental contextual displacement: Towards using LLMs to understand dynamic, utterance-specific meaning construction]{Contextual trajectory and incremental contextual displacement: Towards using LLMs to understand dynamic, utterance-specific meaning construction}


\author[1,2]{\fnm{Grayson Wycliffe} \sur{Storer}}\email{grayson.storer@uvm.edu}

\author*[3,4]{\fnm{Julia Witte} \sur{Zimmerman}}\email{julia.zimmerman@uvm.edu}

\affil*[1]{\orgdiv{Department of Mathematics and Statistics}}

\affil[2]{\orgdiv{Department of Computer Science}, \orgname{University of Vermont}, \orgaddress{\city{Burlington}, \postcode{05405}, \state{Vermont}, \country{United States}}}

\affil[3]{\orgdiv{Computational Story Lab}}

\affil[4]{\orgdiv{Computational Ethics Lab}, \orgname{Vermont Complex Systems Institute}, \orgaddress{\city{Burlington}, \postcode{05405}, \state{Vermont}, \country{United States}}}


\abstract{
Transformer-based large language models (LLMs) such as RoBERTa represent text using contextual word embeddings (CWEs), which alter the embeddings associated with each token based on surrounding context. We construct token-wise incremental trajectories by repeatedly recomputing a token's CWE as successive words are added to a sentence, yielding a representation of how contextualized embeddings evolve as the utterance unfolds. We evaluate this approach using garden-path sentences as a test case with characteristic features. Token-wise trajectories reproduce known features of garden-path processing, including disruption around the critical region, and reliably distinguish garden-path sentences from matched disambiguated controls. We introduce several metrics for quantifying representational displacement across contextual increments and show that trajectory information can be highly predictive of sentence type. We find that ambiguity-related information is recoverable not only from the sentence-level CLS representation but also from ordinary vocabulary tokens, suggesting that utterance-level information is distributed across multiple representational scales. In exploratory analyses, we find qualitatively similar trajectory structures in other ambiguity- and misdirection-related linguistic phenomena. Together, these results establish token-wise incremental trajectories as a promising framework for studying utterance-specific meaning construction using LLMs.
}



\maketitle
\begin{figure}[h]
    \centering
    \includegraphics[width=0.6\linewidth]{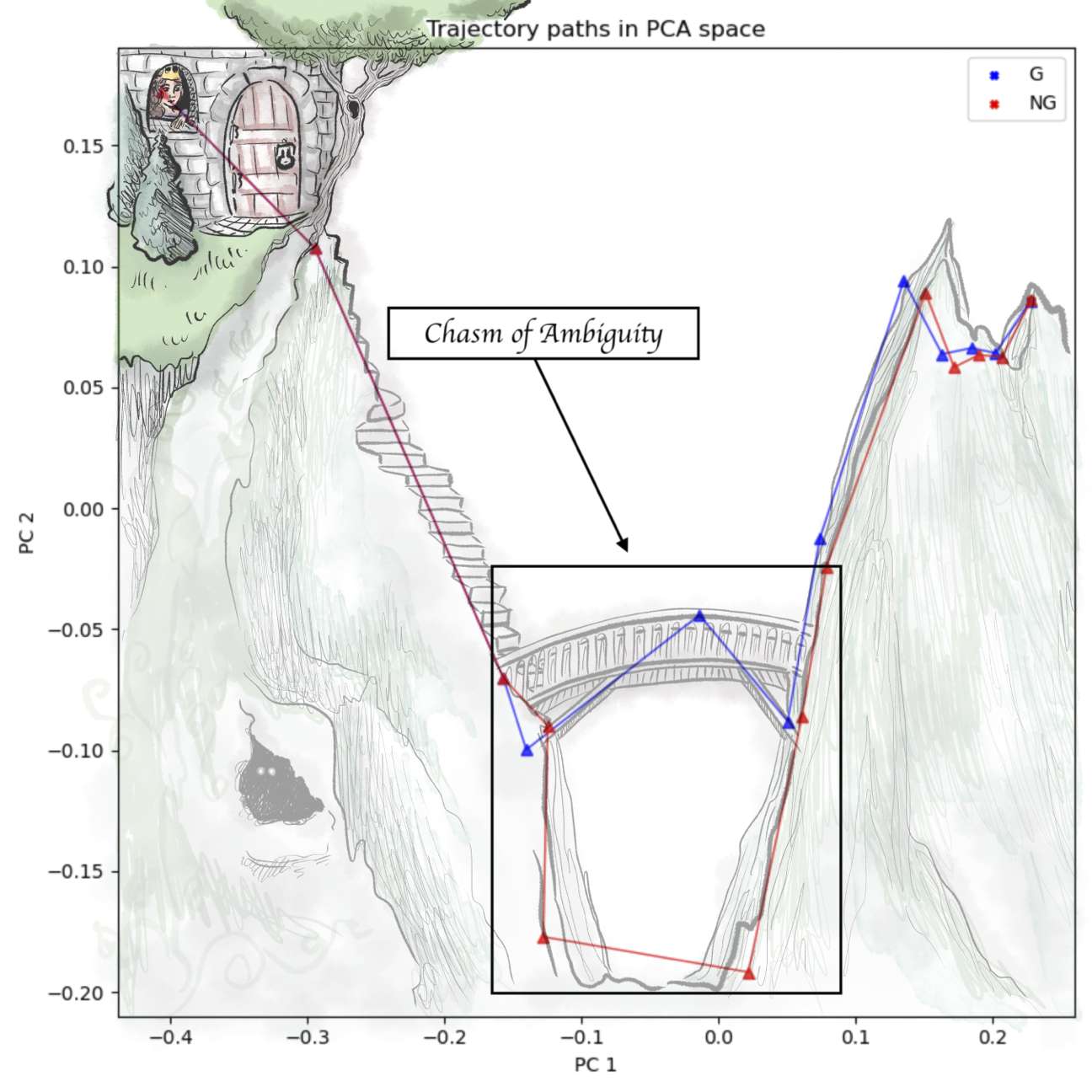}
    \caption{An illustration of characteristic semantic gap between garden path and equivalent disambiguated sentences.}
    \label{fig:chastleambiguity}
\end{figure}
\section{Introduction} 
\label{sec:introduction} 

This project is intended to be a small step in a much more ambitious direction. The big picture motivation -- explained in Sec.~\ref{sec:motivation} -- is that we want to study local meaning construction, and we think the utterance-specific, dynamic representations LLMs rely on hold significant promise for that. In this work, we explore token-wise contextual trajectories as one such candidate representation. To validate that representation is meaningful -- as in, correlated with the process of meaning construction in the LLM in some interpretable way -- we test it on garden path sentences (GPs). Prior work on GPs (in LLMs and in people) provides us with some baseline expectations to draw from.

The primary result of this paper is that token-wise contextual trajectories do appear to capture aspects of meaning construction as the utterance unfolds. We find that this representation reproduces known garden path phenomena. Furthermore, the representation contains enough information for a classifier to reliably distinguish garden path from non-garden-path sentences. In exploratory work, we see that similar trajectory structures appear in other ambiguity/misdirection phenomena, which is consistent with our intuition that LLM representations may be able to capture characteristics of multiscale linguistic phenomenon. The incremental contextual trajectories may therefore be useful more broadly for studying utterance-level meaning construction.

\section{Motivation}
\label{sec:motivation}

For its entire evolutionary history~\cite{fedorenko2024primarily}, only humans\footnote{Or at least, hominins.} produced human-like language. Remarkably, LLMs now represent the first extra-human demonstration of something that convincingly looks like this ability. Deep learning models may represent a minimally viable language technology that can be disentangled from many hallmark properties of human cognition~\cite{mahowald2024dissociatinglanguagethoughtlarge,zimmerman2025locality}. Although there are significant unknowns and differences, they inarguably offer unique opportunities to deepen our understanding of individual and collective cognition and problem-solving, and of language as an abstract system. These models appeal across the many disciplines interested in bowtie models of cognition~\cite{levin2024stigmergy}. As the most performant computational language models to date, the internal states of LLMs promise richer quantitative representations of linguistic features than ever before. Their potential for studying fine-grained, utterance\footnote{In this paper we use \textit{utterance} to mean an instance of language (speech or writing) that has some meaning and is evaluated as a whole, but is not necessarily a complete sentence or set of sentences.}-specific, dynamic meaning is largely still developing.

We want to help develop methodologies for using LLMs to study utterance-specific meaning construction rather than diachronic semantic change. The latter has traditionally been more prevalent in computational linguistics~\cite{hamilton-etal-2016-diachronic,Card_2023,Liu_2021,periti2025semantic,kishino2025quantifyinglexicalsemanticshift}, in large part because earlier computational tools were better suited to coarse-grained, corpus-level analyses. By aggregating statistical regularities from large text corpora and representing them via meaning-mediated abstractions~\cite{tinyblitz} in lower-dimensional latent spaces,\footnote{The knot in the bowtie model.} LLMs provide quantitative representations that can be used to investigate fine-grained aspects of meaning construction. This provides a way to study the role of context in meaning construction, and the relationship between these phenomena and distributional accounts of meaning~\cite{harris1954structure,sahlgren2008distributional,mcdonald2001testing} and language acquisition~\cite{chiang-yogatama-2023-distributional}.

The approach introduced here offers a step towards a quantitative framework for examining relationships among tokens, words, and contexts using LLM representations. Although our focus is on meaning construction, related methods could potentially be applied to other domains, including machine translation.\footnote{As research explores correspondences between human cognition and machine learning~\cite{binz2025foundation}, opportunities for applying LLM-based methods in linguistics and cognitive science may continue to expand. However, our focus here relies on linguistic behavior rather than underlying cognitive alignment (or lack thereof).}

\begin{center}\textbf{Key takeaways:}\end{center} 
\begin{itemize} 
    \item Garden-path incremental trajectories reproduce known garden-path processing behavior, showing disruption near critical regions and subsequent recovery. This supports token-wise incremental trajectories as a plausible framework for studying utterance-specific meaning construction. 
    \item This signal survives even in the first vocabulary token, despite its not being a dedicated sentence-level representation like CLS. This suggests distributed, multi-scale representations of linguistic information: individual token representations appear to reflect both local lexical information and broader utterance-level structure, implying that sentence-level information is neither purely localized nor confined to dedicated sentence-level representations. 
    \item Incremental trajectories almost perfectly separate garden-path and control sentences. A classifier trained on unshuffled trajectories achieves high accuracy, while performance collapses when trajectory order is randomized, indicating that meaningful information is encoded in the relationships of the trajectory itself rather than solely in static representational properties. 
    \item Similar trajectory signatures appear across ambiguity-related phenomena at multiple linguistic scales. Although these analyses are exploratory, they suggest that trajectory-based methods may generalize beyond garden-path sentences for studying utterance-specific meaning construction. 
\end{itemize}

\section{Literature Review} 
\label{sec:Lit_review}

When introduced, static word embeddings (SWEs)\footnote{For example, GloVE and Word2Vec~\cite{pennington2014glove,church2017word2vec}.} enabled significant advances in natural language processing (NLP), and they continue to be suitable representations for many tasks, such as distant reading methods over large corpora and coarse-grained diachronic language change. 

However, SWEs are limited when it comes to fine-grained contextual differences in meaning, a prominent example being their difficulty representing polysemous words (e.g. ``run", ``bank")~\cite{miaschi_contextual_2020}. Various extensions of the SWE approach have been used to mitigate these shortcomings, especially with respect to polysemy, including but not limited to finetuning pretrained embeddings~\cite{Taghipour2015SemiSupervisedWS, 10.1145/3543507.3583191}, but none of these approaches have been as powerful and general-purpose as contextualized word embeddings (CWEs). 
Transformer architectures and the development of contextual word embeddings leading to the release of bidirectional transformers such as BERT greatly improved the semantic interpretation of polysemous words in computational models\cite{devlin_bert_2019,peters_dissecting_2018, vaswani2023attention}, though at the cost of additional computation and increasingly complex latent space geometry \cite{cai_isotropy_2021, peters_dissecting_2018, mikolov2013efficientestimationwordrepresentations, mickus_isotropy_2024}. 

The concomitant increase in performance of computational models of language offers incredible promise for computational linguistics research, although the complexity of latent space geometry complicates the application of these tools and adds to computational complexity. Specifically, although CWEs are much better at handling linguistic ambiguity within a single surface form, they are not infallible.\footnote{Various methods for word sense disambiguation (WSD) using CWEs such as pretraining embeddings and algorithms for learning decision boundaries within the embedding space \cite{yuan-etal-2016-semi,vial2018improvingcoveragegeneralizationability}.} However, their reliance on the frequency of surface forms in specific contexts within training data, draws helpful though limited parallels to how people acquire language. Importantly, when given incrementally longer contexts, the displacement across subsequent embeddings may align with models of how humans process sentences. \cite{tikochinski_incremental_2025} \citet{hadiwinoto-etal-2019-improved} explored pre-training CWEs such that senses of polysemous tokens are more easily and accurately determined. \citet{wiedemann2019doesbertmakesense} studied how a simple nearest neighbors method can be used for word sense disambiguation in the contextual embedding space of the BERT model, demonstrating a simple-yet-effective approach to WSD and arguing that further improvements would require a more complex method due to the non-linearity of the decision boundaries in BERT's contextual embedding space. 
However, much remains unknown. Ambiguity is an important feature of language (particularly for optimizing efficiency as described in ~\citet{zipf1965}); thus it is important to make sure that advances in computational models of language are consistent across ambiguous linguistic phenomena. 

Garden path sentences are one such ambiguous phenomenon. Garden path sentences are relatively strictly defined as they require a likely misinterpretation after the first parse, followed by reparsing (and sometimes also characterized as inducing lingering echoes of the original misparse). This effect is driven by relatively higher prior likelihood of co-occurence between linguistic forms used in the Garden Path sentence in an atypical way. \citet{futrell2019neurallanguagemodelspsycholinguistic} proposed the idea of measuring classically psycholinguistic phenomena, including the misinterpretation of garden path variety sentences, by examining surprisal at each increment using a combination of two datasets from \citet{STURT1999136,PMID:12690829}, which contain garden path sentences and their unambiguous equivalents (also used here and in \citet{wang_causality_2024}). 

\citet{futrell2019neurallanguagemodelspsycholinguistic, wang_causality_2024, li_incremental_2024} used incremental internal states such as surprisal metrics and next token logits. \citet{li_incremental_2024} also used the models' self-reported ``comprehension'' as determined by answering yes/no questions about the sentences given to the model.

\citet{jurayj_garden_2022} used CWEs for garden-path variety sentences by analyzing Manhattan distance and cosine similarity of the CWEs of tokens in garden-path sentences compared to disambiguated versions of sentences with similar meanings.

Our methodology builds on the implementations of contextual increments as used by \citet{futrell2019neurallanguagemodelspsycholinguistic,wang_causality_2024, li_incremental_2024}, while incorporating CWEs as in \citet{jurayj_garden_2022}, although we examine the displacement between each embedding rather than the final position of each embedding (since we are interested specifically in developing an analog of how meaning is constructed as the model's representation of the utterance evolves with context). Our approach is to represent an utterance as the sequence of changes undergone by a token embedding as additional context is added.

We extend the contextual trajectory work of \citet{bigelow2026storiesspaceincontextlearning}. Where that work focused on global linguistic phenomena over relatively large increments such as character or emotional arcs in stories as observed through sentence-length increments, we apply similar techniques at the token-scale. Whereas \citet{bigelow2026storiesspaceincontextlearning} investigates global behavior (e.g. character arcs), we focus on utterance-level meaning construction. We choose garden path sentences as our subject because they exhibit local, token-scale ambiguity; existing work on garden path sentences should allow us to start making connections between our ideas about meaning construction, what is known already, and the internal states of LLMs~\cite{zimmerman2025locality,zimmerman2025tokensoftoverlookedappetizerlarge,tinyblitz}. In future work, we want to further explore such connections, working towards a greater understanding of meaning construction that unites micro (e.g. morphemes) and macro (e.g. narrative) scale structures.

\section{Methods} 
\label{sec:Methods}

We primarily use the RoBERTa base model~\cite{liu_roberta_2019}. Because here we are interested in probing meaning construction of received utterances, rather than producing utterances, we chose a model that focuses on the encoding half of the bowtie model of cognition~\cite{levin2024stigmergy} and which comes from a family that has been widely used in LLM research~\cite{devlin_bert_2019,liu_roberta_2019}.

Our unit of analysis is the CWE for the tokens in the utterance as its context grows. We use individual token vectors as well as the embedding of the CLS token. In RoBERTa, the CLS token is not trained with its own objective, which often takes place during fine-tuning before classification tasks. However, we are only interested in the CLS token as a way of representing an accumulation of token-level context, which it does, as it attends to all the other (pretrained) tokens in the context window, so we did not fine-tune~\cite{liu_roberta_2019}. In fact, we are more interested in what the model learns to store in CLS during base training, and fine-tuning would likely have obscured that.

In Appendix~\ref{sec:rq3}, we also use DeBERTa and DistilBERT, which, like RoBERTa, are variants of the BERT model, in order to establish that it is plausible to think that the incremental contextual trajectories of garden path sentences could be fairly architecture independent~\cite{devlin_bert_2019,liu_roberta_2019,sanh2020distilbertdistilledversionbert,he2021debertadecodingenhancedbertdisentangled}. In future work, this will need to be tested with greater variety of models and sentences, but we wanted to establish antecedent plausibility of that hypothesis. We used DeBERTa because it uses separate embedding vectors to represent position and semantic content, which might help alleviate the caveat addressed in Appendix~\ref{appendix:caveat_context_change_interpretation}, and could significantly alter the incremental trajectories. We used DistilBERT because it is a much smaller model, which also may significantly alter incremental trajectories. Therefore these were reasonable variations to try out here.

\subsection{Operationalization}

We introduce an implementation of incremental context similar to those used in \citet{li_incremental_2024,jurayj_garden_2022}. We use the same incremental scale where each increment is a new embedding in the latent space; however, we also examine the displacement between each increment in addition to the final position in latent space. 

\begin{figure}[h]
    \centering
    \includegraphics[width=0.6\linewidth]{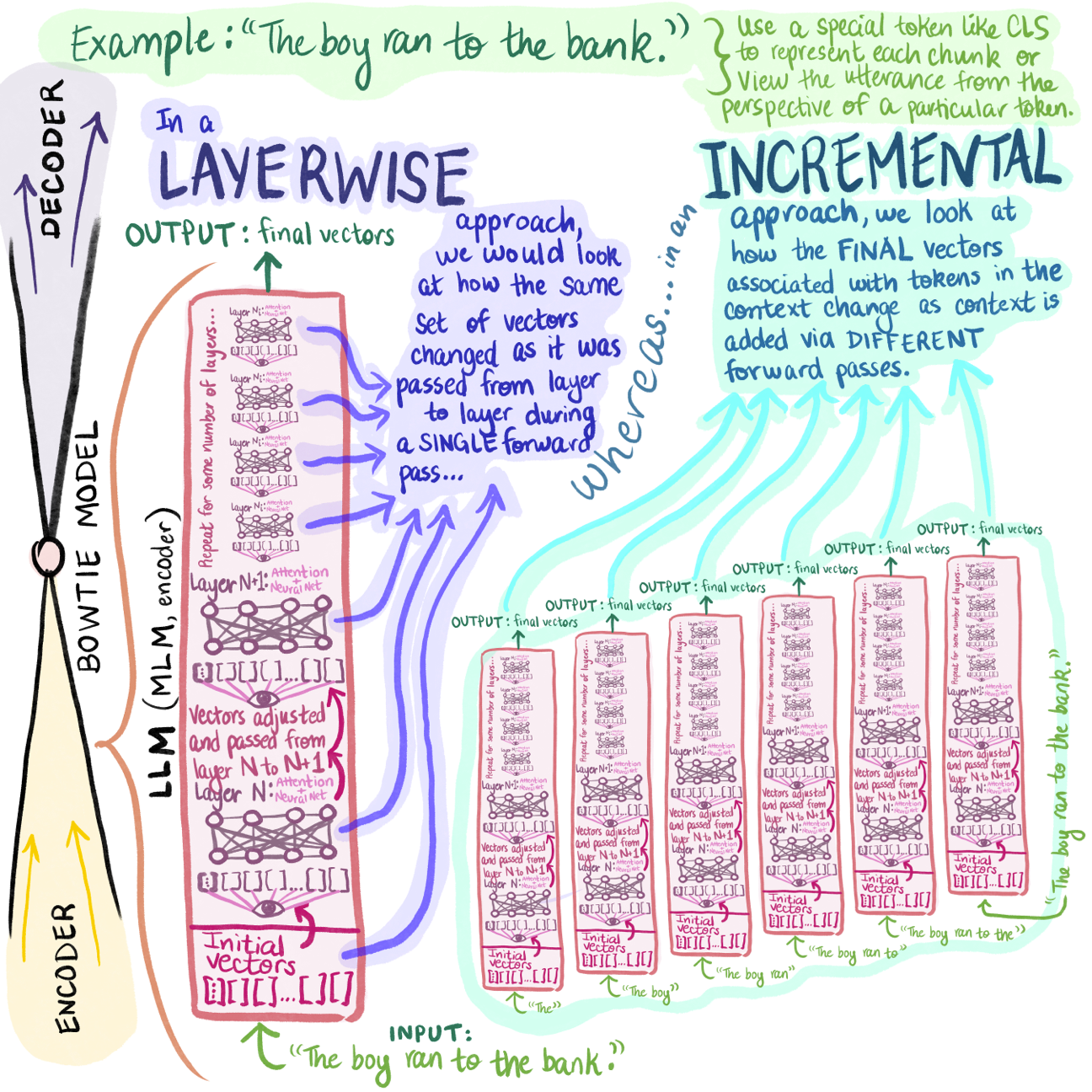}
    \caption{An illustration of incremental context, as opposed to layerwise LLM analysis.}
    \label{fig:incrementalvslayerwise}
\end{figure}

Incremental context trajectories aggregate evolving vector positions across independent forward passes as a sequence (see Figure~\ref{fig:incrementalvslayerwise}. Although computationally expensive (requiring $N$ forward passes for every $N$-length utterance), they may be a good way to leverage the significant commonalities in linguistic performance across some architectural differences, such as number of layers, resulting in a partial picture of meaning construction that is easily comparable across many models. This seems especially promising in conjunction with focusing on different tokens as a way to shift the perspective on what aspect of meaning is being constructed, something which is usually done via layerwise analysis.

\subsection{Moving from coarse-grained to fine-grained meaning construction: caveat}
If we want to use fine-grained context evolution to examine meaning construction, as one token after another is added, we then have to contend with the impact of individual token choice. \footnote{When casually discussing human language, it is easy to make statements like, ``these two sentences mean the same thing but say it in different ways'', accepting some stability of some aspects of meaning, such as semantic or propositional meaning at a broad level, but acknowledging variations in other aspects of meaning, such as pragmatic meaning (implicature, tone, etc.). Moreover, on closer examination, these glosses likely become much hairier even in the most familiar and comfortable cases. However, LLMs do not have human architecture, motivations, or experiences, so we have to tread extra carefully.} No two distinct contexts have identical meanings from the perspective of the LLM (since, for example, every token in the context will influence every other token via typical attention implementations). And we can't assume that those differences in representation will align with how we might describe our own experience (``there is a typo, so I know this really means the same thing''; ``there is a different choice of tone, so I know the speaker had a different intent''; ``here the number nine is represented as a string, here as a digit, but clearly these statements have identical meanings''~\cite{marjieh2025numberlargelanguagemodel}), especially since our goals in learning and using language are undeniably significantly different than the goals that guide LLMs as they learn and use language.\footnote{Architectural choices are epistemic choices. Consider a plausible token objective: this goal collapses aspects across many sociolinguistic strata into a single point. For more discussion of this topic, see \citet{zimmerman2025tokensoftoverlookedappetizerlarge}, \citet{zimmerman2025locality}, and works cited therein, especially \citet{mahowald2024dissociatinglanguagethoughtlarge} and \citet{mccoy2023embersautoregressionunderstandinglarge}.}

To that end, although much is yet unknown, we draw on existing literature and our own experiments and conclude we can most safely proceed by trying to stick with utterances that conform to the most likely mold, keep in mind position, length, and potential tokenizations, and deal with relative, rather than absolute, vectors.\footnote{Various papers have examined the ability of LLMs to parse sentences with slight perturbations in formatting, exploring aspects of a fuzzy spectrum where some aspects of meaning are stable to some perturbations, but some display what might be called sensitive dependence on initial conditions. For example, \citet{freeburg2026fingerprintmarkdowntrainingshapes} approaches this problem by examining how prompt engineering and tweaks to training data can retain model performance while avoiding classical markers of LLM generation, particularly the em dash. Broader perturbations to sentence formatting varying the specificity of prompts are studied in \citet{he2024doespromptformattingimpact}, who find that LLM performance is improved by eliminating sources of ambiguity in prompts. \citet{11413757} test how the insertion of atypical characters into prompts (specifically emojis) affects model performance and find that strategic placement of emojis can augment the performance of punctuation by emphasizing different subsections of prompts. See also prior section and Appendix~\ref{appendix:caveat_context_change_interpretation}.}

Word level tokenization allows us to take advantage of CWEs updating for each utterance in order to more accurately reflect the process of parsing an individual word in an utterance. We use this tokenization to study if changes in latent space after the introduction of an ambiguous word aligns with other models of human cognition of ambiguous phenomena. This method of studying the fine-grained meaning construction of a word is fundamentally exclusive to CWEs rather than SWEs, as we rely on studying the change of individual tokens in the LLMs vocabulary, something that does not occur within SWEs. 

\subsection{Definition of incremental context}

We represent the trajectory of an utterance using a function $D: T \rightarrow \mathbb{R}^{768} \text{ s.t. } \forall t_n \in T, D(t_n) = t_n - t_{n-1}$, where $T$ is the set of all tokens in the phrase. This definition of $D$ was selected by testing four different function operationalizations against a sentence type (interrogative, declarative, exclamatory, imperative) classification task. This process is explained in more detail in Appendix~\ref{appendix:defining_contextual_trajectory}. 

This representation of context probes how the latent space evolves as words (specifically, tokens)\footnote{For all tasks in the paper, we checked that all tokenizations would be word-level} are added to the context window. We explore the perspective of the special CLS token as well as the perspective of specific vocabulary tokens within the utterances. Illustrated in Figure~\ref{fig:incrementalvslayerwise}: Unlike studying the layer-wise evolution of CWEs, this representation of contextual increments only examines the position of embeddings after a complete forward pass, such that this representation may be more consistent across various model architectures (leveraging commonalities across linguistic competence in related models, and not requiring precise mappings between different amounts of layers or other specific architectural components). 

\subsection{Research Questions}

\textbf{Case study: Using garden path sentences to establish consistency between our operationalization of incremental trajectories and existing psycholinguistic models}

If we can draw parallels between the incremental trajectories of garden path sentences and aspects of garden path sentences put forward in previous works (with computational models and with humans), that provides face validity to incremental trajectories as a model of meaning construction in at least one linguistic phenomena, and leaves open the potential for more general utility. On the other hand, if we do not see parallels between our own subjective experience of GP sentences or any of the models, observations, or results put forward in prior work, then that suggests (though does not prove) we may need to reconsider whether representing meaning construction in this way is an effective approach for meaningful alignment between human cognition and computational linguistics.

\citet{jurayj_garden_2022} results show that the embeddings of ambiguous tokens in garden-path variety sentences show a non-trivial spike in Manhattan distance when compared to the targeted meaning in the disambiguated version of the sentence. This result offers insight as to how the internal states of models process ambiguous tokens, the authors do not investigate whether this behavior is characteristic of garden-path variety sentences. We expect that the semantic disparity observed in \citet{jurayj_garden_2022} will also be reflected by our method of incremental contextual trajectory, such that the vectors representing displacement of tokens will reflect greater displacement when the next token is located within the critical region. 

We use the method of incremental context as it better aligns with how humans perceive spoken linguistic structure.\footnote{In the sense that when you hear an utterance, the sounds are produced and perceived in linear, temporal order.} We know from the work by \citet{tikochinski_incremental_2025} that the way LLMs process text using long context windows and parallel processing does not align with human comprehension in at least some notable respects and instead a closer analog would be processing smaller chunks of text incrementally. To that end, we pursue the following research questions, looking for apparent and functional connections between what is already known about GPs and features of the incremental context trajectories.

\subsubsection{RQA1}
\label{sec:rqa1}
What do the token-wise incremental trajectories of garden path sentences look like? Specifically, based on what we know about the processing of garden path sentences as shown in  \citet{jurayj_garden_2022} and \citet{futrell2019neurallanguagemodelspsycholinguistic}, do we see a characteristically jagged trajectory in the critical region of garden path sentences corresponding to the dips in semantic similarity observed in garden path traversal experiments from \citet{jurayj_garden_2022}?
We examine the trajectories by computing the embeddings of the contextual increments and calculating the distance vectors between each subsequent embedding. We use principle component analysis decomposition in order to project the vectors onto a plane in $\mathbb{R}^2$ can be used to plot the trajectory of sentences through the PCA decomposition of contextual embedding space. We plot garden path sentences and equivalent unambiguous sentences on the same plane to see if and how their trajectories are different. 

\subsubsection{Results for RQA1}
\label{sec:gptraj}

We introduce the following formalization: For two sentences, one of which is a garden path and the other is the disambiguated equivalent, with a critical region occurring between indices $i$ and $j$ where $i < j$. The incremental trajectories will be equivalent for all increments until the index $i$ (cosine similarity of 1), the minimum cosine similarity (maximum semantic distance) will occur within increments $i$ and $j$ (the critical region), then all all remaining increments will be within $\epsilon$ of each other where $\epsilon <$ max distance $\in [i, j]$. 

\begin{figure}[h]
    \centering
    \includegraphics[width=0.4\linewidth]{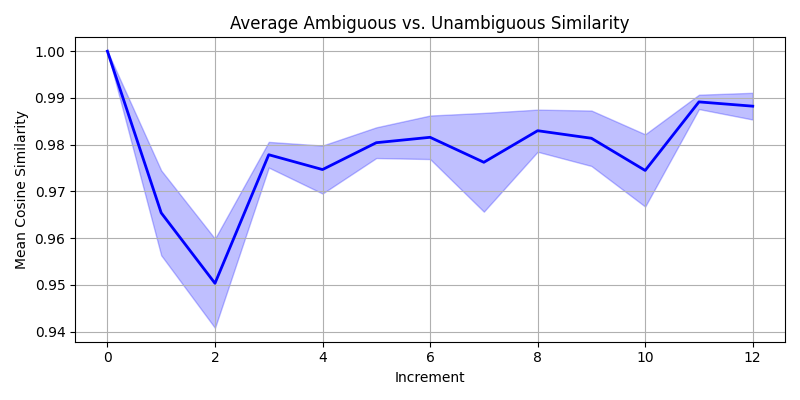}
    \includegraphics[width=0.4\linewidth]{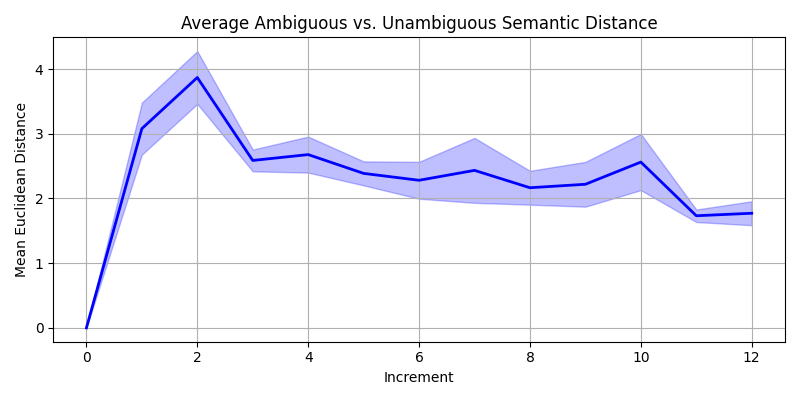}
    \caption{Cosine similarity and Euclidean distance between garden path and disambiguated sentences, all having a critical region occurring at the same point, specifically the 3rd word in the sentence. 
    We see the same behavior that we describe in our formalization: the two trajectories are identical at the beginning of the sentence, the largest separation between the trajectories occurs in the critical region (2nd index, 3rd word), and the semantic similarity increases through the last increment.}
    \label{fig:formalization}
\end{figure}

From our projection into $\mathbb{R}^2$ (shown in \ref{fig:chasm_with_no_art}), we observe the characteristic gap between the trajectory of the garden path sentence and the trajectory of the equivalent unambiguous sentence, at a spot corresponding to the ambiguous token. We expect that the latent space representation of the two different sentences may correspond with model surprisal and model misinterpretation after processing an ambiguous token, which can be verified in future work. Furthermore, the difference in trajectories around the ambiguous region may align with what we know about increases in human reading times of garden path sentences as described in \citet{mainakilaas2026revisitingrealtimediggingineffects}, though we leave further analysis of this relationship for future work. 

\subsubsection{RQA2}
Assuming we see, qualitatively, what looks like a characteristic GP trajectory (which we informally call a ``U-turn''), how do we know it is truly meaningful, and even definitive of garden path sentences in this representation? We test this by training a classification deep learning model which receives the trajectory of one sentence as input and determines whether the sentence is a garden path sentence or not, based on the token-wise incremental trajectories. We hypothesize that the ``U-turn'' component of the trajectories will be a signature of garden path trajectories and therefore a large portion of the signal that the classifier uses. 

To test our hypothesis that the ``U-turn'' component of the trajectories makes up the majority of the signal, we shuffle the increments of the trajectory to verify if both the global and local structure of the representation are interpretable. We will shuffle by increment size, down to 1 token. The ambiguity that we hypothesized exists in garden path sentences exists for at least 3 tokens in most sentences in our dataset, so shuffling by 1 token chunk effectively negates the presence of a characteristic interval.  

\subsubsection{Results for RQA2}

The model we constructed performs a 1 dimensional convolution over the contextual trajectory of the sentence. We trained the model on the Grodner dataset of garden path sentences and disambiguated equivalent sentences using an $80\%-20\%$ train-test split. Over 100 randomly instantiated models trained for 25 epochs each, the average accuracy on the validation dataset was $\sim98.5 \%$. We verified that the model is forming a legitimate classification boundary by using the Sturt dataset as a holdout test set. The average accuracy on the holdout test set was $ \sim 99.8 \%$.

Although these GPs do not represent all GPs, for this subset, we find that trajectory information almost perfectly separates garden path sentences and their disambiguated control sentences. Therefore, some aspect of the trajectory distinguishes between these two cases; we suspect the main feature the classifier relies on could be the jaggedness we observed to be characteristic of GP trajectories. We hone in on that explanation by disrupting order in RQA3 (Sec.~\ref{sec:rqa3}).

The training and validation curves shown in \ref{fig:unshuffled} show slight overfitting to the training data, which is expected. Furthermore there is no observable collapse in the models ability to generalize to the validation dataset at any point in the training process. The results of the model on the holdout dataset as well as the consistent performance across validation datasets indicate that the model is forming a meaningful decision boundary and is able to generalize to data outside of the training distribution.

We perform a statistical reconstruction of the average trajectory of garden path sentences and render the trajectory in PCA space. The average trajectory shown in \ref{fig:average_garden_path_traj} illustrates the characteristically jagged trajectory through PCA space, which we suspect is the main feature the classifier is relying on. 

\subsubsection{RQA3}
\label{sec:rqa3}

We want to increase the likelihood that our hope that our model is forming the classification decision boundary as a result of trajectory (an inherently order sensitive phenomenon) is correct. We retrained the model with each trajectory randomly shuffled, such that there would be no characteristic jagged interval within the trajectory. Randomly shuffling the trajectories creates input data that does not contain characteristic intervals; in the case of the sentences shown in \ref{fig:formalization}, this would require 3 or more contextual increments within the trajectory to maintain their order after a random permutation. We hypothesize that if the classifier is forming a decision boundary on something other than the trajectory (for example, the common presence of the word ``that'' in the disambiguated controls), then this decision boundary will be order invariant -- it would survive this disruption. In that case, the model trained on shuffled data should exhibit similar performance metrics to the model trained on unperturbed, order-preserved data.

\subsubsection{Results for RQA3}
\label{sec:resultsforrqa3}

After shuffling the trajectories of each sentence in the Sturt and Grodner datasets, we trained the same model architecture from RQ2 using the same training parameters. On the shuffled trajectories, the model quickly overfit to the training data and was unable to generalize to the validation dataset. Average accuracy on the validation dataset after 25 epochs on 100 randomly instantiated models was $\sim 56 \%$, indicating that the trained models were unable to perform much better than random guessing. This performance is shown in \ref{fig:shuffled}. 

The poor performance of the model on the shuffled trajectories shows that the model trained on un-shuffled data was forming a decision boundary based on order-sensitive characteristics within the trajectory. Therefore, we can much more confidently say that the classifier model is fitting to the jaggedness of the characteristic interval, although we do not definitively show that. In conjunction with the high performance of the classifier model, these results align with our hypothesis that contextual trajectories of garden path sentences both have characteristic intervals and that those characteristic intervals are sufficient to distinguish them from disambiguated controls.

\subsubsection{RQA4}
\label{sec:rqa4}

Our results in RQA1, RQA2, and RQA3 with the CLS token establish that the incremental trajectory representation does share characteristics established in prior GP research. In RQA4, we test whether these characteristics persist in representations at multiple resolutions within the model, by examining the trajectories as seen from the perspective of specific vocabulary tokens: the initial token, the first token located at the beginning of the critical region as reflected in the ``U-turn'', and the last interval (between the penultimate and final tokens). Due to the implementation of the CLS token within RoBERTa, we expect to see broadly similar behavior from the trajectory of the first token in the utterance. As we move further into the sentence, information locality constraints will limit the amount of signal in the trajectory of single word tokens. 


\subsubsection{Results for RQA4}
\label{sec:resultsforrqa4}

To test the extent to which individual tokens reflect an ambiguity at a greater scale (multiple tokens; clauses or sentences), we retrained the classifier using the trajectory of the first word of each sentence in the dataset as the input data.\footnote{We used the first word because it has the most opportunity to change as the context evolves, without a priori assuming whether the representation of a specific word will change meaningfully or not. All tokens will change, so we don't want to assume change is meaningful in the way we care about.} We found that when the classifier was trained on these new data, it was able to reach similarly high performance metrics: $\sim97.8 \%$ on the validation dataset and $\sim99.7 \%$ accuracy on the holdout dataset. These results indicate that the latent space represented via individual vocabulary tokens can reveal larger scale information about the nature of the utterance in which they exist. Training and validation accuracy curves of this model are shown in \ref{fig:first_token}. 

When we look at the embedding of the second to last token, we have a single vector $\vec{v} \in \mathbb{R}^{768}$ such that $\vec{v} = \phi(t_{n-1}, n-1) - \phi(t_{n-1}, n)$ where $\phi$ is RoBERTa's embedding function, $t_{n-1}$ is the penultimate word in an $n$-length utterance, $n-1$ is the size of the context window of the first embedding, and $n$ is the size of the context window of the second embedding. This calculation yields a single vector instead of a time-series of vectors. For this reason we use a fully-connected feed-forward neural network for the classification task instead of a convolutional neural network. The best accuracy that the model achieved was $\sim59 \%$ on the validation dataset and roughly $\sim50.9 \%$ on the holdout dataset. Training and validation accuracy curves of this model are shown in \ref{fig:last_token}.

Furthermore, we examine the trajectory of the first token in the critical region of the sentences, which is often located in the middle third of the sentence. When we retrained the classification model on the trajectories of the first token in the critical region, we were able to achieve a validation accuracy of $\sim94.8 \%$ but our accuracy on the holdout dataset was only $\sim54.9\%$. We suspect that the reason for the decreased accuracy on the holdout dataset is due to the significant difference in sequence length (sentences in the Sturt dataset are significantly longer than sentences in the Grodner dataset). When we switch to using the trajectory of a token that is not present for the entire sentence, we are creating a more difficult classification task, which is exacerbated by the significant difference in sequence length between the holdout and validation datasets. Nonetheless, the high validation accuracy is indicative that there is a significant amount of signal present in the trajectory of the first token in the critical region, but the shorter window of representation prevents the classifier from being able to generalize on this signal. Training and validation accuracy curves are shown in \ref{fig:divergent_hinge}.

\begin{center}
\textbf{Exploratory connections}
\end{center}
\subsection{Establishing plausible correspondence across qualitatively different linguistic scales}

Garden path sentences involve local ambiguity and reinterpretation. We next ask whether similar trajectory signatures appear in phenomena involving broader forms of reinterpretation or misdirection, which would support the hypothesis that incremental trajectories may capture scale invariant properties of linguistic phenomena such as ambiguity (and, more broadly, that internal LLM representations can preserve connections across distinct phenomenon that are qualitatively related for people).

\subsubsection{RQB1}
In particular, we are interested in whether this representation illustrates the presence of global ambiguity as well as the specific GP variety. Garden path sentences exhibit local ambiguity (reparsing a certain word or clause), whereas jokes often exhibit broader global ambiguity \cite{Lew2010AmbiguitygeneratingDI, Oaks1994CreatingSA}.
For example, both garden path sentences and certain kinds of jokes rely on a qualitatively similar re-analysis of a key word of phrase, a point of surprise or realization in which the initial prediction is contraindicated. We hypothesize that the characteristic interval (``U-turn") that (we hypothesize) shows up for garden path sentences will also surface in certain categories of joke. 
We test this hypothesis by calculating the average trajectory of ``dad jokes" from the dataset compiled by Silvia and Burnett \cite{silvia_burnett_2025} through embedding space to see if there is a characteristic trajectory that these jokes have in common. 

\subsubsection{Results for RQB1}
\label{sec:JokesRQ}
To examine how incremental trajectory represents other linguistic phenomena that rely on misdirection of some kind, we performed the same incremental trajectory analysis from RQ1 on dataset of ``dad jokes" from taken Reddit and compiled by Silvia and Burnett \cite{silvia_burnett_2025}. We observe consistent global ``U-turn" phenomena that is complexity invariant. 

\begin{figure}[h]
    \centering
    \includegraphics[width=0.3\linewidth]{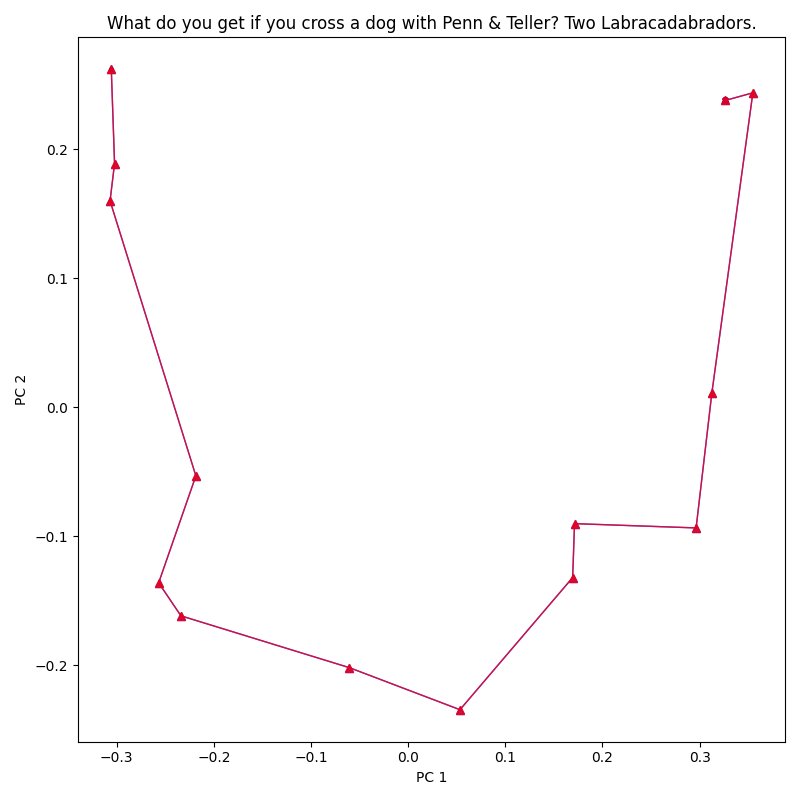}
    \includegraphics[width=0.3\linewidth]{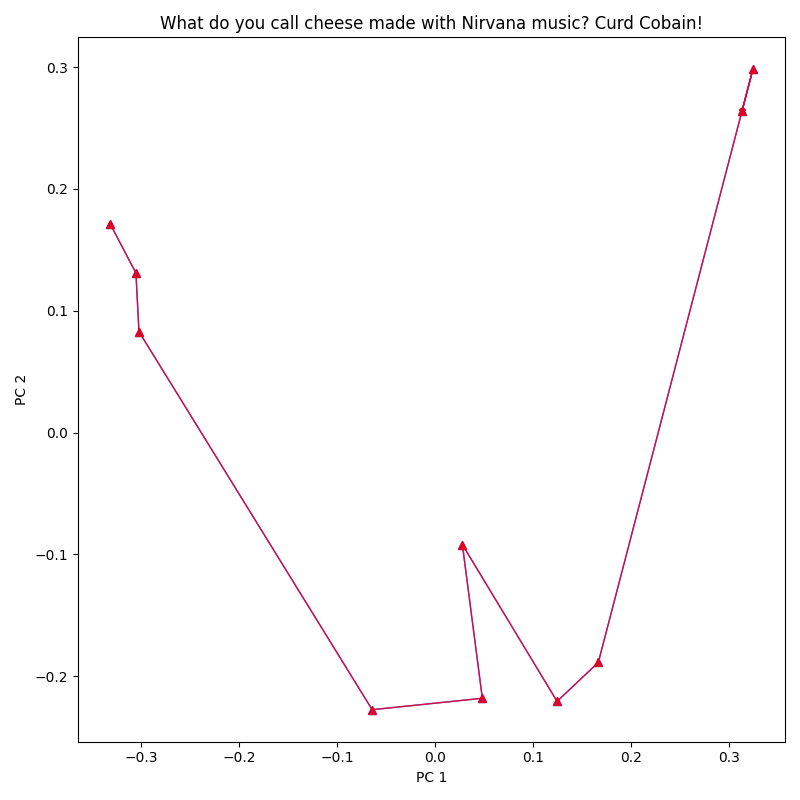}
    \includegraphics[width=0.3\linewidth]{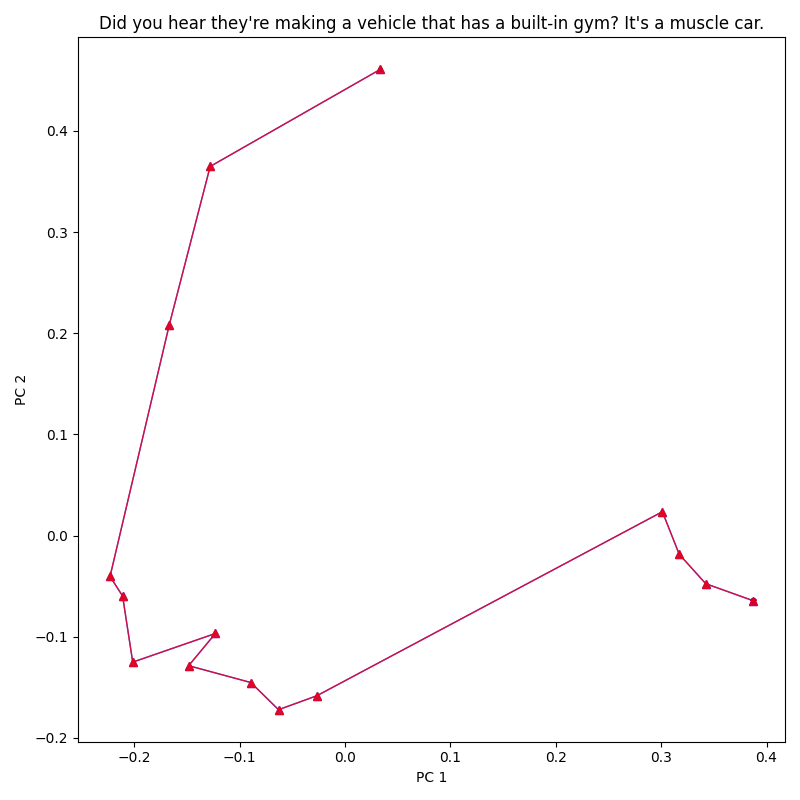}
    \includegraphics[width=0.9\linewidth]{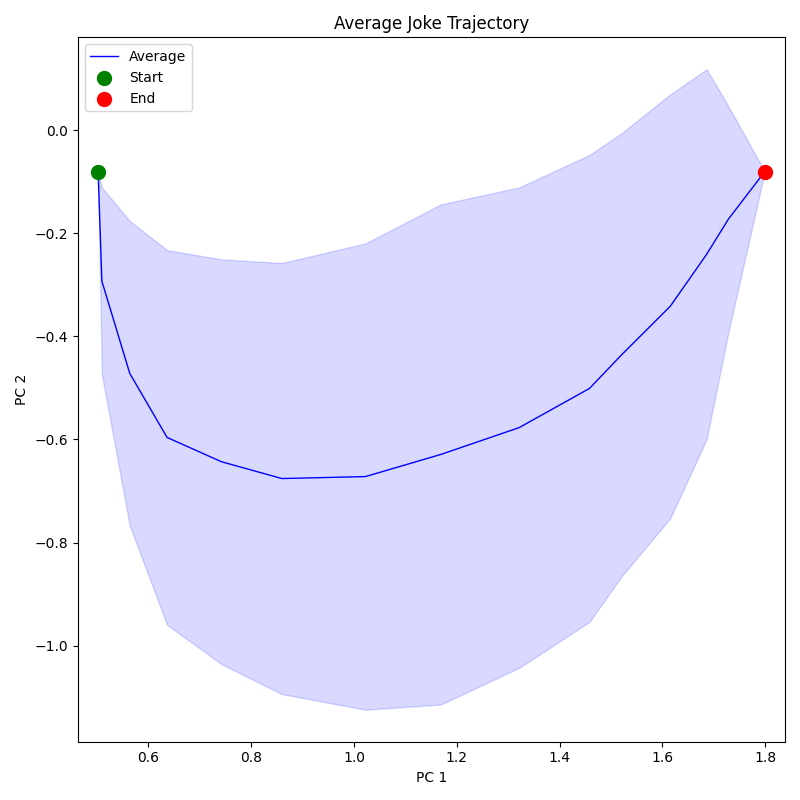}
    \caption{Three different jokes of varying complexity exhibiting similar global ``U-turn'' behavior. From left to right, their transcripts are as follows: ``What do you get if you cross a dog with Penn $\&$ Teller? Two Labracadabradors.'', ``What do you call cheese made with Nirvana music? Curd Cobain!'', ``Did you hear they're making a vehicle that has a built-in gym? It's a muscle car.''}
    \label{fig:placeholder}
\end{figure}

We construct an average trajectory of ``dad jokes'' through contextual embedding space by calculating mean direction and magnitude of trajectory vectors and using this information to reconstruct the average trajectory. 

The mean reconstructed trajectory is indicative of the presence of global curvature in the trajectory of ``dad jokes". The standard error of these data are significant but predictable as many of the jokes have different subject matter and thus different amounts of variance covered by different principle components. This would be a problem for any dataset and is exacerbated by the fact that the points we are decomposing are $\in \mathbb{R}^{768}$. 

\subsection{Exploration of latent space organization by token role}

The garden path analyses establish that incremental token trajectories track ambiguity-related processing. We next ask what kinds of linguistic information appear to organize those trajectories. Since these trajectories are token-based, we want to see if the trajectories look significantly different when the tokens they describe play different roles in the utterance. We do some initial exploration, but many more questions remain under this umbrella for future work.

\subsubsection{RQB2}
We perform an exploratory contextual analysis of sentences containing a common polysemous token, in this case the word ``run'', to examine how structure emerges in the latent space of the model. We know from prior work by \citet{Digutsch2023OverlapIM} and \citet{zimmerman2025tokensoftoverlookedappetizerlarge}, that the semantic content will be a dominant organizing factor for the common polysemous token ``run'', but that other factors such as sentence type may be more dominant in determining the organization of the CLS token. We construct a dataset of sentences with the polysemous token ``run'' consisting of four different sentence types (imperative, declarative, interrogative, $\&$ exclamatory) and examine the contextual increments of each of these sentences to see if the emergent clusters correspond more with sentence type or token sense. We perform this analysis for both the CLS token as well as the common polysemous token. In future work, we want to further understand how these different LLM representations of language connect: in this project, we use features from model internal states as a partial representation of meaning construction. In \citet{zimmerman2025tokensoftoverlookedappetizerlarge}, we explore in more detail what information models aggregate within tokens. In recent work, our experiment implicates necessary features of model representations of language in order to sufficiently explain model behaviour~\citep{tinyblitz}. Eventually we hope these different perspectives will coalesce into a rich computational account of language and how it could be realized under different architectures.

\begin{center}
\begin{figure}[p] 
    \centering 
    \includegraphics[width=\textwidth, height=0.45\textheight, keepaspectratio]{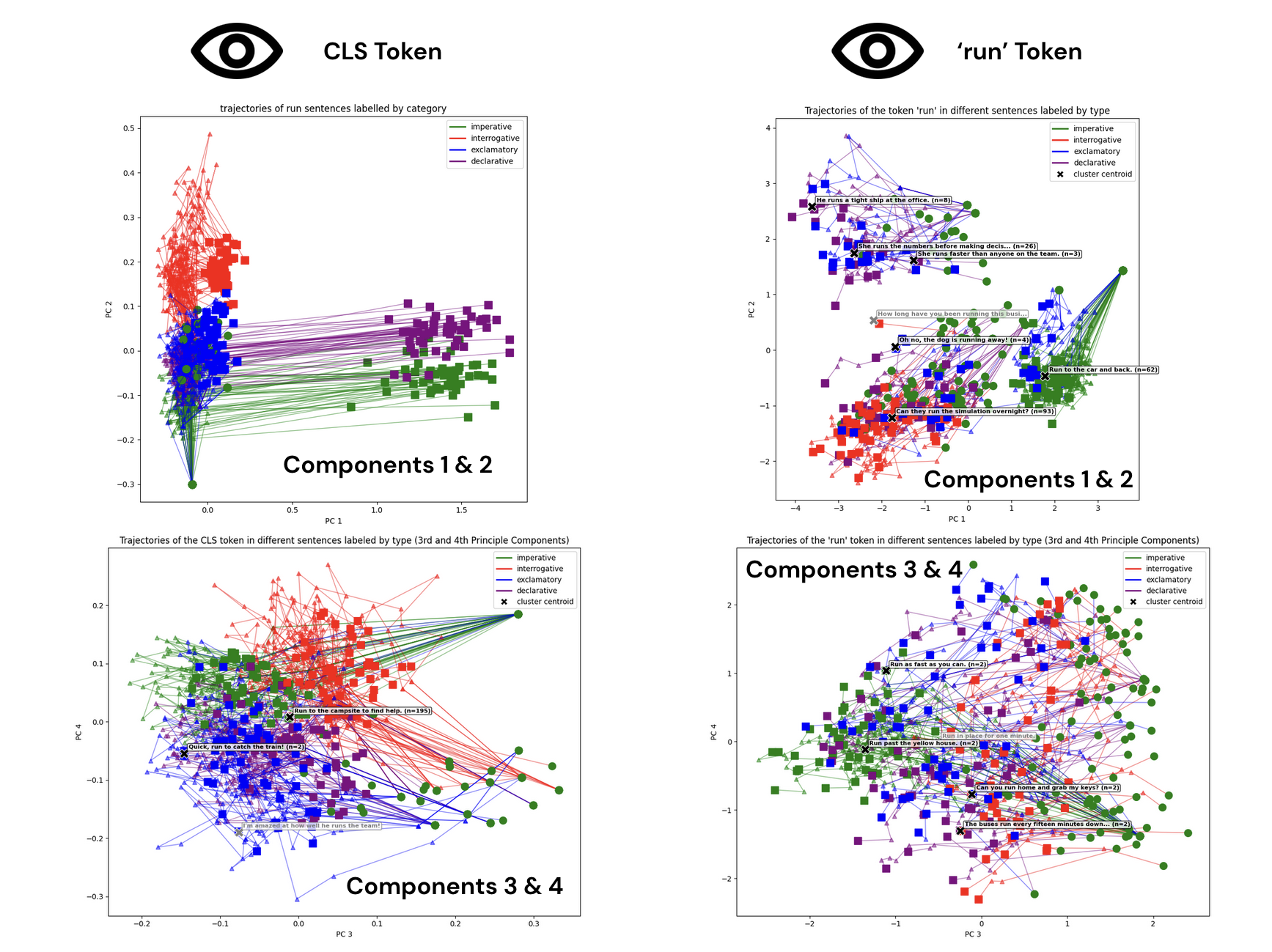} 
    \includegraphics[width=\textwidth, height=0.45\textheight, keepaspectratio]{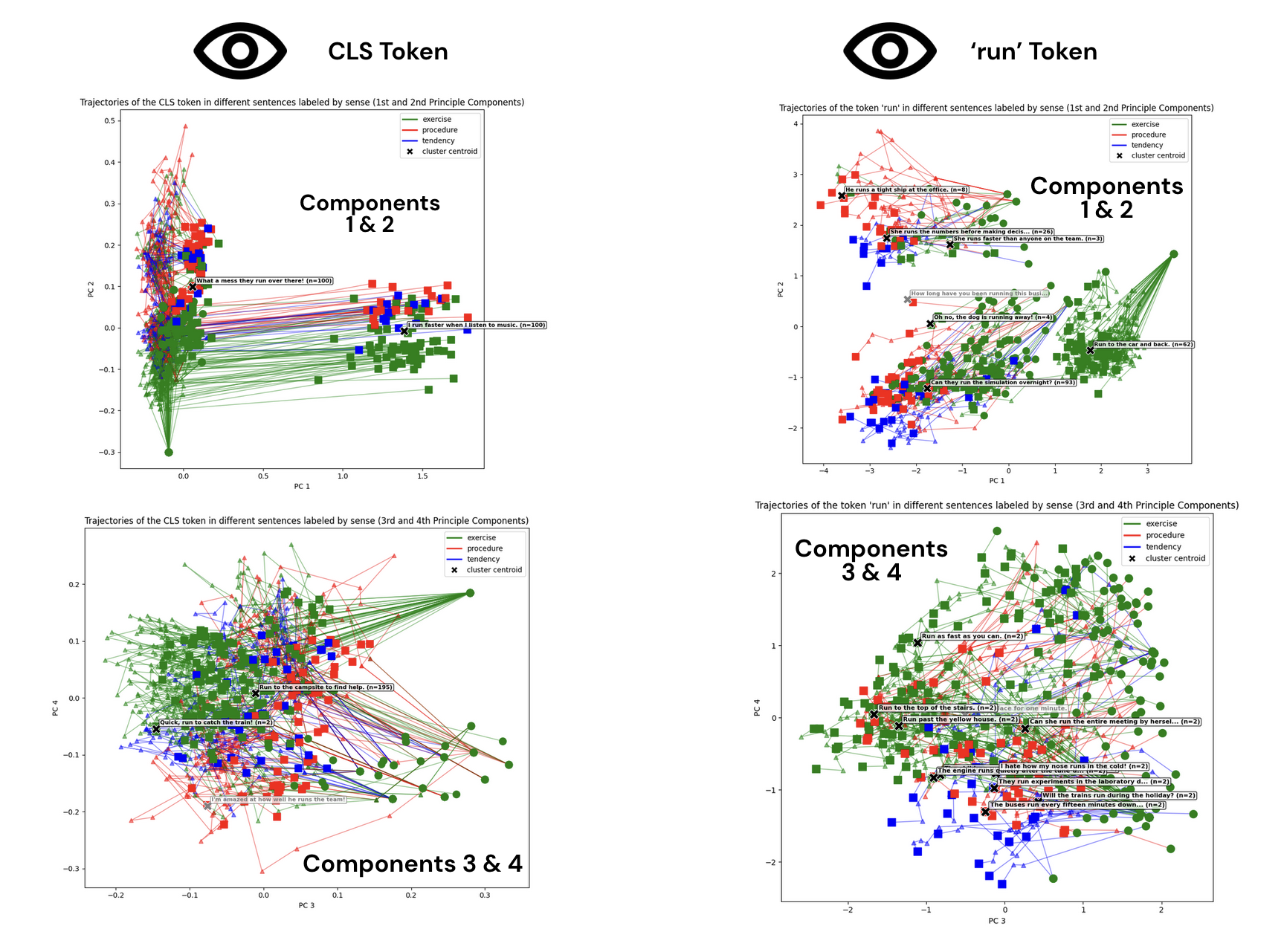}
    \label{fig:flow_chart}
    \caption{Flow charts depicting emergent organization in latent space. The trajectories in the upper 4 plots are colored according to the sentence type and the trajectories in the lower 4 plots are colored according to the sense of the polysemous token. Note that this CLS token is not trained separately, but incorporates information from the other tokens in the context.} 
\end{figure} 
\end{center}

\subsubsection{Results for RQB2}
We visualize these embeddings using dimensionality reduction techniques including PCA and we perform a uniform manifold approximation clustering task in order to see if the latent space representation is more representative of the sense of the polysemous token or the sentence category \cite{mcinnes2020umapuniformmanifoldapproximation}. We see significant separation between different types of sentences when plotting their representation as depicted by CLS tokens. It seems as if the last increment is what determines whether a CLS embedded sentence falls into a certain sentence type category, specifically in the case of imperative and declarative sentences. We come to this conclusion because the last increment in the trajectory of imperative and declarative sentences exhibits an extreme amount of displacement towards the final cluster. We hypothesize that this is due to the presence of the punctuation mark in the last chunk, and that without the punctuation, we might have different clusters more organized by verb sense for the polysemous token, in this case ``run" or some form of the word ``run" including gerund form. 

A possible explanation of the emergent separation of interrogative sentences is the fact that interrogative sentences are the most syntactically distinct from the other sentence types in the dataset. Specifically, the syntax trees of interrogatives often differ from those of other sentence types via ``wh''-movement \cite{chomsky1977wh}.

However, \ref{fig:run_sentences_with_traj} is a PCA decomposed plot of only the CLS token of each sentence containing the polysemous token run. We expected that the emergent organization of the final position of each CLS token would be more dependent on sentence scale information such as sentence type. We compare the emergent organization by sentence scale phenomena of the CLS token to the emergent organization of token level phenomena by plotting the positions of the token ``run'' in each of the sentences. We expect that these clusters will not be as dominantly organized by sentence type and instead will be influenced more by polysemous verb sense.  

The clusters (or lack thereof) observed in \ref{fig:run_token_sentences} are indicative that the sentence type is more dominant in the latent space representation of CLS tokens, whereas the local features such as word sense are more dominant in latent space representations of targeted, single word tokens. These findings correspond with the work of \citet{Digutsch2023OverlapIM} as well as \citet{zimmerman2025tokensoftoverlookedappetizerlarge}. This also conforms with the much more erratic behavior of tokens in the critical region which we observed when examining garden path effects in \ref{sec:gptraj}.

The UMAP clustering shown in \ref{fig:umap_clusters} identified 3 different clusters. The clustering algorithm accurately separated imperative sentences and interrogative sentences, but exclamatory and declarative sentences were frequently drawn into different overlapping clusters. This could potentially be due to the similar narration style between common declarative and exclamatory sentences; they are both often conversational in nature. Nonetheless, we are surprised that the UMAP clustering does not appear to be entirely dictated by the punctuation used in each sentence type.  

The previous plots shown have all been examining the first and second components identified by a 2 dimensional principle component decomposition. We reconstruct these same figures (specifically \ref{fig:run_latent}, \ref{fig:latent_traj}, \ref{fig:emergence_run_sentences}, \ref{fig:run_sentences_with_traj}, \& \ref{fig:run_token_sentences}) in the 3rd and 4th principle component plane as determined by a 4 dimensional principle component decomposition. We analyze these figures to determine whether there are new or similar emergent clusters within the 3rd and 4th principle components of the embedding trajectories. 

The 3rd and 4th principle component decompositions of the ``run'' token trajectories seem to reveal emergent organization by the sense of the token ``run'' rather than the type of the sentence. However, for the 3rd and 4th principle component decompositions of the CLS token, we see less organization around word sense, and more organization by sentence type. The apparent emergence of sentence scale information as an organizing factor for the CLS token in latent space is consistent with the fact that RoBERTa's CLS tokens attend to all of the vocabulary tokens within its respective context window; though it is interesting nonetheless that sentence level information appears without in the form of an aggregate of individual tokens \cite{zimmerman2025locality, zimmerman2025tokensoftoverlookedappetizerlarge, Digutsch2023OverlapIM}. These results hint at a compositionality of CLS tokens that does not necessarily align with with reality: the meaning of a sentence is not necessarily the sum of its individual words.  
A flow chart of various PCA plots is shown in \ref{fig:flow_chart}.

\section{Discussion} 
\label{sec:disc}



In our case study of incremental context, we introduce gross turbulence, net turbulence, stability, and inverse normalized stability as metrics for measuring token trajectories. We calculate average distributions of these metrics and find garden path sentences tend to have more tokens exhibiting less inverse normalized stability than disambiguated sentences. 

As a result of our experiments in \ref{sec:gptraj}, we were able to confirm the findings of \citet{jurayj_garden_2022} that the ambiguity exhibited by garden path sentences can be observed via token level semantic differences in sentences. Furthermore, we make the stronger claim that characteristic token level differences are indicative of garden path sentences. We conclude that, of the variants we compared, the $D_2$ representation of a contextual increment is the best suited for fine grained analysis of ambiguity as exhibited by language models when interpreting garden path sentences and other psycholinguistic phenomena. We come to this conclusion as our results in \ref{section:vectorvsscalar} confirm our hypothesis that the vector representation of token trajectory contains valuable information that is lost when trying to represent trajectory using only a scalar value, specifically the $L_2$ norm. 

To test the plausibility of using incremental trajectories to analyze more general psycholinguistic phenomena, we perform an exploratory analysis on corpora that rely on more global ambiguity or misdirection, specifically ``dad jokes". We calculate the average trajectory of several hundred jokes from a dataset of ``dad jokes" compiled by \citet{silvia_burnett_2025} from the \textit{r/dadjokes} subreddit in RQB1. Upon constructing the average trajectory of dad jokes and plotting this trajectory, we notice a global, sentence scale ``U" shape similar to that exhibited by garden path sentences around the critical region. This is consistent with sentence-scale misdirection being a hallmark of token trajectories for (at least some kinds of) ``dad jokes"; however, we leave further analysis to future work. 

Overall, garden-path information is clearly present in model representations as an utterance unfolds. CLS incremental trajectories distinguish garden-path from non-garden-path sentences with high accuracy. This makes sense, as CLS is a sentence-based representation (occurring at the beginning of every sentence). More surprisingly, comparable information is also recoverable from the trajectory of the first vocabulary token; unlike CLS, this token is not a dedicated sentence representation but whatever word happens to appear first in the sentence.

Together, these findings suggest that utterance-level information is not confined to dedicated sentence-level representations: vocabulary tokens appear capable of encoding both local lexical information and more global properties of the sentence, potentially the utterance.\footnote{Future work should test with longer utterances.} Rather than being concentrated solely near the disambiguating region, ambiguity-related information appears in the relatively far-removed CLS and first vocab token trajectories.

The results suggest both specialization and distribution. CLS appears particularly useful for sentence-level distinctions, implying the emergence of sentence-scale representational organization during training despite the absence of any explicit objective requiring such organization. At the same time, ordinary vocab tokens retain substantial sentence-level information, with different kinds of tokens treating it differently. This has implications for understanding information locality as realized in LLMs. Future work should explore how tokens variously encode sentence-level, local, and additional resolutions, comparing trajectories across token positions and linguistic phenomena. For example, do sentence-terminal punctuation tokens specialize like CLS? To what extent is token content redundant?

\citet{tinyblitz} argued from behavioural evidence that form and function are too entangled in plausible-text-generating architectures. Here, internal-state analyses could be consistent with that interpretation, with individual token representations appearing to combine lexical and utterance-level information, suggesting relatively distributed, potentially redundant, multi-scale representations of linguistic structure.

\subsection{Main Limitations}

In \ref{sec:JokesRQ}, we conduct an analysis of latent space representation of ``dad jokes" as compiled by Silvia and Burnett \cite{silvia_burnett_2025}. These jokes contain elements of misdirection, but the analysis performed is not indicative of misdirection appearing in all varieties of jokes. Our analysis should serve as a study of how sentence scale misdirection manifests itself in latent space, not as an argument that certain linguistic devices are necessary requisites to write jokes.  

The majority of embeddings used in this paper are generated by RoBERTa. These embeddings are located in $\mathbb{R}^{768}$, which makes visualization non-trivial. In many cases, the vast majority of the variance in CWEs is located outside of the 1st and 2nd principle components. In future work, we are interested in exploring other methods of understanding the content of CWEs and in exploring their content beyond a few dominant organizing forces. 

Experiments conducted in this paper are intended to highlight how token level information can reveal meaningful information about meaning construction in larger linguistic scales. We do not present these findings as an argument that incremental token-level analysis is superior to other NLP techniques; in many situations, a simpler, less-resource-intensive approach would probably be more appropriate.

We address these limitations in more detail in \ref{appendix:detailed_limitations}.

\subsection{Future Work}

In future work, we would like to draw more concretely from \textit{stigmergic epistemology} as described in \citet{MARSH2008136}, drawing parallels between stigmergic traces and trajectories that represent utterances as they contextually evolve.

Although we primarily examine target-token-level or CLS-token-level trajectories of sentences for characteristic differences, we would like to examine additional trained (and aggregate) token representations of clause level phenomena, such as clauses in the critical region of garden path sentences. We would like to examine how the ``token perspective'' on the representation of the utterance changes under different architectures (e.g. an encoder-decoder model such as T5~\cite{raffel2023exploringlimitstransferlearning}), and particularly, objectives. Specifically, we would like to look at SpanBERT (which trains with n-gram prediction spans)~\cite{joshi-etal-2020-spanbert}, BERT (which trains CLS with a next sentence objective)~\cite{devlin_bert_2019}, and Lexinvariant LMs (which randomize token vector initializations per forward pass)~\cite{huang2023lexinvariantlanguagemodels}.\footnote{We are interested in how linguistic information manifests across different architectures: Presumably lexinvariant models would have to store some information that would typically be stored in vocabulary tokens elsewhere~\cite{huang2023lexinvariantlanguagemodels}.}

In \ref{sec:JokesRQ}, we conduct exploratory analyses on the trajectory of ``dad jokes", which parallels some of the analysis performed in \ref{sec:gptokens} and \ref{sec:gptraj}. However, the datasets used in those analyses contained equivalent, disambiguated entries for each garden path sentence, which allowed for meaningful statistical validation of many of our hypotheses \cite{silvia_burnett_2025, STURT1999136, PMID:12690829}. For future work, we propose the analysis of monologues as an effective control against jokes as they may reflect a similar tone to many jokes, which are often presented to a crowd like delivering a monologue. If monologues are an effective control group, then we may be able to perform statistical validation of the hypotheses presented in \ref{sec:JokesRQ}. 

For future work we propose the analysis of sentences that do not meet the typical criteria of garden path sentences but may exhibit a parallel pattern of high surprisal followed by reparsing, specifically with respect to word sense, when the co-occurrence statistics of the context are more closely aligned with an irrelevant word sense (in other words, when the correct word sense is statistically disfavored by the context). One such example would be the word ``murder'': this is a polysemous token (although arguably, though punny, potentially a borderline homophonous word with respect to these two senses) as it could refer to a crime or group of crows, but it is much more frequently used to refer to a crime.


Extending the work of \citet{5401099} in developing context-aware bag of words methods, we propose using token-wise methods to develop a ``bag-of-trajectories'' method of clustering to better compensate for polysemy.\footnote{For example, in a corpus consisting of multiple different uses of the word ``bank'', we would collect the embeddings of all the different uses of the word bank, perform a clustering algorithm on the embeddings and then count each appearance within the different clusters. With this method, assuming the clustering algorithm accurately identifies the clusters as the different senses of the token ``bank'', we would have as our result $n$ uses of the noun ``bank'' referring to a financial institution and $m$ uses of the verb ``to bank'' referring to the action of changing the angle of a plane's aileron. This would offer more fine-grained analysis than the classical bag-of-words approach which would only indicate that there were $n + m$ uses of the word ``bank''.}

\section{Acknowledgments}
\label{sec:acknowledgments}

The authors wish to express their appreciation for support from Juniper L. Lovato, Peter Sheridan Dodds, Christopher M. Danforth, Neil Traft, and Alejandro Javier Ruiz Iglesias, and for conversation with Ted Chiang and Kyle Mahowald. This project series has been developed by a working group of Katie Ekstr{\"o}m, Pablo Rosillo-Rodes, Thomas Pashby, Kathryn Cramer, Tabia Tanzin Prama, Ashley Fehr, Calla Beauregard, Parisa Suchdev, Fitzwilliam Keenan-Koch, Collin Andrew Coil, Milo Z. Trujillo, Sharaj Kunjar, Sagar Kumar, Mohsen Ghasemizade, Dakota Marosi, Mikaela Irene Fudolig, Juniper L. Lovato, Peter Sheridan Dodds, Christopher M. Danforth, Alejandro Javier Ruiz Iglesias, and Julia Witte Zimmerman, with additional early support from Denis Hudon, Jonathan St. Onge, Yoshi Meke Bird, and Bradford Demarest.
The authors are grateful for 
support furnished by 
the National Science Foundation 
(Grant \#2242829).
The authors acknowledge the Vermont Advanced Computing Center (VACC) at the University of Vermont for providing computational resources that have contributed to the research results reported within this paper. Specifically, this material is based upon work supported by the National Science Foundation under Award No. 2242829.
Through the duration of this project, G.W.S. was supported as an intern under the mentorship of J.W.Z. as a part of the Vermont EPSCoR SOCKS REU, a program supported by the National Science Foundation under Award No. 2242829.

\bibliography{sn-bibliography}

\begin{appendices}

\section{Author Contribution Statement}
\label{sec:ACS}
G.W.S. led investigation, validation, formal analysis, software, writing, and visualization, and contributed to methodology, resources, and conceptualization. J.W.Z. led the conceptualization, project administration, resources (including data availability),  and supervision, and contributed to methodology, writing (original draft as well as review $\&$ editing), and visualization. 
\textit{Note: some content in this paper is adapted from \citet{zimmerman2025locality} (doctoral dissertation) and \citet{zimmerman2025tokensoftoverlookedappetizerlarge}, as well as earlier drafts of related projects.}

\section{Ethical Considerations Statement} 
\label{sec:Ethics}
We are not aware of any privacy, surveillance, or (mis)representation concerns associated with the datasets directly\footnote{However, there are ethical concerns around the creation and use of Generative AI models including LLMs (including MLMs). This topic is out of scope of this project, but we want to acknowledge its existence and its relevance to this work. Besides using such models as subjects in our experimental pipeline, G.W.S. and J.W.Z. used an institutional Copilot account at various points throughout this work.} used in this paper as the data used are not representative of any human subjects, but instead used to form a representation of linguistic phenomena. Furthermore there is no personal identifying information (PII) within any of the data used nor is there any anonymized PII that susceptible to a re-identification style attack~\cite{sondeck2025practicalreadytousemethodologyassess}. All of the original data used in this project were created by the authors for the sole purpose of fulfilling the linguistic requirements associated with each RQ.

The data used in this paper that are not the authors' original work are freely available, specifically the dataset from Wang et al., which compiles the datasets from \citet{STURT1999136} and \citet{PMID:12690829}, as well as the dataset from \citet{silvia_burnett_2025}. The \citet{silvia_burnett_2025} dataset contains jokes from the \textit{r/dadjokes} subreddit and may contain content that some find offensive. 

\section{Code and Data Availability Statement}
\label{sec:dataavail}
The code and data for this paper are available on \href{https://github.com/graysonstorer/contextual-trajectories/}{github}. Code is licensed under the MIT license, and data is shared under the MIT license unless otherwise noted. The garden path sentences used in \ref{sec:gptraj} and \ref{sec:gptokens} were taken from Wang et al. \cite{wang_causality_2024}, which compiles two different datasets, one from Grodner et al. \cite{PMID:12690829} and one from Sturt et al. \cite{STURT1999136}. This dataset is available on the following public GitHub \href{https://github.com/wangdaphne/garden-path-SF-dataset}{repository}. The jokes used in \ref{sec:JokesRQ} were taken from Silvia and Burnett \cite{silvia_burnett_2025} and are freely available at \href{https://osf.io/kynm7/files/osfstorage}{osf.io}.

\section{Caveat: difficulties in interpreting contextual changes in LLM representations}
\label{appendix:caveat_context_change_interpretation}

\begin{figure}
    \centering
    \includegraphics[width=0.45\linewidth]{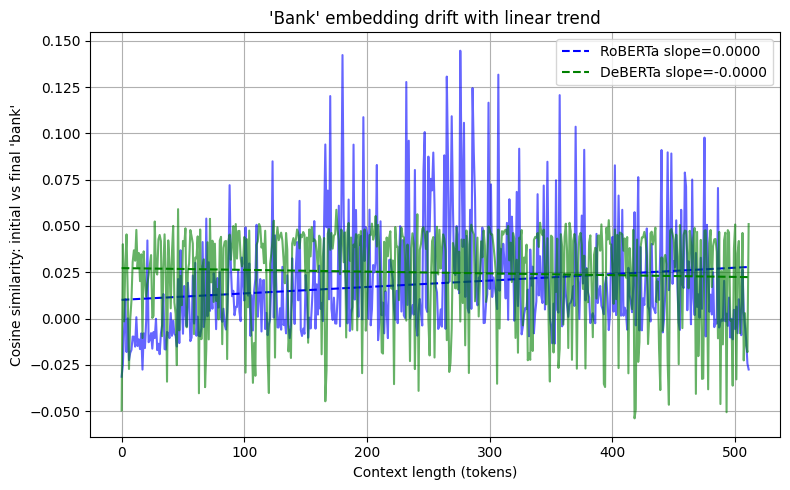}
    \includegraphics[width=0.45\linewidth]{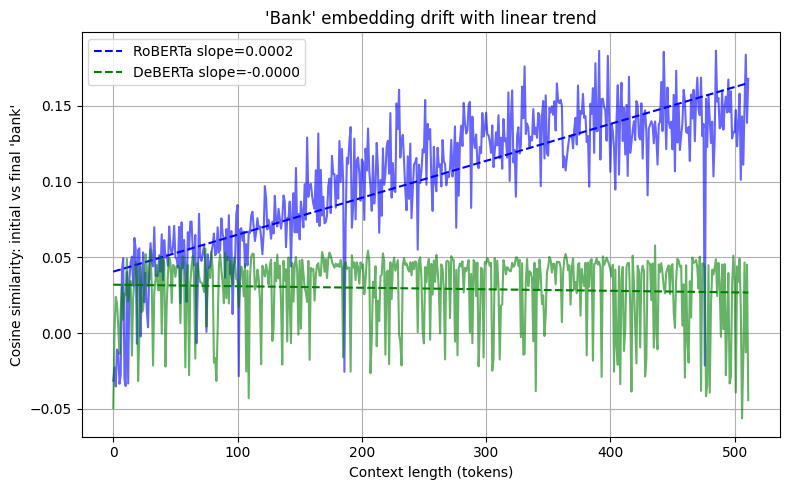}
    
    \caption{These charts are not particularly meaningful in and of themselves, but we included them as an example of the caveats we are bringing up. First, architecture matters to epistemology: In the RoBERTa model, the attention process compresses position along with all other linguistic information accessible to the model. In DeBERTa, position is stored separately, leading to different behaviour in the representations between the two models as (architectural) context is added, even though both models do share significant aspects of their behaviour. Second, because (broadly-construed) architecture matters, we cannot assume our experience of language (via any representation: formal, informal, etc.) will map cleanly to what happens in LLMs, even though the language we both generate has significant commonalities. As what is essentially random gibberish is added to the (architectural) context, we doubt most people would perceive much difference between ``... (234 tokens of gibberish) ... bank'' and ``... (240 tokens of gibberish) ... bank''; however, the model continues to adjust its representations.
    These two graphs show when the \textit{bank} token is anywhere in the context (left) versus when it is at the end (right). The ``context'' here is a toy context: just random other tokens from the model's vocabulary. Additionally, as we were hoping would be the case, DeBERTa, the model with disentangled attention does not obviously show a correlation between context length and magnitude of vector change, compared to RoBERTa.}
    \label{fig:bank_random_positions}
\end{figure}

It is not obvious how to interpret the change in the vectors from initial to final position. It is some kind of ``change in meaning'' for the model, but all the details are unresolved. For one thing, most LLM architectures ensure the position will change,  whether or not there is any sense in which the relevant token's meaning is invariant. The positions will probably change more given the size of the architectural context simultaneously propagated through the model. For example, see Fig.~\ref{fig:bank_random_positions}. This is one of the very first sub-problems that needs to be addressed somehow. One option would be to normalize exemplar contexts of length N using an average change based on that context size, or to group contexts by length or position. Another option would be to try to estimate how much variation in vector similarity is possible between two contextualized words we judge as having ``the same meaning'' (similar to some existing WSD approaches). There are plenty of ways to chip away at this, but we wanted in this paper to clearly signpost it as an open question. 

\section{Defining Contextual Increments}
\label{appendix:defining_contextual_trajectory}

\subsection{RQ1:} We propose various different mechanisms for quantitative representation of the displacement between each chunk in an incremental representation of a sentence. We propose the following 4 functions:

\begin{itemize}
    \centering
    \item $D_1 : T \rightarrow \mathbb{R}$ s.t. $\forall t_n \in T$, $D_1(t_n) = ||t_n - t_{n-1}||_2$. 
    \item $D_2 : T \rightarrow \mathbb{R}^{768}$ s.t. $\forall t_n \in T$, $D_2(t_n) = t_n - t_{n-1}$. 
    \item $D_3 : T \rightarrow \mathbb{R}$ s.t. $\forall t_n \in T$, $D_3(t_n) = ||t_n - t_0 ||_2$.
    \item $D_4: T \rightarrow \mathbb{R}^{768}$ s.t. $\forall t_n \in T$, $D_4(t_n) = t_n - t_0$.
\end{itemize}

$D_2$ and $D_4$ will represent trajectory as a vector quantity while $D_1$ and $D_3$ will represent trajectory as solely the displacement between subsequent word embeddings. We hypothesize that the vector representations will contain more valuable information than the scalar representation due in part to the significance of direction within latent space representations of ambiguous tokens \cite{wiedemann2019doesbertmakesense}. 

We perform a test of each representations resilience to perturbation in order to ensure that our representation of incremental context is more sensitive in changes to semantic meaning rather than changes to syntax. We measure resilience to syntactic perturbation by analyzing the performance each representation on a broad sentence variety classification task. We consider a representation to be resilient to syntactic perturbation if the representation can reveal value information about the category of sentence despite changes to syntax including capitalization and punctuation. We test this quality in each representation by evaluating the performance of a classifier on a sentence type classification problem using data generated using $D_{1, \dots, 4}$ on textual samples with altered capitalization and removed punctuation. We hypothesize that the function $D$ that yields data which the classifier can best classify despite syntactic perturbation will offer the clearest representation representation of sentence scale information such as sentence type. We use resilience to perturbation as a proxy measurement of each representation's ability to prioritize accurate representation of information such as sentence type through pragmatic, natural language utterances.

\subsection{RQ2:} LLMs assign each token in the context a position. Which position is the most relevant for our goals? We test this by looking at the trajectories according to each token in the context. We define various metrics to quantify different properties of token trajectories and test how these metrics reflect properties of garden path sentences and their equivalent disambiguated versions. We present summary statistics of trajectory behaviors across a variety of different tokens from garden path and disambiguated sentences. 

\subsection{RQ3:}\label{sec:rq3} We present an exploratory analysis of whether our implementation of incremental context is invariant across different model architectures. We hypothesize that our representation of incremental context will yield similar results as shown by the average Euclidean distance and cosine similarity between garden path and disambiguated sentences for every increment of context added. We collect these statistics by examining the behavior of hinge tokens and CLS tokens from garden path sentences across contextual trajectory for a variety of models based on BERT but using different weight quantization and attention mechanisms, specifically DeBERTa, RoBERTa, and DistilBERT \cite{he2021debertadecodingenhancedbertdisentangled,liu_roberta_2019,sanh2020distilbertdistilledversionbert}. 

\subsection{Results for RQ1}
\label{section:vectorvsscalar}

We test 4 different representations of trajectory for single token increments, 2 that represent trajectory as a vector quantity calculated by performing vector subtraction and 2 that represent trajectory as a scalar quantity calculated by taking the $L_2$ norm between two vectors. We evaluate each representation of trajectory by training a model to classify sentence type given the trajectory of the CLS token as represented by the functions $D_{1, \dots 4}$ altered by various perturbations to text including altered capitalization and removed punctuation. We chose this test as it is a reasonable proxy for the ability of the representation to reflect sentence scale information despite common perturbations in natural language (e.g. Tumblr speak or missing punctuation). The performance of each classifier confirms our hypothesis that the richer vector structure preserved by the functions $D_2$ and $D_4$ contains information about the location of embeddings in latent space that is at least beneficial and at most necessary for a task like sentence classification. $D_1$ and $D_3$ representations were able to be classified despite perturbations to capitalization; however, the classifiers using these representations showed significant declines in performance after removing punctuation. This is indicative that the $D_1$ and $D_3$ representations of trajectory are syntax dominant, potentially due to the loss of information preserved in the vector representations. The training and validation accuracy curves for the classifiers trained on each representation are shown as \ref{fig:representationtesting_perturbation_D_4}, \ref{fig:representationtesting_perturbation_D_2}, \ref{fig:representationtesting_perturbation_D_3}, and \ref{fig:representationtesting_perturbation_D_1} in the appendix.


\subsection{Results for RQ2}
\label{sec:gptokens}
We examine the trajectory through embedding space of every single token in the garden path and disambiguated sentences. We see that hinge variety tokens (tokens the meaning of which changes in garden path sentences) exhibit characteristics of lingering misinterpretation while less ambiguous tokens such as function words have similar starting and ending points regardless of whether or not they are in a garden path sentence. 
\begin{figure}[h]
    \centering
    \includegraphics[width=0.6\linewidth]{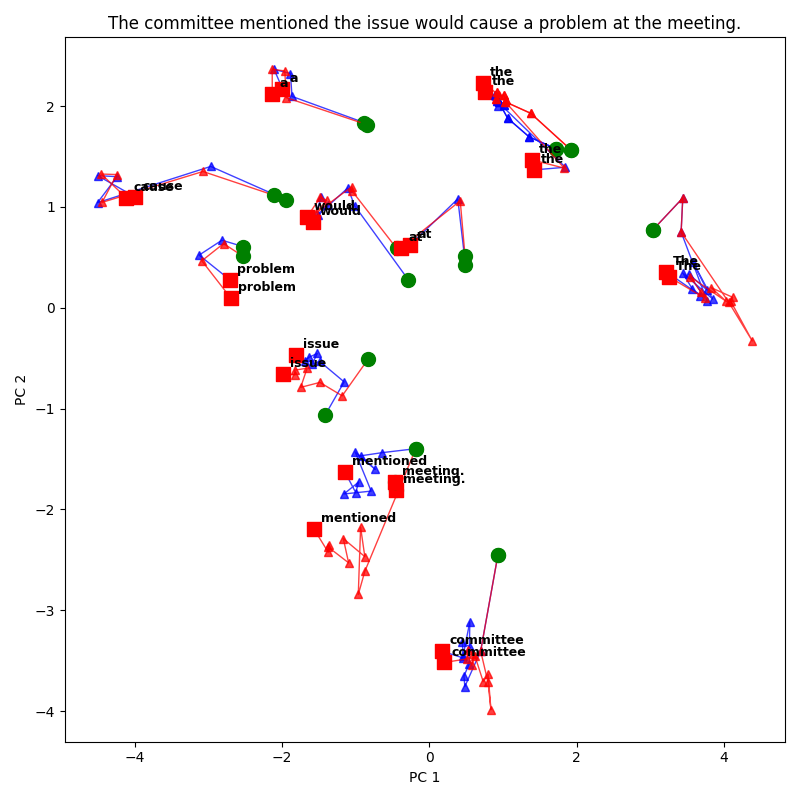}
    \caption{Every token in a sentence has its own respective trajectory for the duration that it is in the sentence. This plot shows the PCA decomposition of the trajectory of every single token in a garden path sentence, plotted in the same PCA space. In this case the sentence is ``The committee mentioned the issue would cause a problem at the meeting.''}
    \label{fig:everysingletoken}
\end{figure}

When performing PCA only on the hinge token embeddings, we see that there is lingering misinterpretation in the model at the end of the sentence. A portion of the distance between final embeddings can be explained by the small syntactic difference between the two sentences, though we also suspect that the lingering disparity in final position is indicative that latent space is not determined entirely by ``meaning'' per se, but also by properties such as a presence or lack of ambiguity.

\begin{figure}[h]
    \centering
    \includegraphics[width=0.3\linewidth]{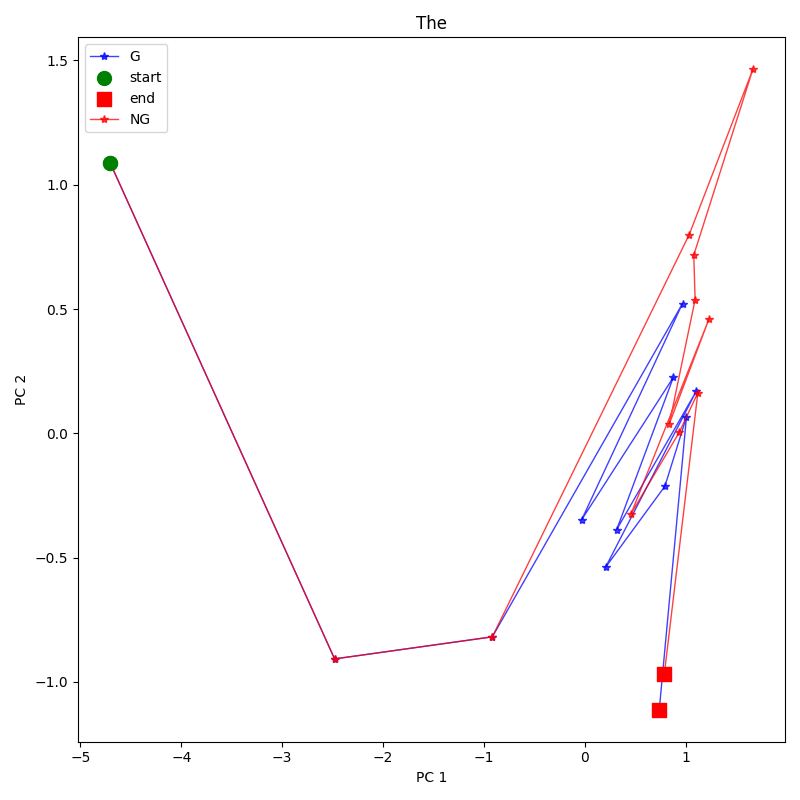}
    \includegraphics[width=0.3\linewidth]{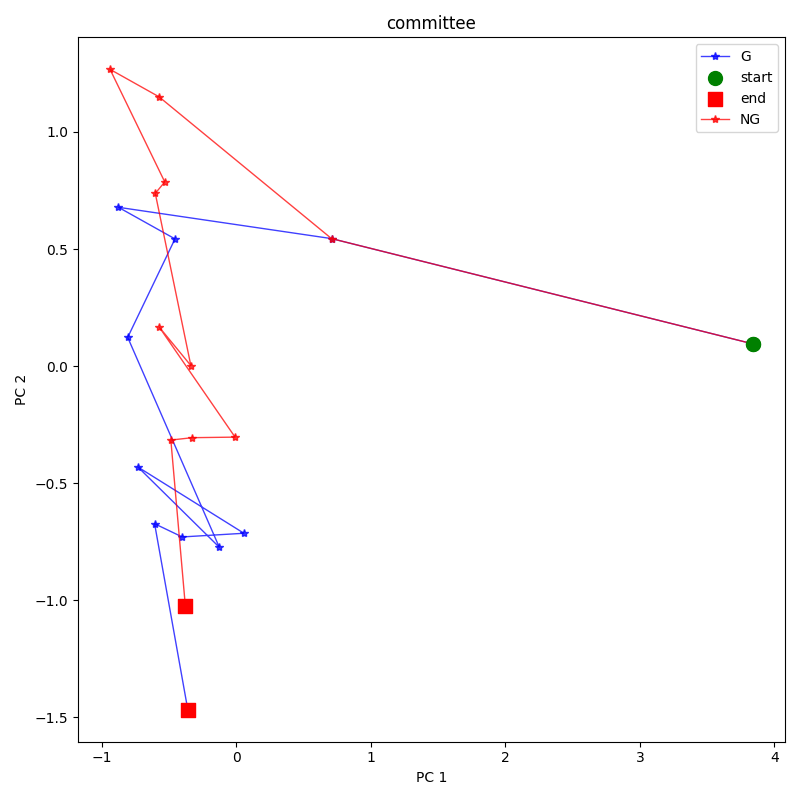}
    \includegraphics[width=0.3\linewidth]{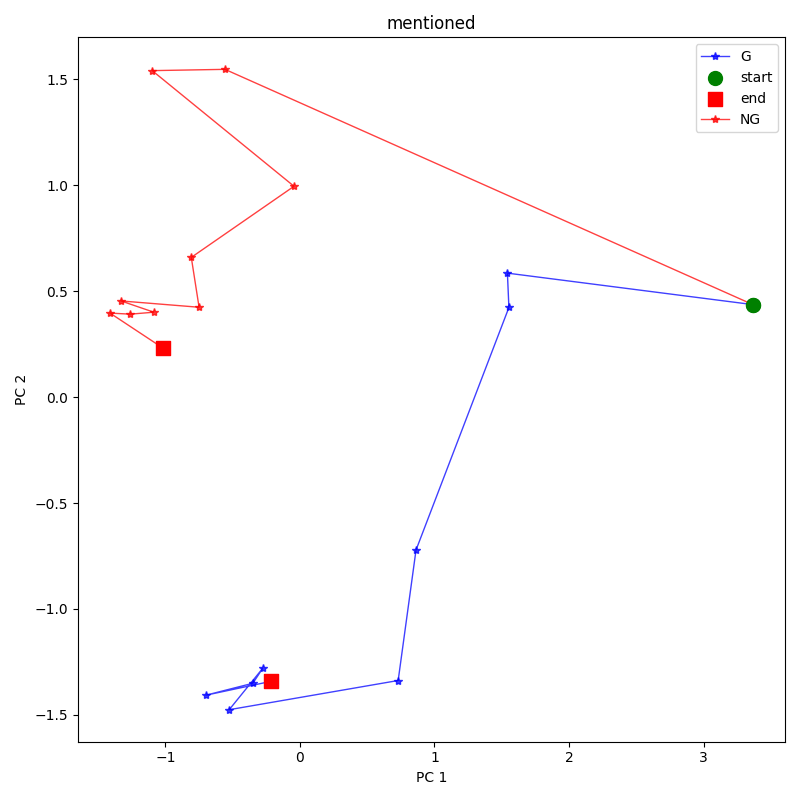}
    \includegraphics[width=0.3\linewidth]{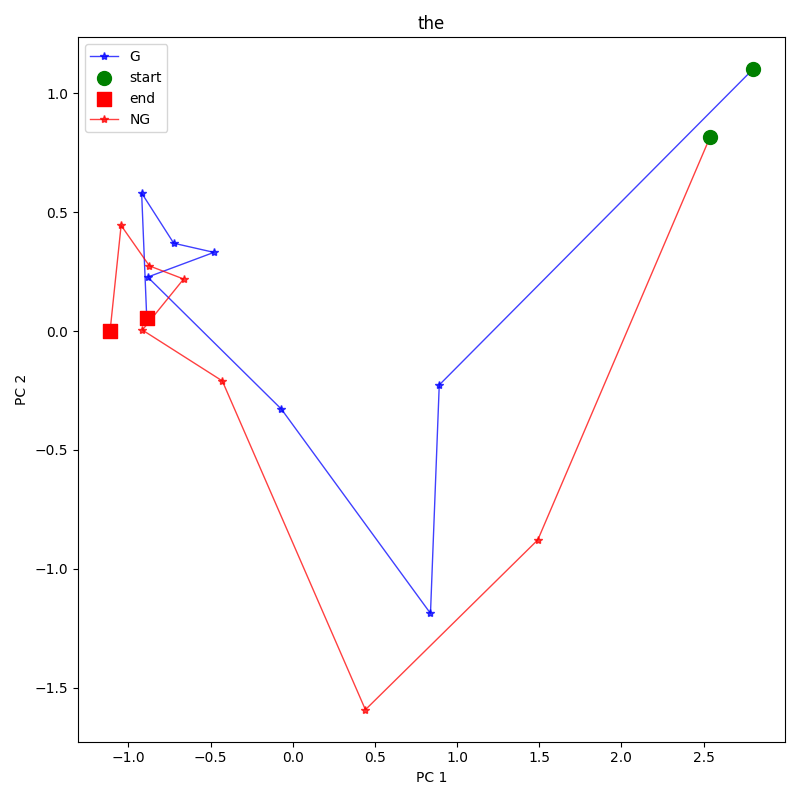}
    \includegraphics[width=0.3\linewidth]{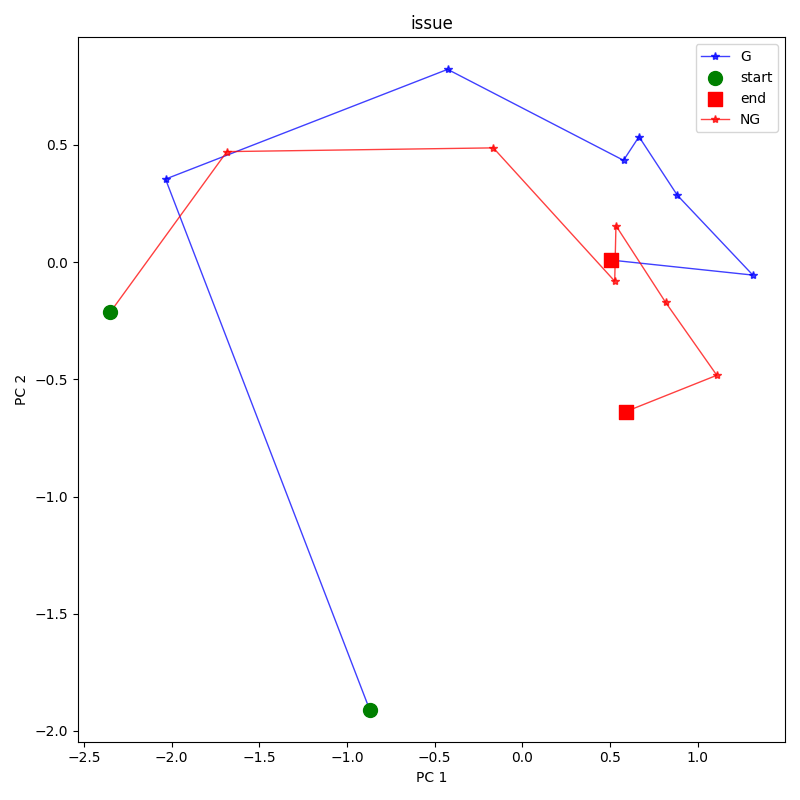}
    \includegraphics[width=0.3\linewidth]{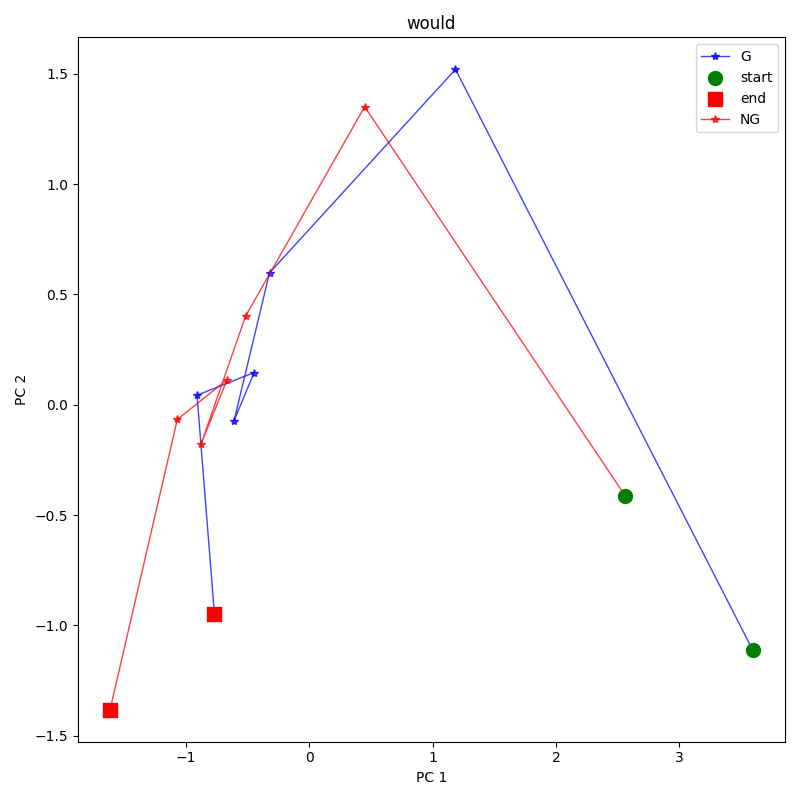}
    \includegraphics[width=0.3\linewidth]{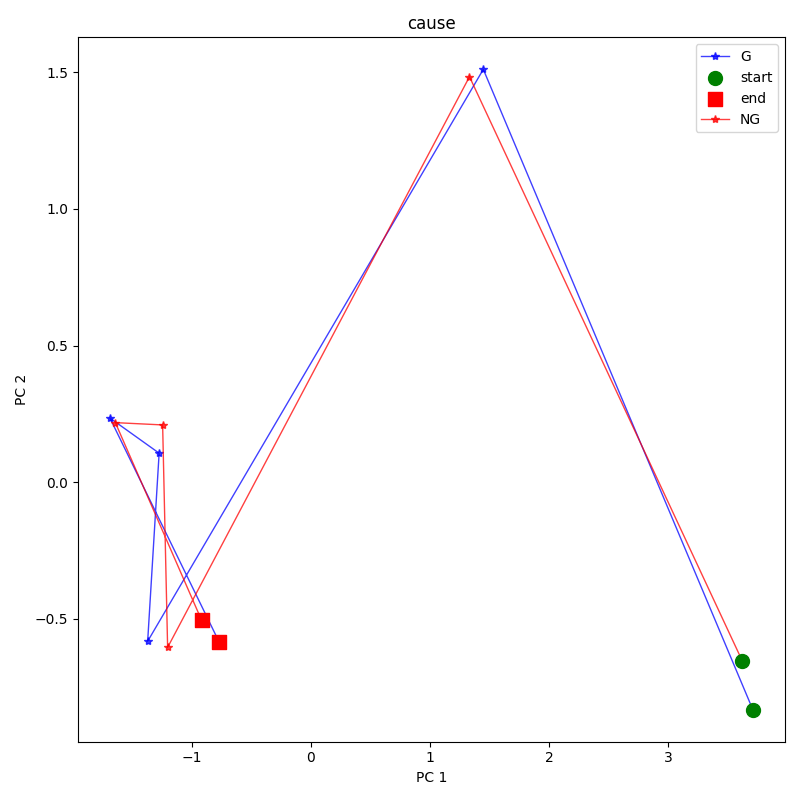}
    \includegraphics[width=0.3\linewidth]{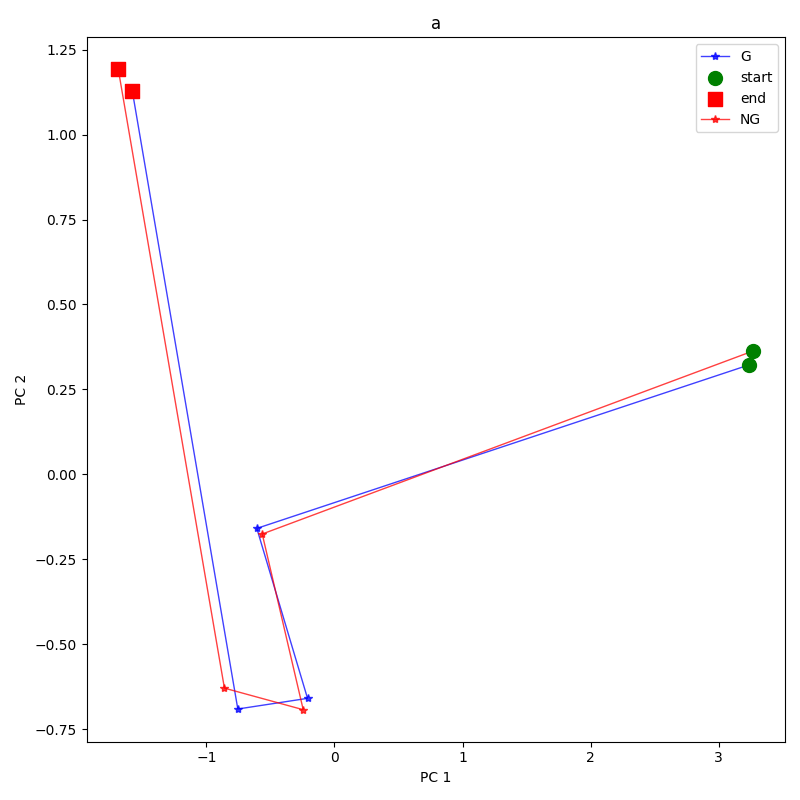}
    \includegraphics[width=0.3\linewidth]{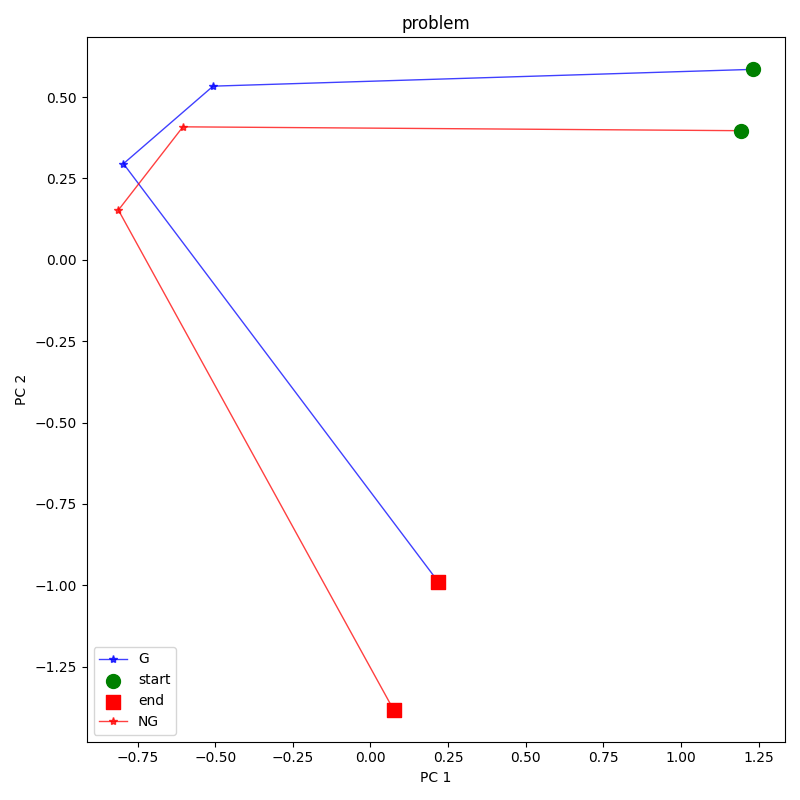}
    \includegraphics[width=0.3\linewidth]{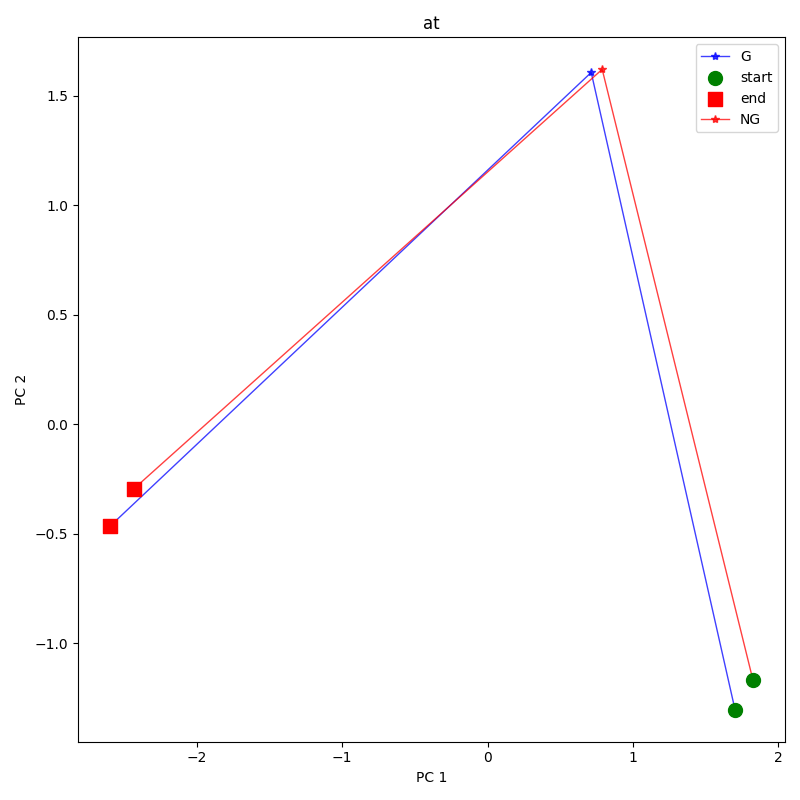}
    \includegraphics[width=0.3\linewidth]{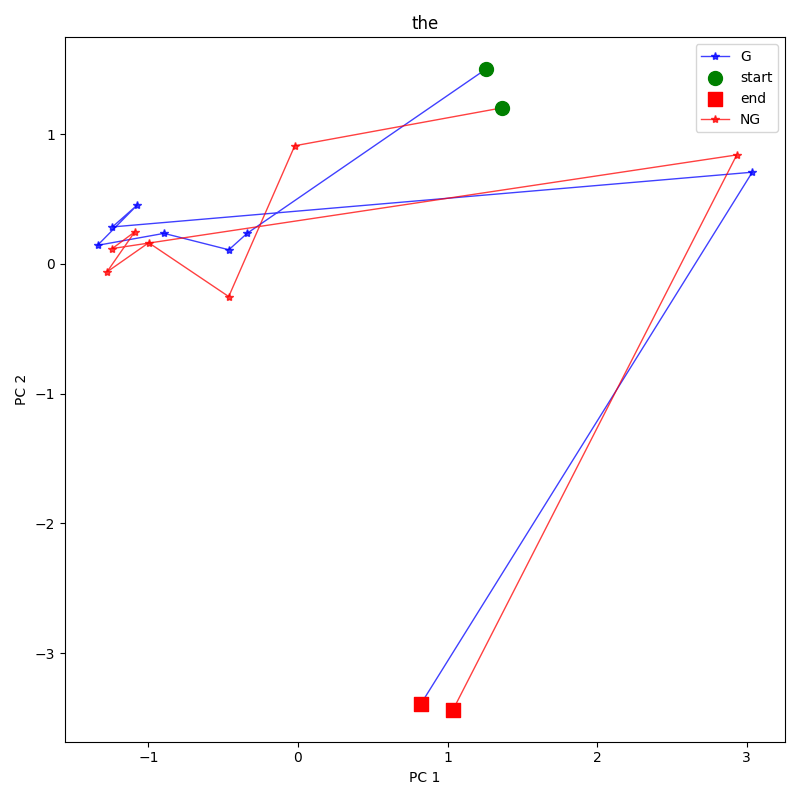}
    \includegraphics[width=0.3\linewidth]{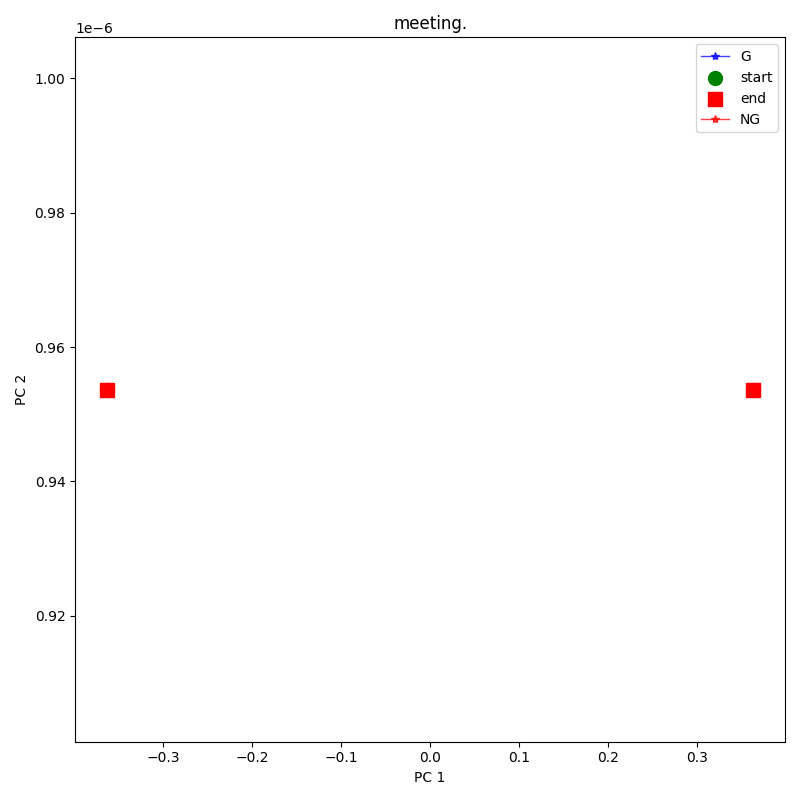}
    \caption{PCA plots of the trajectory of every word in the garden path sentence: ``The committee mentioned the issue would cause a problem at the meeting.'' Note that in the last PCA plot, the red square on the left is the last token in the garden path sentence and the red square on the right is the last token in the disambiguated sentence.}
    \label{fig:hugeploteverysingletoken}
\end{figure}


Our central hypothesis regarding garden path detection is that tokens within a garden path sentence will exhibit more erratic movement than tokens in an equivalent disambiguated sentence. One such measure is gross token turbulence $\kappa(t)$ which we define as follows: 
\begin{displaymath}
    \kappa(t) = \sum_{i= 1}^{|T|}|| \phi(t_{i-1}) - \phi(t_i)||_2
\end{displaymath}
where $T$ is the set of increments containing the token $t$ and $\phi$ is a function mapping the token $t$ into the embedding space. 
We define net turbulence $\kappa_n(t)$ as follows:
\begin{displaymath}
    \kappa_n(t) = || \phi(t_0) - \phi(t_{|T|})||_2
\end{displaymath}
The stability of a token $S(t)$ is defined as the ratio of net turbulence to gross turbulence: 
\begin{displaymath}
    S(t) = \frac{\kappa_n(t)}{\kappa(t)}
\end{displaymath}
The inverse normalized stability of a token $E(t)$ is defined as follows: 
\begin{displaymath}
    E(t) = \frac{|| \phi(t_0) - \phi(t_{|T|})||_2}{\sum_{i= 1}^{|T|}|| \phi(t_{i-1}) - \phi(t_i)||_2} \cdot |T|
\end{displaymath}
We perform an analysis of the most turbulent token in each sentence and look at whether there are shared tokens that seem to exhibit more gross turbulence across different sentences. 


We also examine the most frequently occurring tokens that exhibit the highest turbulence in disambiguated sentences in order to find any categorical differences. 


The key difference between the two categories is that there are more words exhibiting high turbulence in the disambiguated sentences than in the garden path sentences. For both categories, the articles ``the" and ``The" are the most frequent highly turbulent words, but they are almost 2 (1.75) times as frequent in the garden path sentences as shown in \ref{fig:frequent_most_turbulent}.

\begin{figure}
    \centering
    \includegraphics[width=0.4\linewidth]{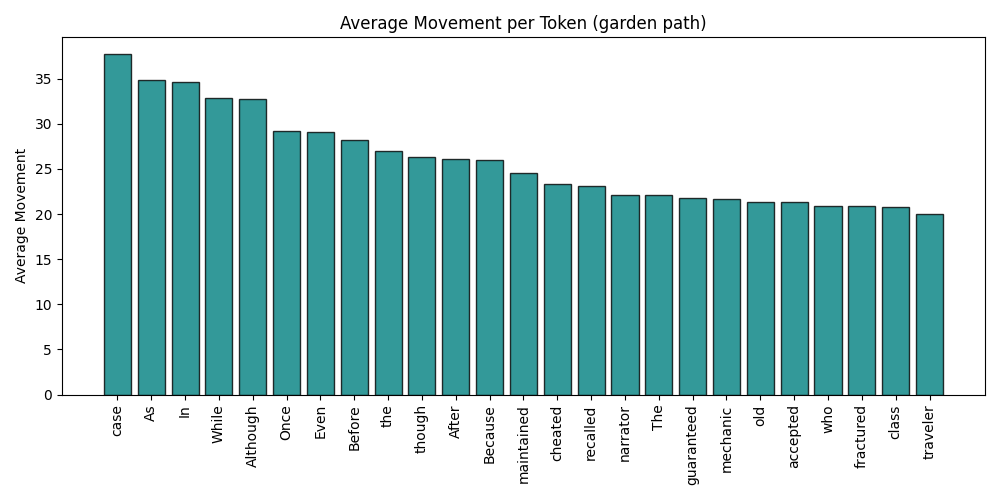}
    \includegraphics[width=0.4\linewidth]{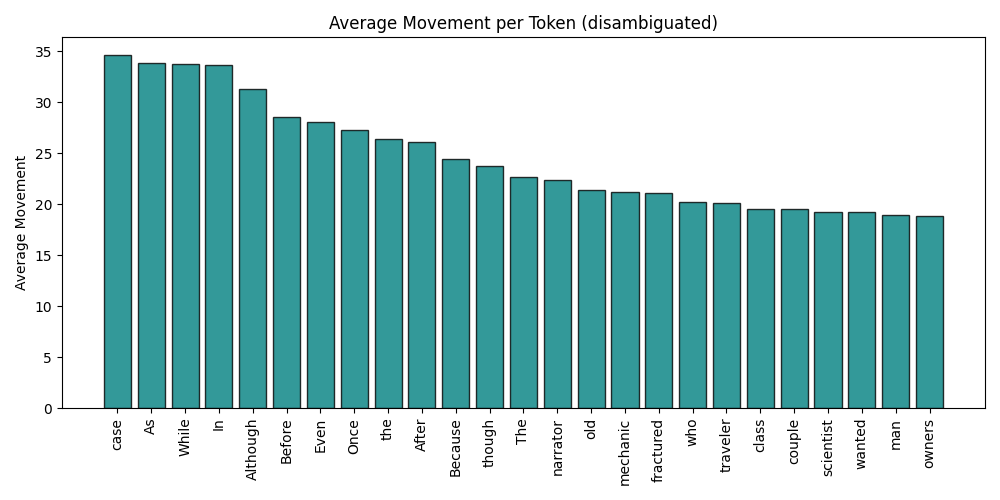}
    \includegraphics[width=0.4\linewidth]{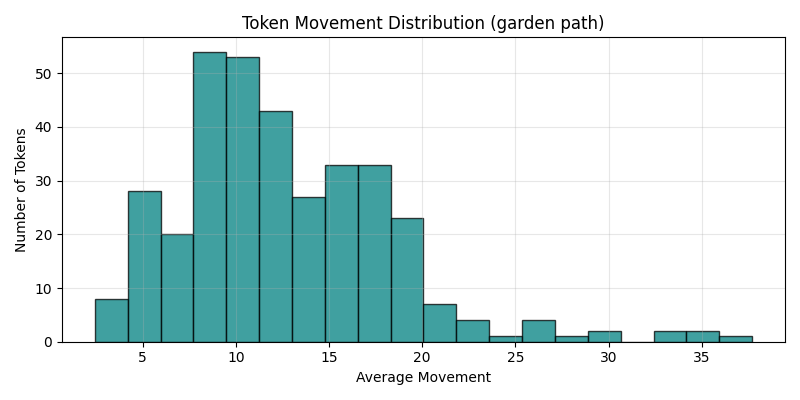}
    \includegraphics[width=0.4\linewidth]{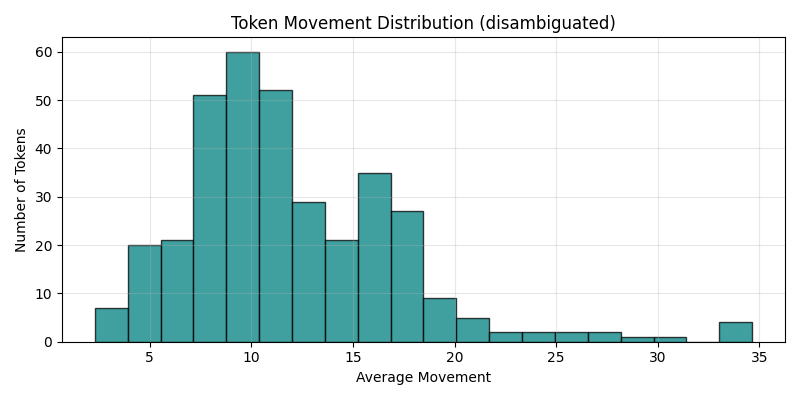}
    \caption{Average movement per token and token movement distribution for disambiguated and garden path sentences.}
    \label{fig:token_distribution}
\end{figure}

A Kolmogorov-Smirnov test of the two distributions of gross turbulence (garden path and disambiguated) revealed a KS-statistic of 0.0891 and a $p$-value of 0.2132, indicating that the difference between the two distributions is not statistically significant.

\begin{figure}
    \centering
    \includegraphics[width=0.4\linewidth]{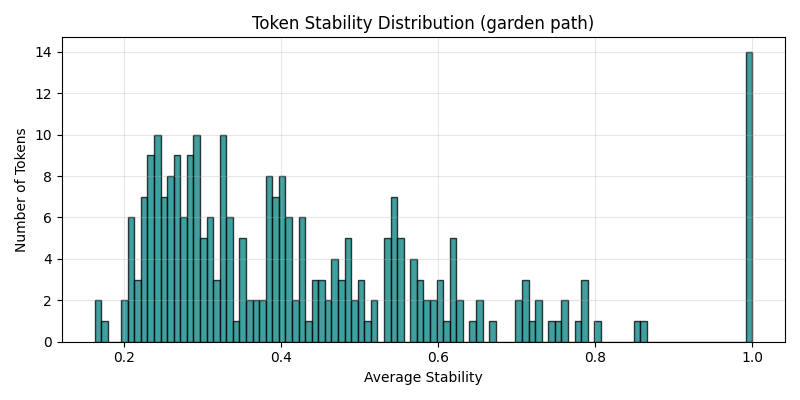}
    \includegraphics[width=0.4\linewidth]{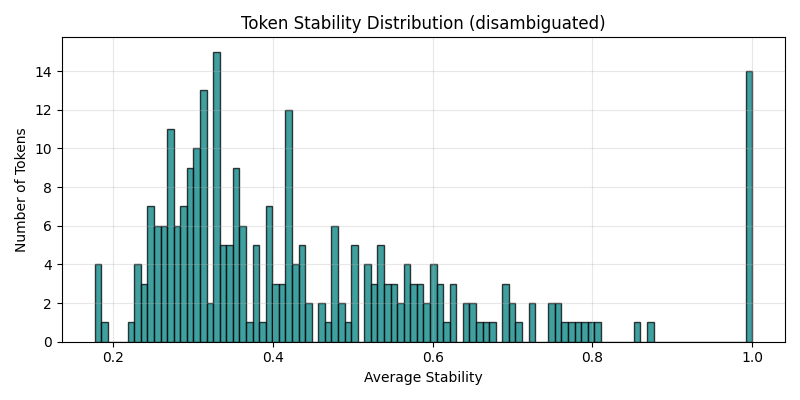}
    \caption{Distribution of token stability across garden path and disambiguated sentences}
    \label{fig:stability_distribution}
\end{figure}

We calculate the distributions of token stability across garden path and disambiguated sentences as shown in \ref{fig:stability_distribution}.

A Kolmogorov-Smirnov test of the two token stability distributions revealed a KS-statistic of 0.1071 and a $p$-value of 0.08138, indicating a more meaningful relationship than gross turbulence, but still not statistically significant.

\begin{figure}
    \centering
    \includegraphics[width=0.4\linewidth]{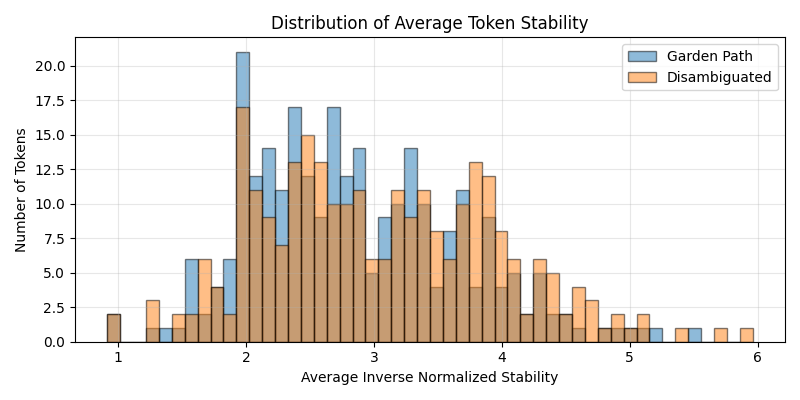}
    \caption{Distribution of the inverse normalized token stability for garden path and disambiguated sentences}
    \label{fig:inverse_norm_stab}
\end{figure}

Finally, we calculate the distributions of inverse normalized token stability across garden path and disambiguated sentences as shown in \ref{fig:inverse_norm_stab}. 

A Kolmogorov-Smirnov test of the two inverse normalized token stability distributions (\ref{fig:inverse_norm_stab}) revealed a KS-statistic of 0.1268 and a $p$-value of 0.0207, which is below the standard threshold of $p = 0.05$ indicating a statistically significant difference between the two distributions. This finding is indicative that, on average, tokens within garden path sentences exhibit characteristically different turbulence throughout their trajectory with respect to the average turbulence in the trajectory of disambiguated sentences. 

When we normalize the distributions of inverse normalized token stability by distance (in terms of word increments) from the critical region, we see that the majority of tokens in disambiguated sentences have an inverse normalized stability below 5, whereas the mass of the distribution of tokens in garden path sentences are not as densely concentrated within the same region. 

\begin{figure}
    \centering
    \includegraphics[width=0.4\linewidth]{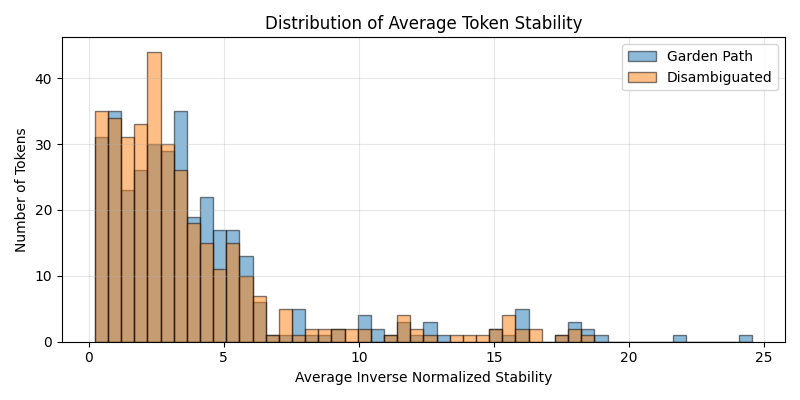}
    \caption{Distribution of the inverse normalized token stability for garden path and disambiguated sentences when normalizing by distance from the critical region in terms of word increments.}
    \label{fig:inverse_norm_stab_dist_norm}
\end{figure}

A Kolmogorov-Smirnov test of the two distributions shown in \ref{fig:inverse_norm_stab_dist_norm} revealed a KS-statistic of 0.1075 and a $p$-value of 0.03203, which is below the standard threshold of $p = 0.05$, indicating a statistically significant difference between the two distributions.

\subsection{Results for RQ3}

\begin{figure}[h!]
    \centering
    \includegraphics[width=0.4\linewidth]{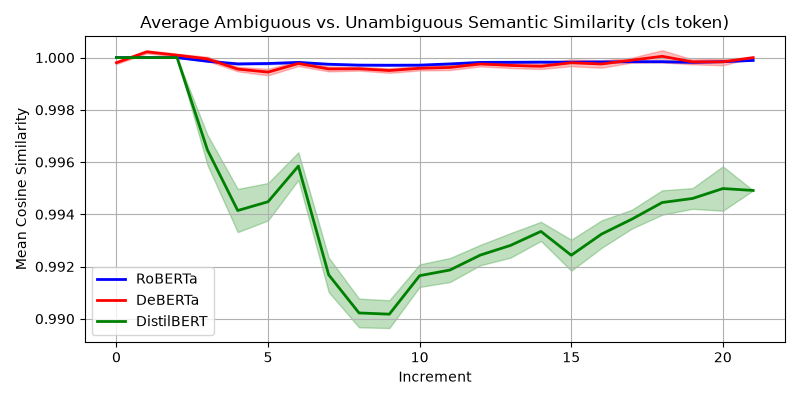}
    \includegraphics[width=0.4\linewidth]{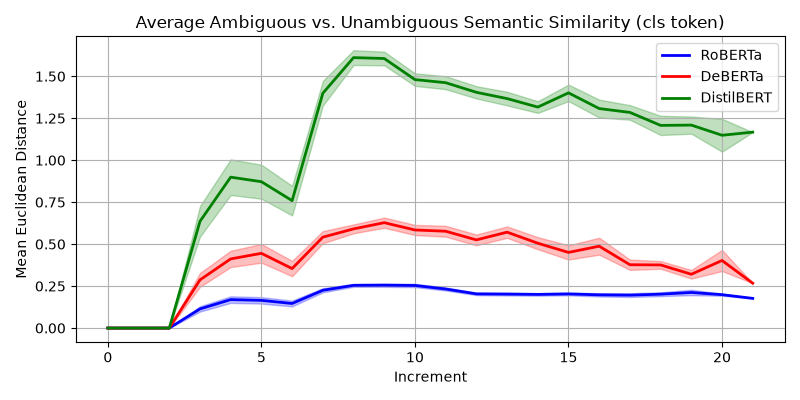}
    \includegraphics[width=0.4\linewidth]{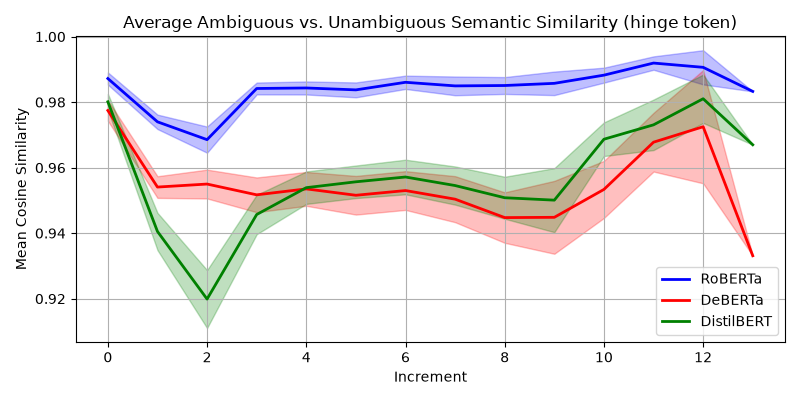}
    \includegraphics[width=0.4\linewidth]{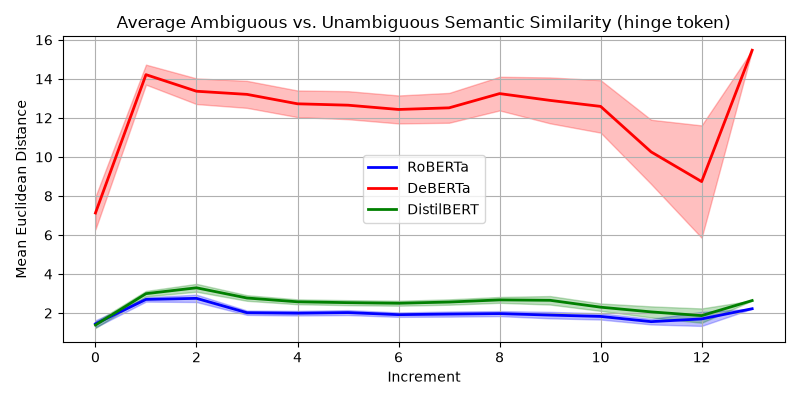}
    \caption{Summary statistics of token embedding distance between model interpretation of garden path and disambiguated sentences across RoBERTa, DeBERTa, and DistilBERT. $y$-axis on the left 2 plots is the cosine similarity between garden path sentences and disambiguated equivalents, $y$-axis on the right two plots is Euclidean distance between garden path sentences and disambiguated equivalents. Upper 2 plots are showing the average behavior of the CLS token and lower 2 plots are showing the average behavior of the hinge token.}
    \label{fig:summarystatsdebertaandroberta}
\end{figure}

We examine the semantic distance and semantic similarity as measured by Euclidean distance and cosine similarity across RoBERTa, DeBERTa, and DistilBERT models. Results are shown in \ref{fig:summarystatsdebertaandroberta}. The similar curves exhibited across the plots could suggest that the different model architectures exhibit similar characteristics when processing incremental garden path sentences despite the different attention mechanisms and weight quantization present in each of the models \cite{he2021debertadecodingenhancedbertdisentangled, liu_roberta_2019, sanh2020distilbertdistilledversionbert}.

We perform similar comparisons between RoBERTa and OLMo models. The trajectories do not align as well with the other models, but we do see the canonical dip and recovery of semantic similarity, although at different points in the sentence. Ultimately, we will need to perform more thorough analysis to come to a better understanding of alignment across architectures, but at this point we believe the discrepancy is due to the vastly larger embedding space of OLMo, as well as the different attention mechanism. 

\begin{figure}[h]
    \centering
    \includegraphics[width=0.4\linewidth]{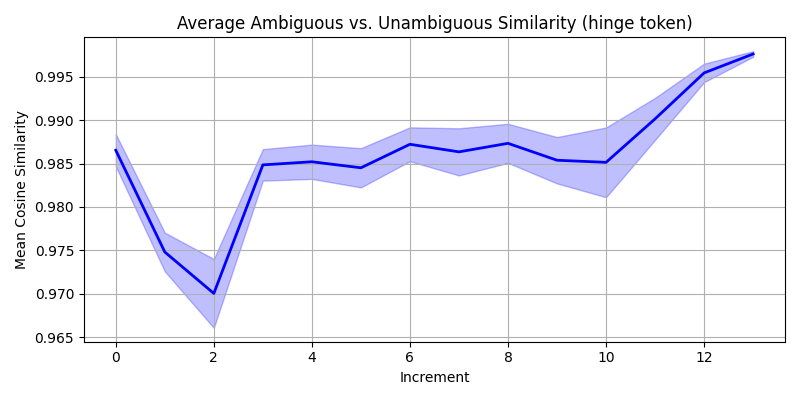}
    \includegraphics[width=0.4\linewidth]{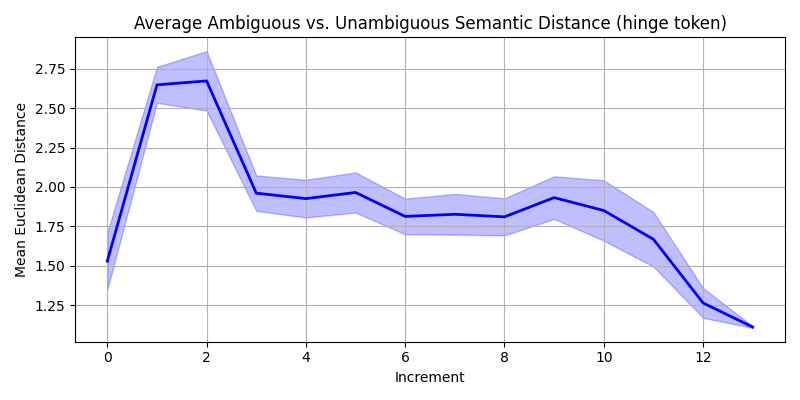}
    \includegraphics[width=0.4\linewidth]{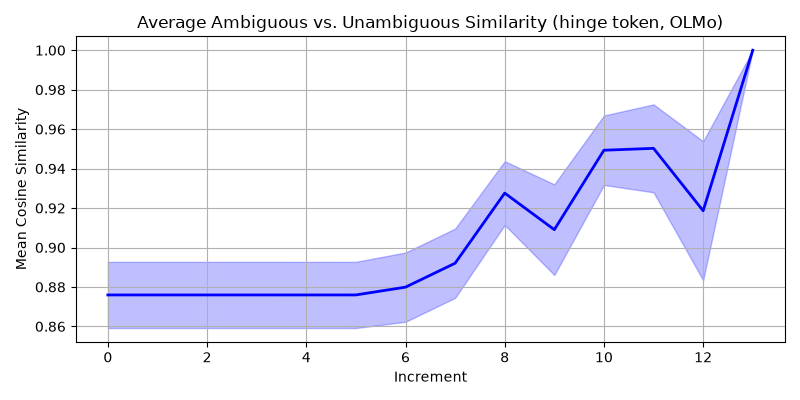}
    \includegraphics[width=0.4\linewidth]{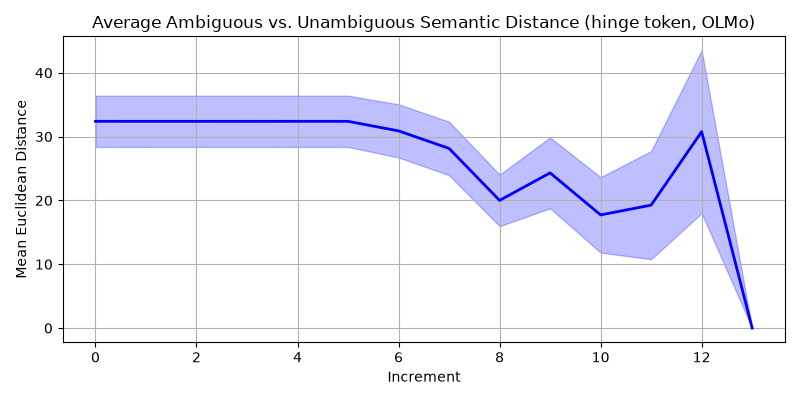}
    \caption{From left to right: Average cosine similarity between the hinge token in garden path and disambiguated sentences as additional tokens are added to context, Average Euclidean distance between the hinge token in garden path and disambiguated sentences as additional tokens are added to context. Upper 2 are RoBERTa, lower 2 are OLMo.}
    \label{fig:roberta_distances}
\end{figure}

\section{Classifier Model Specifications}
\label{appendix:classifier_model_specifications}

The classifier that we are using for garden path sentence classification is a 1 dimensional convolutional neural network. We use 3 convolution blocks with filters of size (in order from input first to last block) 5, 3, and 1. Each block uses a stride of 1 and a padding of 1.

After feature extraction is performed by the convolutional blocks, the features are fed into a feed forward neural network to perform the classification task. Further details as well as the complete implementation can be found on \href{https://github.com/graysonstorer/contextual-attractors/}{github}.

\section{Word Sense Disambiguation on Polysemous Tokens}
\label{appendix:WSD_on_polysemous_tokens}

\subsection{Methods}

We hypothesized that the incremental trajectory of polysemous tokens through contextual embedding space has a non trivial relationship with the sense of the token, such that a deep learning model can be trained to predict word sense given contextual trajectory. To test this hypothesis, we constructed a 1-dimensional convolutional neural network for prediction of word sense given a truncated list of difference vectors between token embeddings given incremental context. Sentences were categorized according to a $k$-means clustering algorithm, which grouped the sentences into 3 clusters such that the distance between each sentence's CLS token in the embedding space is minimized \cite{devlin_bert_2019}. A plot of what this looks like in PCA space is shown in figure 2. The categories roughly correspond to the prototypical uses of the word ``run". Cluster $0$ corresponds to the procedural use e.g. ``run the program", cluster 1 corresponds to the physical activity e.g. ``run on the treadmill", and cluster 2 corresponds to the political action e.g. ``run for office". After each sentence in the dataset was assigned a cluster, we calculated the incremental trajectory of the target token with algorithm 1. These trajectory data were used as input for a convolutional neural network, which was trained to fit trajectory data to the corresponding cluster as determined by the $k$-means clustering algorithm.

It is worth noting that the silhouette scores for all of the different values of $k$ that we tested were in the range of $\sim 0.19$ to $\sim 0.23$. These scores are not indicative of a perfect match; however, this is an expected result due to the nonlinearity of decision boundaries for regions of semantic similarity within contextual embedding space \cite{wiedemann2019doesbertmakesense}. To ensure accurate classification, the clusters identified by the $k$-means algorithm were checked manually to ensure that the machine classification reflected semantic similarity.

\subsection{Results}

To test the relationship between polysemous token sense and contextual trajectory, we trained a 1-dimensional convolutional neural network to recognize word sense given a list of trajectory vectors. Over an average of 100 different random model instantiations, we were able to reach an average accuracy of $\sim 85.38\%$ on the validation dataset using an $80 \% -20\%$ train-test split.  

Figure 2 shows the clusters in PCA space as identified by a $k$-means clustering algorithm for $k = 3$. Each cluster roughly corresponds to the prototypical uses of the word ``run", with cluster $0$ representing the procedural use e.g. ``run the program", cluster 1 representing the physical activity e.g. ``run on the treadmill", and cluster 2 representing the political action e.g. ``run for office". The clustering was performed without any PCA as this process would be a significant enough reduction ($\mathbb{R}^{768} \rightarrow \mathbb{R}^2$) such that the clusters may not actually represent true nearest neighbors \cite{wiedemann2019doesbertmakesense} \cite{cai_isotropy_2021}. 

\begin{figure}[h]
    \centering
    \includegraphics[width=0.5\linewidth]{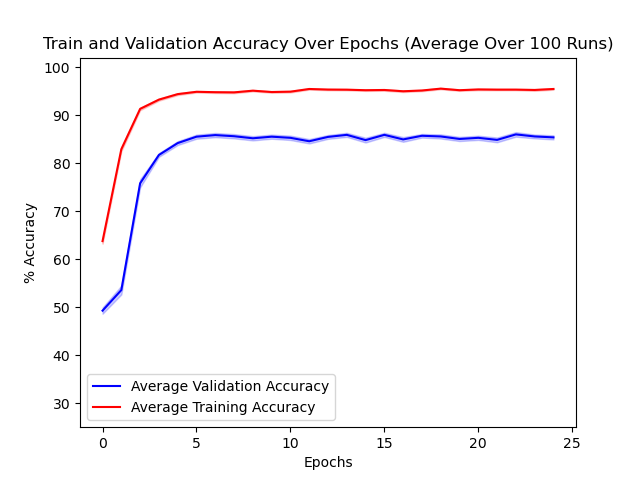}
    \caption{Training and Validation accuracy curves for the word sense classifier}
    \label{fig:placeholder}
\end{figure}


As shown in figure 3, each of the 100 randomly instantiated models was trained for 25 epochs. The training accuracy and validation accuracy curves show clear overfitting on the training data, but the model is still able to generalize to a non-trivial accuracy on the validation dataset.




\section{Testing Different Increment Sizes}
\label{appendix:different_increment_sizes}
\subsection{Methods}
We propose several different implementations of ``contextual increment''. There are several plausible definitions, so we compare X, Y, Z on the test X (probably seeing which works better, classification-wise, maybe something like which has the most variance across representations?). From that, we conclude that the most promising initial candidate is X. Our current working definition of contextual increment is as follows: Given a sentence $T$ (which we will treat as a set of words), an embedding function $f : T \rightarrow \mathbb{R}^{768}$, and sentence length $L = |T|$. We instantiate $X$ as an empty list of lists. $\forall i \in L,$ append the embedding $f(\{T_0, \dots, T_i\})$ to $X$ where $f(\{T_0, \dots T_i \})$ is a point in $\mathbb{R}^{768}$ corresponding to the CLS token of the sentence $\{T_0, \dots, T_i \}$. We analyze the effects of different embedding functions such as embedding only the first token in the sentence or a previously specified target token. The various embedding functions are as follows:
\begin{itemize}
    \item $f_1: (T_0, \dots, T_I) \rightarrow \mathbb{R}^{768}$. This function will contain $I - 1$ difference vectors where $I = |T|$ and each difference vector is taken as an additional word is added to the context. 
    \item $f_2 : (T_0, T_I) \rightarrow \mathbb{R}^{768}$. This function will calculate the trajectory as a difference vector between the initial embedding and the final embedding with full context.
    \item $f_3: (T_0, T_{\lfloor{\frac{I}{2}}\rfloor,}, T_I) \rightarrow \mathbb{R}^{768}$. This function will calculate the trajectory as 2 difference vectors: the first vector will be the difference from the initial embedding and the recomputed embedding with given half of the words $ \in T$, the second vector will be the difference between the embedding given half of the words $\in T$ and the embedding given all of $T$.
    
\end{itemize}
We test the resilience of various representations of incremental context to small perturbations. We do so by training a classifier model which matches the trajectory as represented by different incrementing functions $(f_1, f_2 \text{ or } f_3)$ to the type of sentence corresponding to the sentence from which the trajectory was derived. 

\subsection{Results}
\begin{figure}[h]
    \centering
    \includegraphics[width=0.3\linewidth]{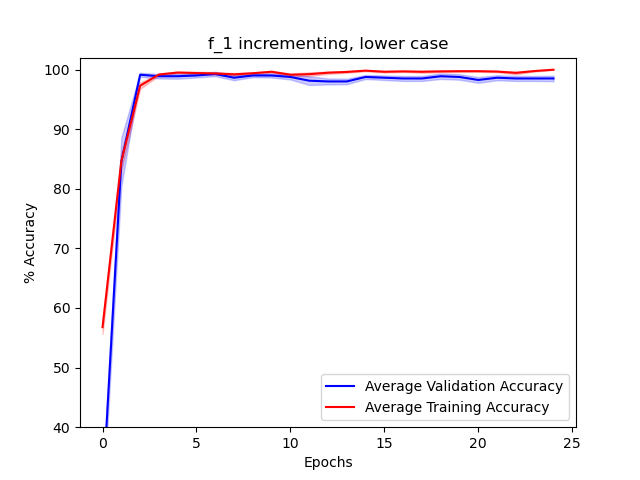}
    \includegraphics[width=0.3\linewidth]{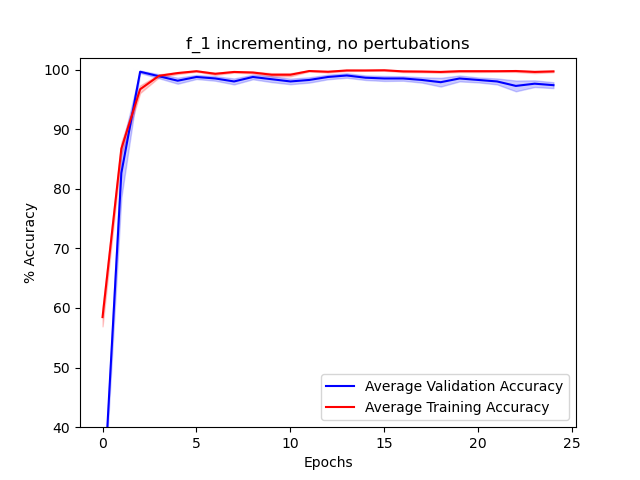}
    \includegraphics[width=0.3\linewidth]{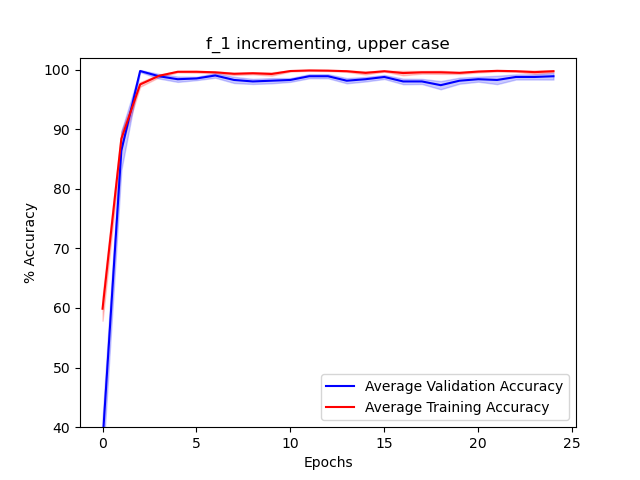}
    \includegraphics[width=0.3\linewidth]{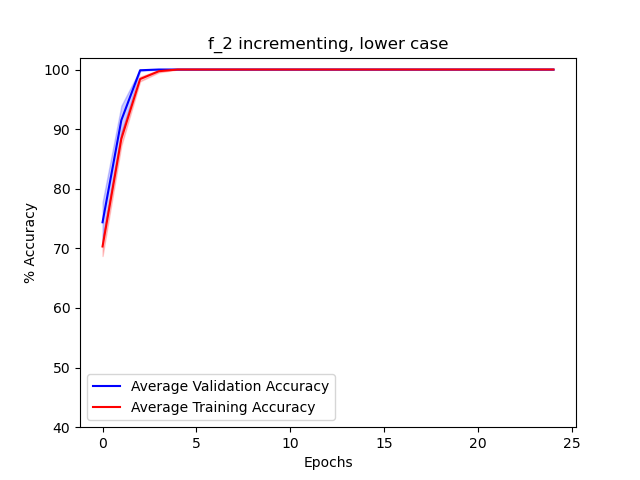}
    \includegraphics[width=0.3\linewidth]{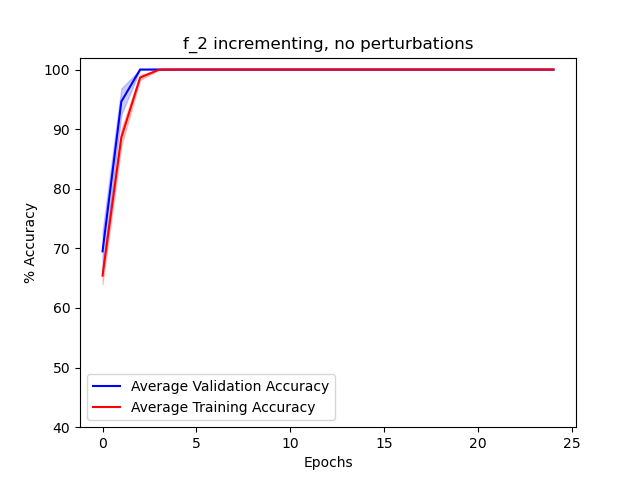}
    \includegraphics[width=0.3\linewidth]{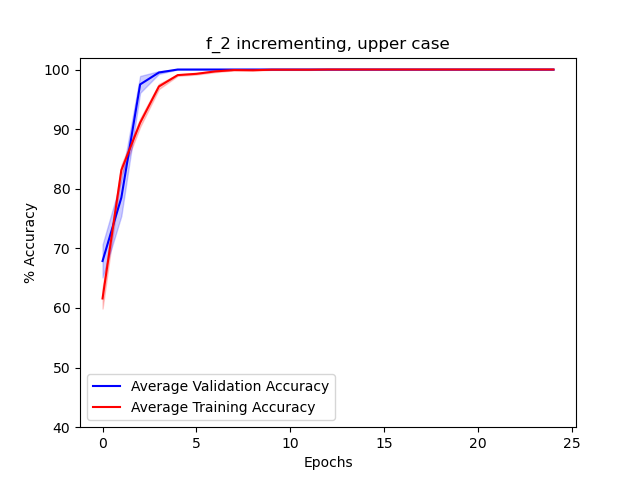}
    \includegraphics[width=0.3\linewidth]{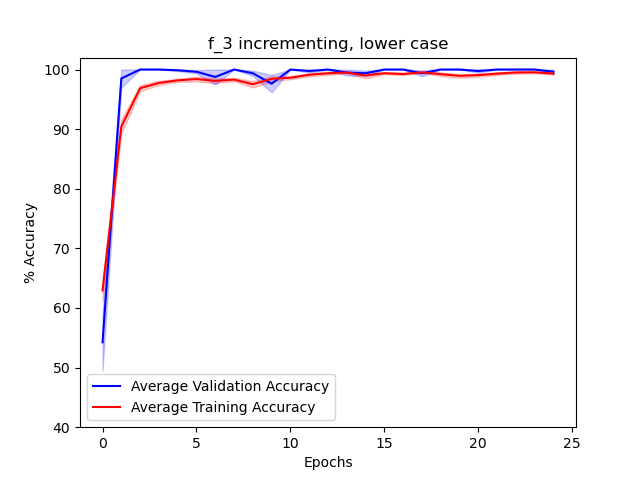}
    \includegraphics[width=0.3\linewidth]{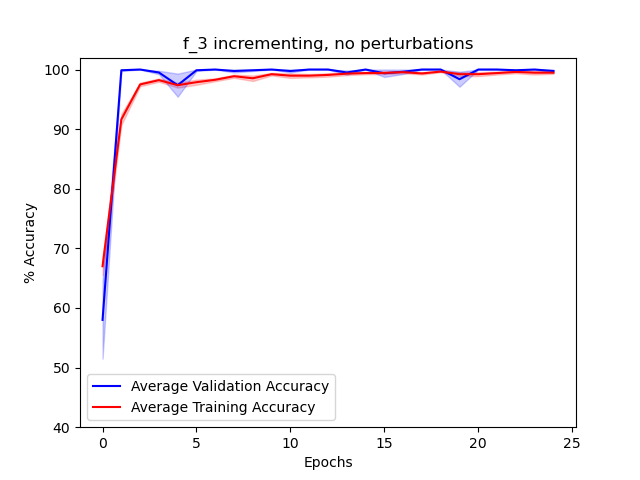}
    \includegraphics[width=0.3\linewidth]{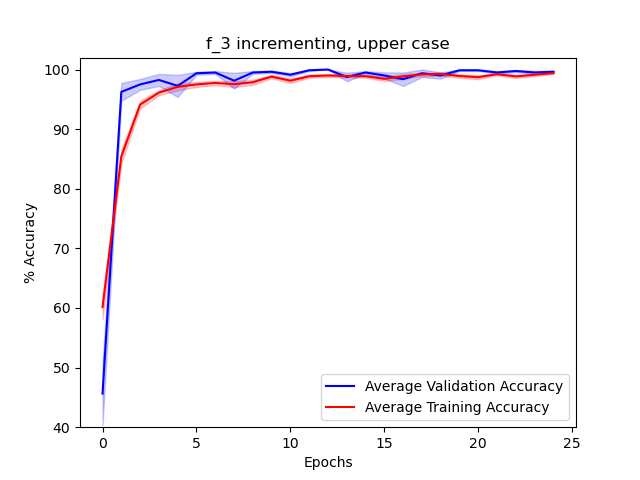}
    \caption{Classifier performance on increments calculated by different functions (from left to right) no perturbations, all characters lower case, and all characters upper case.}
    \label{fig:placeholder}
\end{figure}

\noindent The $f_2$ incrementing function shows the highest resistance to perturbation as training and validation accuracy curves maintain a consistent shape across all datasets no matter the perturbation. The $f_1$ and $f_3$ incrementing functions show similar resistance to perturbation, though they are less resistant to perturbation than $f_2$. The $f_2$ incrementing function has the best resistance to capitalization related perturbation, but all of the incrementing functions exhibit similarly high resistance. The performance of each incrementing function is most likely a result of the amount of data generated by each function; $f_1$ and $f_3$ yield more data for per sentence, which results in more points to reflect perturbation.

We test resilience to punctuation related perturbations by replacing all punctuation that could reveal the sentence category with a period; i.e. question marks and exclamation points would become periods. 

\begin{figure}[h]
    \centering
    \includegraphics[width=0.3\linewidth]{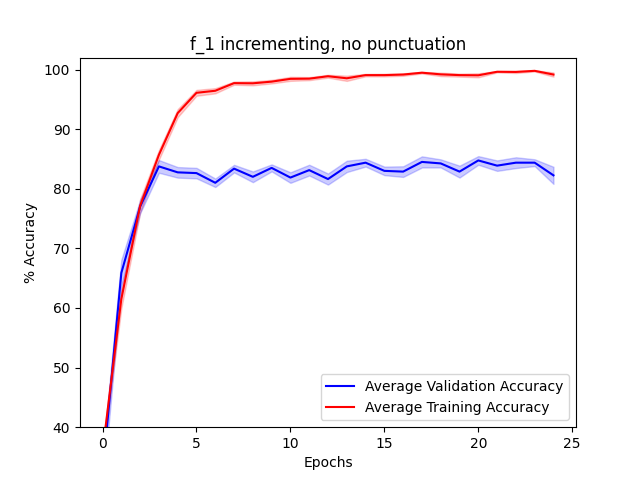}
    \includegraphics[width=0.3\linewidth]{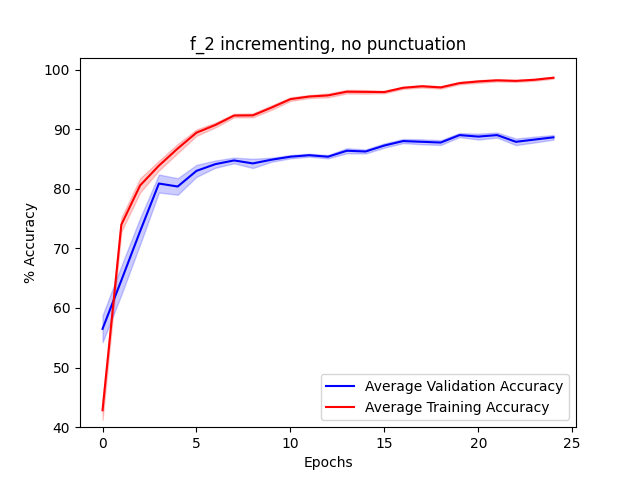}
    \includegraphics[width=0.3\linewidth]{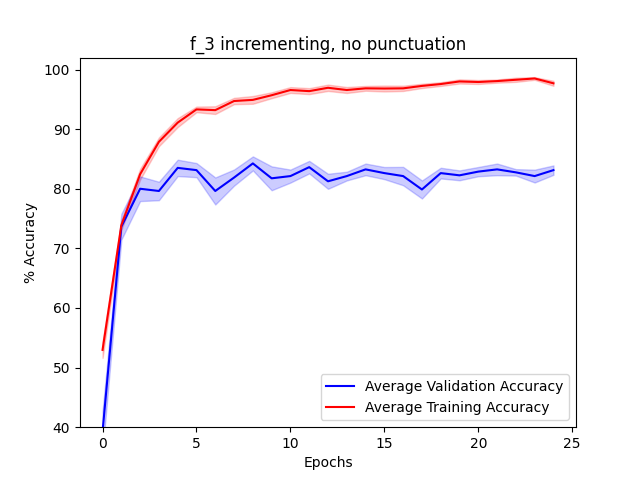}
    \caption{Classifier performance on increments as represented by $f_1$ with (from left to right) no perturbations, all characters lower case, and all characters upper case.}
    \label{fig:placeholder}
\end{figure}

\noindent The training and validation accuracy curves for all of the different representations of contextual increments show that the models rely heavily on the punctuation associated with each different sentence type. Despite perturbation, all of the models are still able to retain performance on the classification task of at least $80 \%$ accuracy on the validation dataset. The $f_2$ incrementing function had the highest resistance to punctuation related perturbations, which we believe is due to the same reasons for the high performance of $f_2$ in the experiment for RQ1. Furthermore, the $f_2$ incrementing function returns one vector which is a more accurate representation of the sentence's location in latent space, as opposed to contextual trajectory through latent space. The $f_1$ incrementing function exhibited the most overfitting of all the functions, while the $f_3$ incrementing function experienced the most variance across the random model instantiations, as shown by the larger error along validation accuracy curves. 

\section{Detailed Limitations}
\label{appendix:detailed_limitations}


In \ref{sec:JokesRQ}, we examine how misdirection in linguistic phenomena other than garden path sentences is processes by LLMs. We answer this RQ by conducting an analysis of latent space representation of ``dad jokes" as compiled by Silvia and Burnett \cite{silvia_burnett_2025}. These jokes do contain elements of misdirection, but the analysis performed is not indicative of misdirection appearing in all varieties of jokes as ``dad jokes" by nature rely on misdirection or some form of play on words that is not endemic to other forms of jokes. Instead this should serve as a study of how sentence scale misdirection manifests itself in latent space.  

The bulk of our results are derived from analysis of embeddings generated by the RoBERTa model.  These embeddings are located in $\mathbb{R}^{768}$. The dimensions of this embedding space make tasks like visualization incredibly complex. In many cases, the first 2 principle components of an embedding will only account for $\sim 30 \%$ of the variance exhibited by the embedding. However, this problem is exacerbated in more recent model releases. For example, the OLMo model uses $\mathbb{R}^{4096}$ as its embedding space as well as a more complicated attention mechanism. 

The experiments conducted in this paper are intended to highlight the ability to glean meaningful linguistic information by examining token level trajectory through contextual embedding space. Therefore the experiments primarily analyze linguistic phenomena from the perspective of token level structure, at times omitting more global syntactic information. We present these experiments with the goal of illustrating that token level syntactic structure can also reveal the presence of phenomena that are traditionally thought of as occurring at a broader, more global scale such as humor, ambiguity, and other linguistic phenomena that occur at a scale of more than 1 word.  

Incremental token-wise analysis comes with a non-trivial computational cost: For an $N$ length utterance, generating a trajectory of the CLS token of this utterance would require computing $\mathcal{O}(N)$ embeddings.

\section{Additional Figures}
\label{appendix:additional_figures}
\begin{figure}[h]
    \centering
    \includegraphics[width=0.3\linewidth]{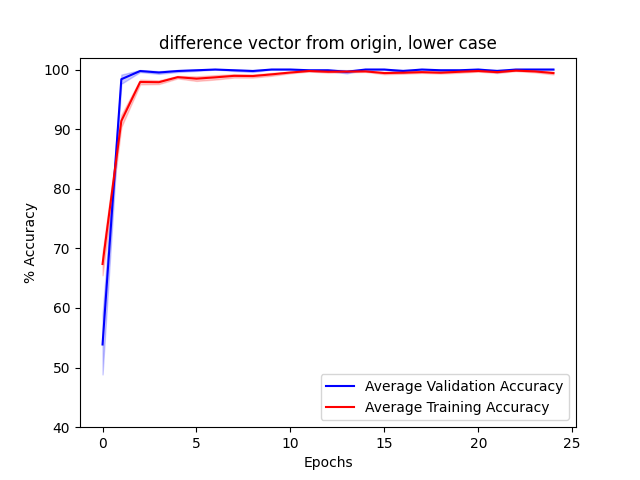}
    \includegraphics[width=0.3\linewidth]{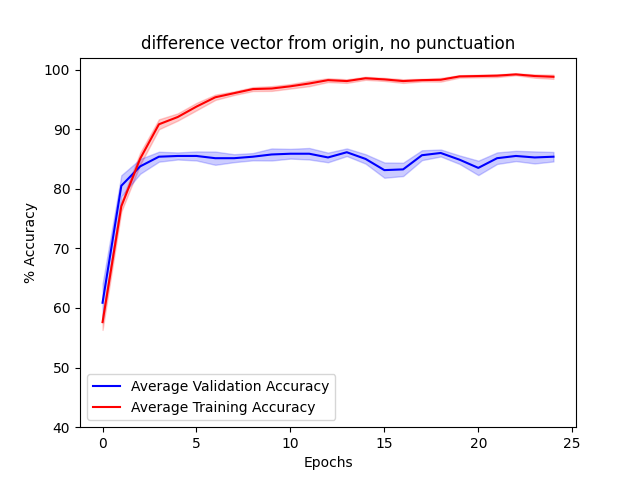}
    \includegraphics[width=0.3\linewidth]{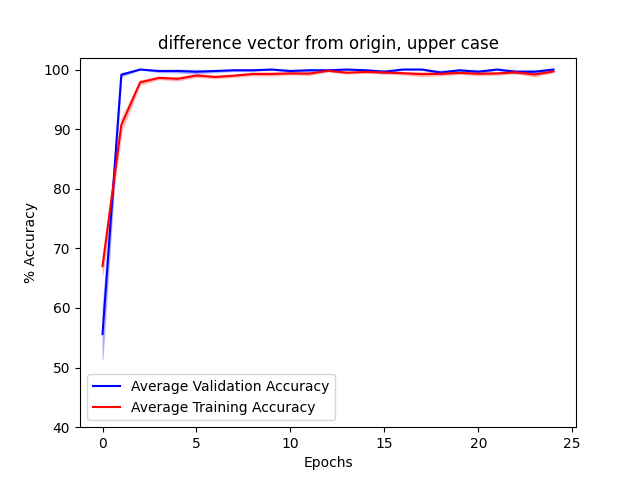}
    \caption{Classifier performance on data generated by $D_4$ with various perturbations including all characters lowercase, all characters upper case, and no punctuation}
    \label{fig:representationtesting_perturbation_D_4}
\end{figure}
\begin{figure}
    \centering
    \includegraphics[width=0.3\linewidth]{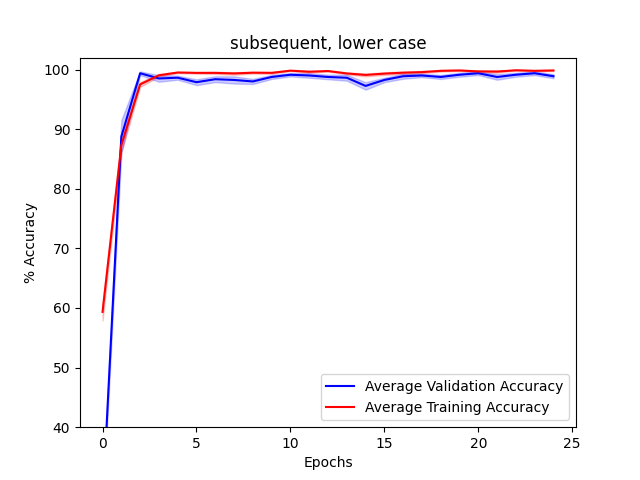}
    \includegraphics[width=0.3\linewidth]{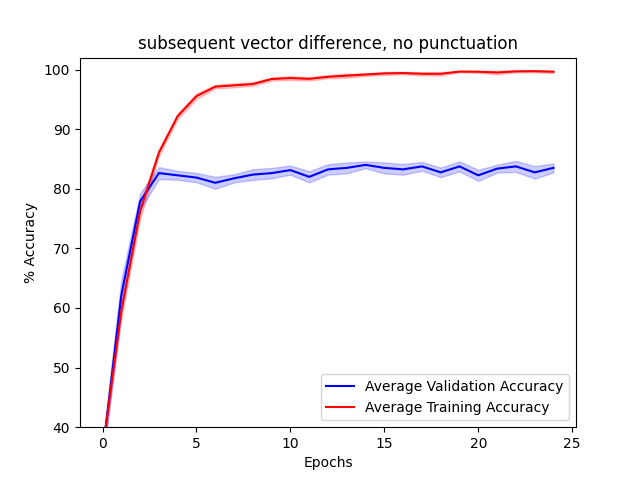}
    \includegraphics[width=0.3\linewidth]{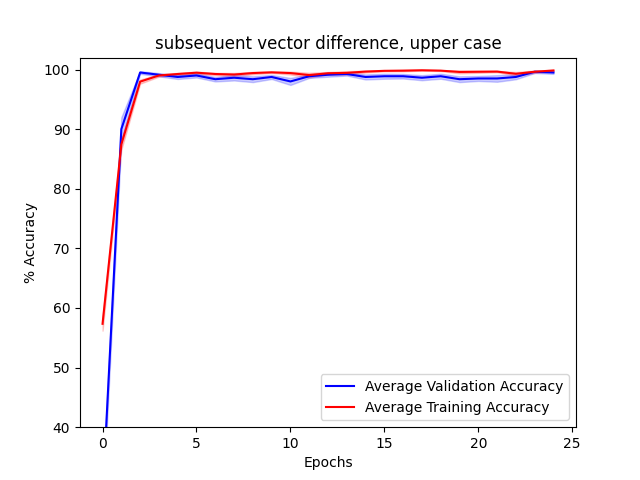}
    \caption{Classifier performance on data generated by $D_2$ with various perturbations including all characters lowercase, all characters upper case, and no punctuation}
    \label{fig:representationtesting_perturbation_D_2}
\end{figure}
\begin{figure}
    \centering
    \includegraphics[width=0.3\linewidth]{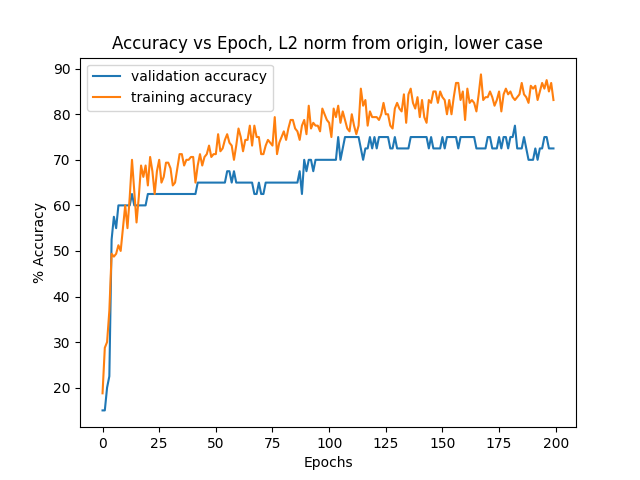}
    \includegraphics[width=0.3\linewidth]{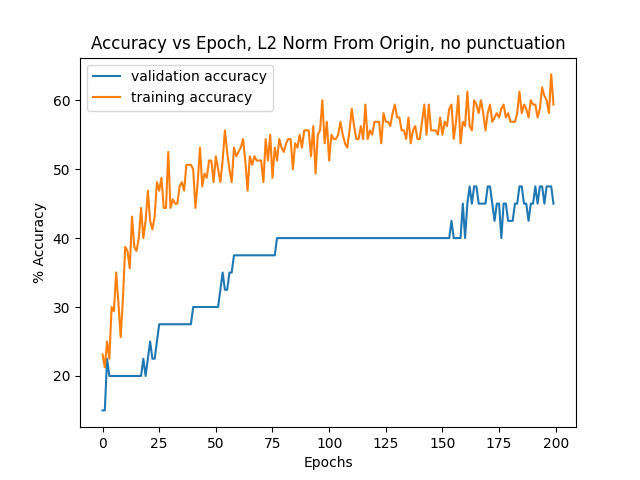}
    \includegraphics[width=0.3\linewidth]{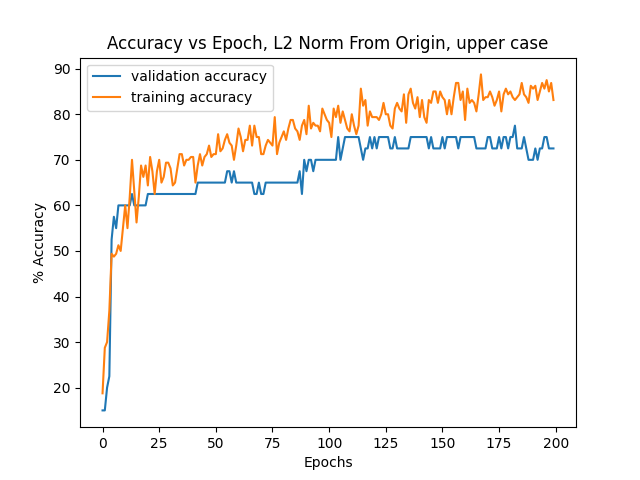}
    \caption{Classifier performance on data generated by $D_3$ with various perturbations including all characters lowercase, all characters upper case, and no punctuation}
    \label{fig:representationtesting_perturbation_D_3}
\end{figure}
\begin{figure}
    \centering
    \includegraphics[width=0.3\linewidth]{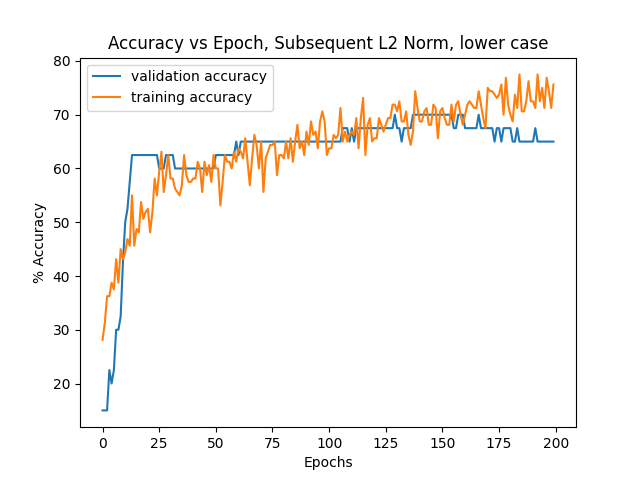}
    \includegraphics[width=0.3\linewidth]{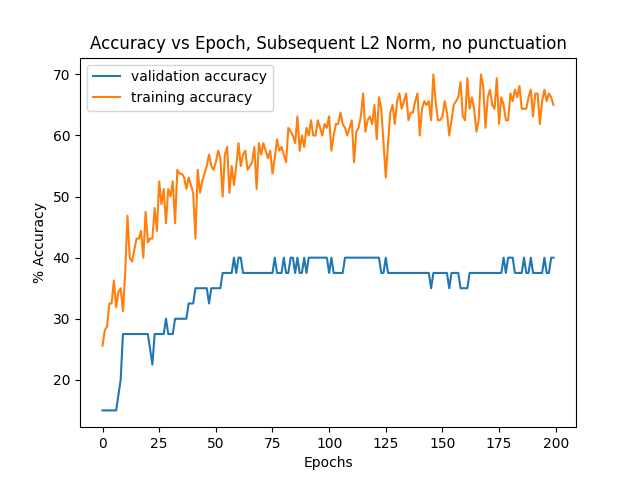}
    \includegraphics[width=0.3\linewidth]{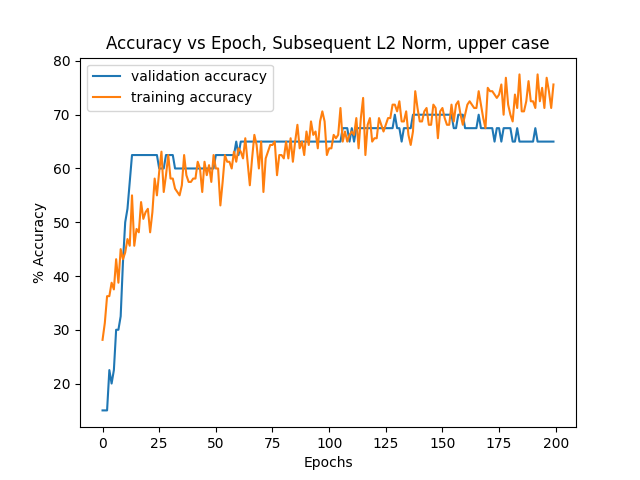}
    \caption{Classifier performance on data generated by $D_1$ with various perturbations including all characters lowercase, all characters upper case, and no punctuation}
    \label{fig:representationtesting_perturbation_D_1}
\end{figure}
\begin{figure}[h]
    \centering
    \includegraphics[width=0.5\linewidth]{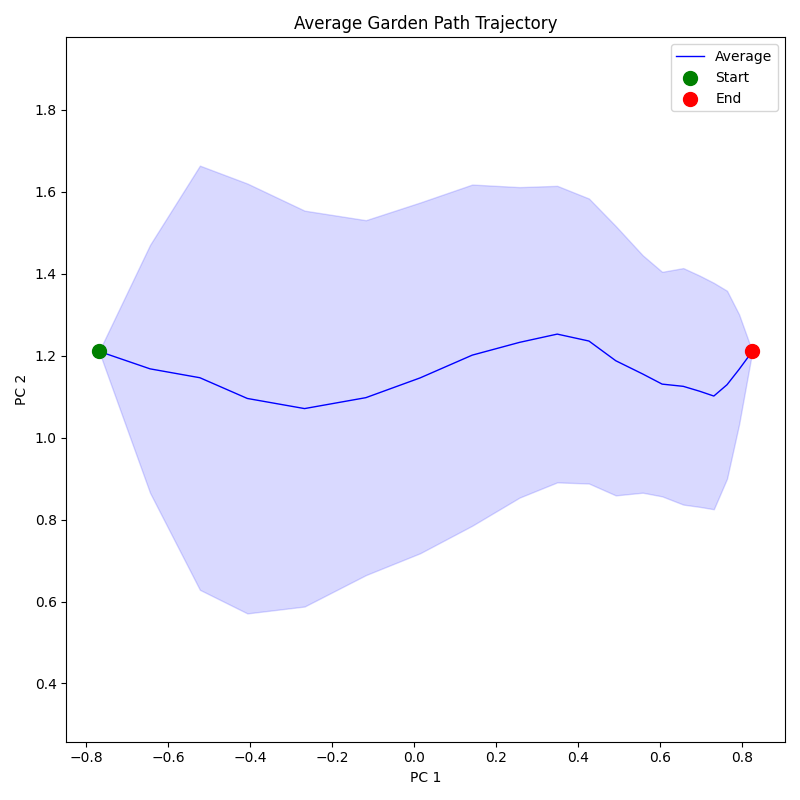}
    \caption{Average trajectory of a garden path sentence through embedding space.}
    \label{fig:average_garden_path_traj}
\end{figure}


\begin{figure}
    \centering
    \includegraphics[width=0.45\linewidth]{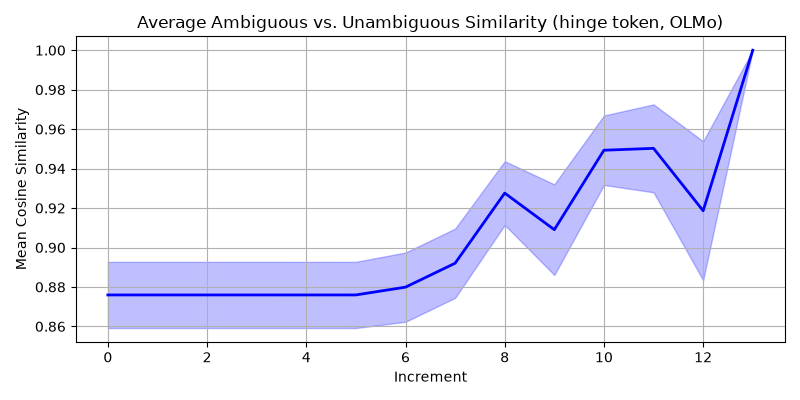}
    \includegraphics[width=0.45\linewidth]{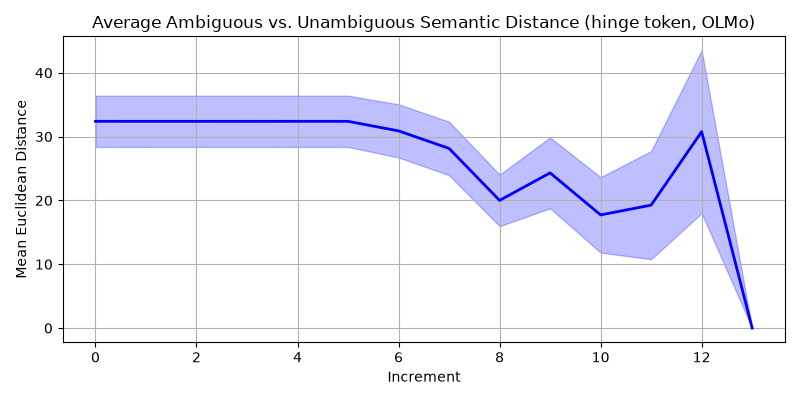}
    \caption{Average similarity statistics across all garden path sentences in the dataset when incrementally embedded into the OLMo embedding space. From left to right: Average cosine similarity between the hinge token in garden path and disambiguated sentences as additional tokens are added to context, Average Euclidean distance between the hinge token in garden path and disambiguated sentences as additional tokens are added to context.}
    \label{fig:olmo_distances}
\end{figure}

\begin{figure}
    \centering
    \includegraphics[width=0.9\linewidth]{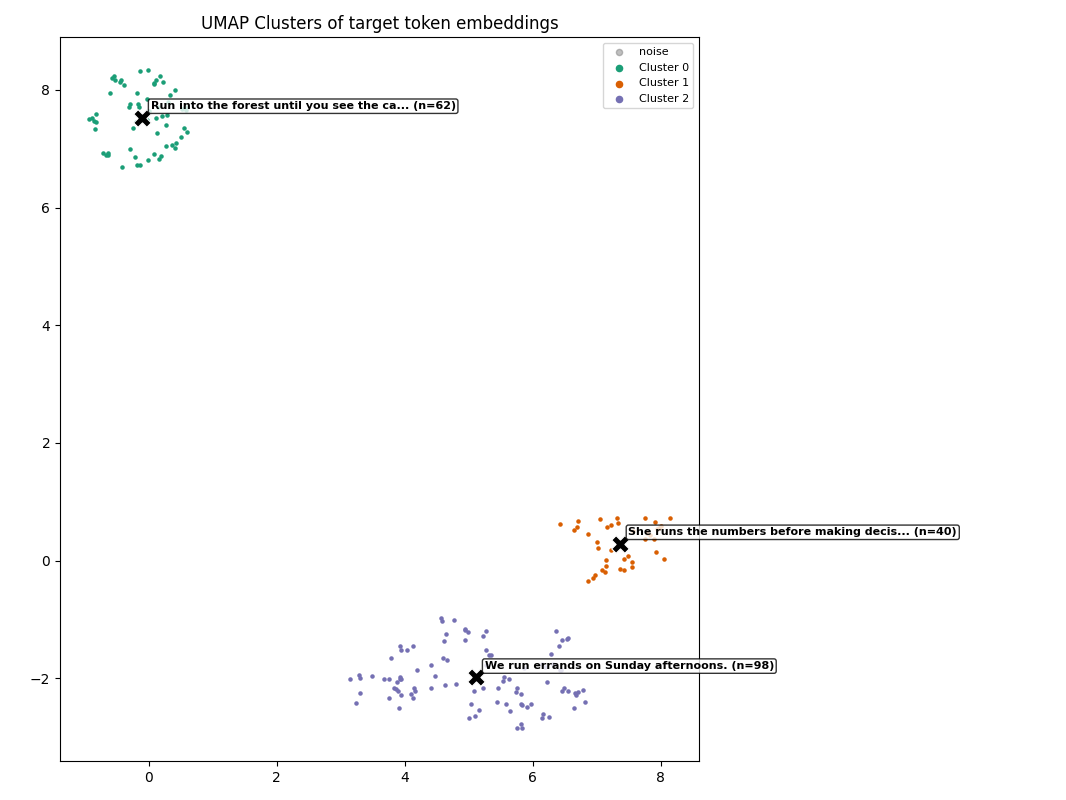}
    \caption{UMAP Clustering of CLS token embeddings of ``run'' sentences. All imperatives were grouped into cluster 0, all interrogatives were grouped into cluster 2, declaratives were split across clusters 1 and 2, and exclamatory sentences were spread across all 3 clusters.}
    \label{fig:umap_clusters}
\end{figure}

\begin{figure}
    \centering
    \includegraphics[width=0.9\linewidth]{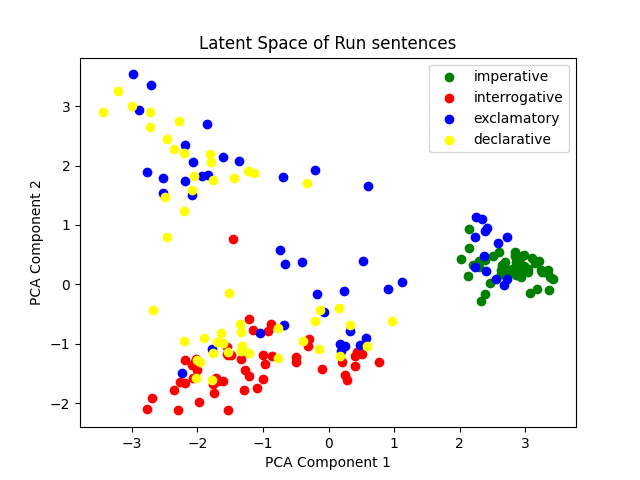}
    \caption{Principle Component Analysis of the latent space representation of the target token (``run'') of sentences, grouped by sentence type.}
    \label{fig:run_token_sentences}
\end{figure}

\begin{figure}
    \centering
    \includegraphics[width=0.45\linewidth]{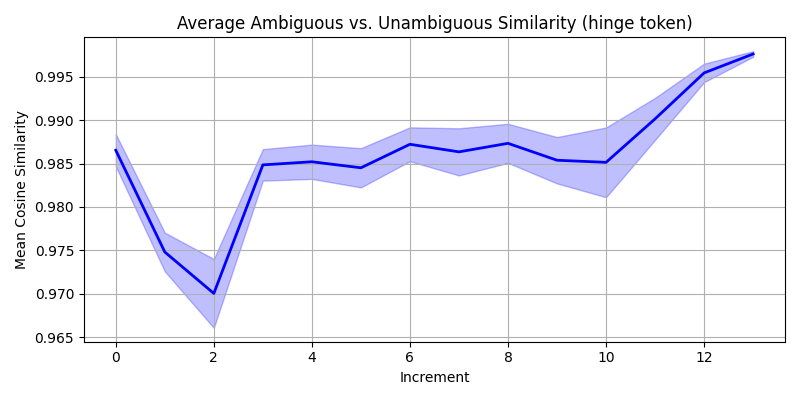}
    \includegraphics[width=0.45\linewidth]{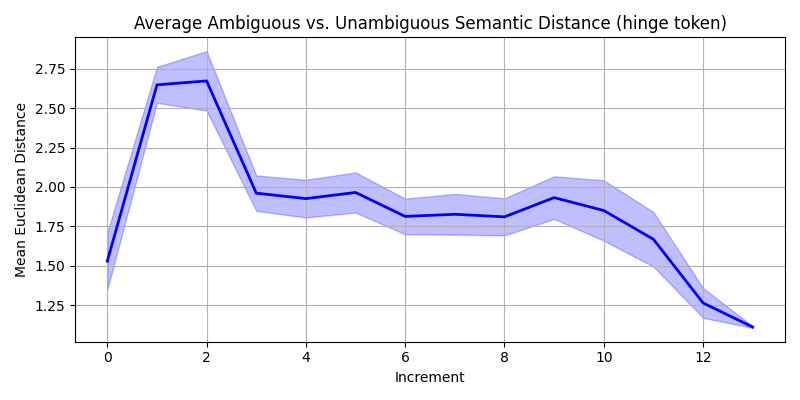}
    \caption{Semantic similarity metrics across the average CLS embedding trajectory of all garden path and negated garden path sentences. The plot on the left shows the cosine similarity between the two average trajectories and the plot of the right shows the Euclidean distance between the two average trajectories.}
    \label{fig:roberta_distances}
\end{figure}
\begin{figure}
    \centering
    \includegraphics[width=0.9\linewidth]{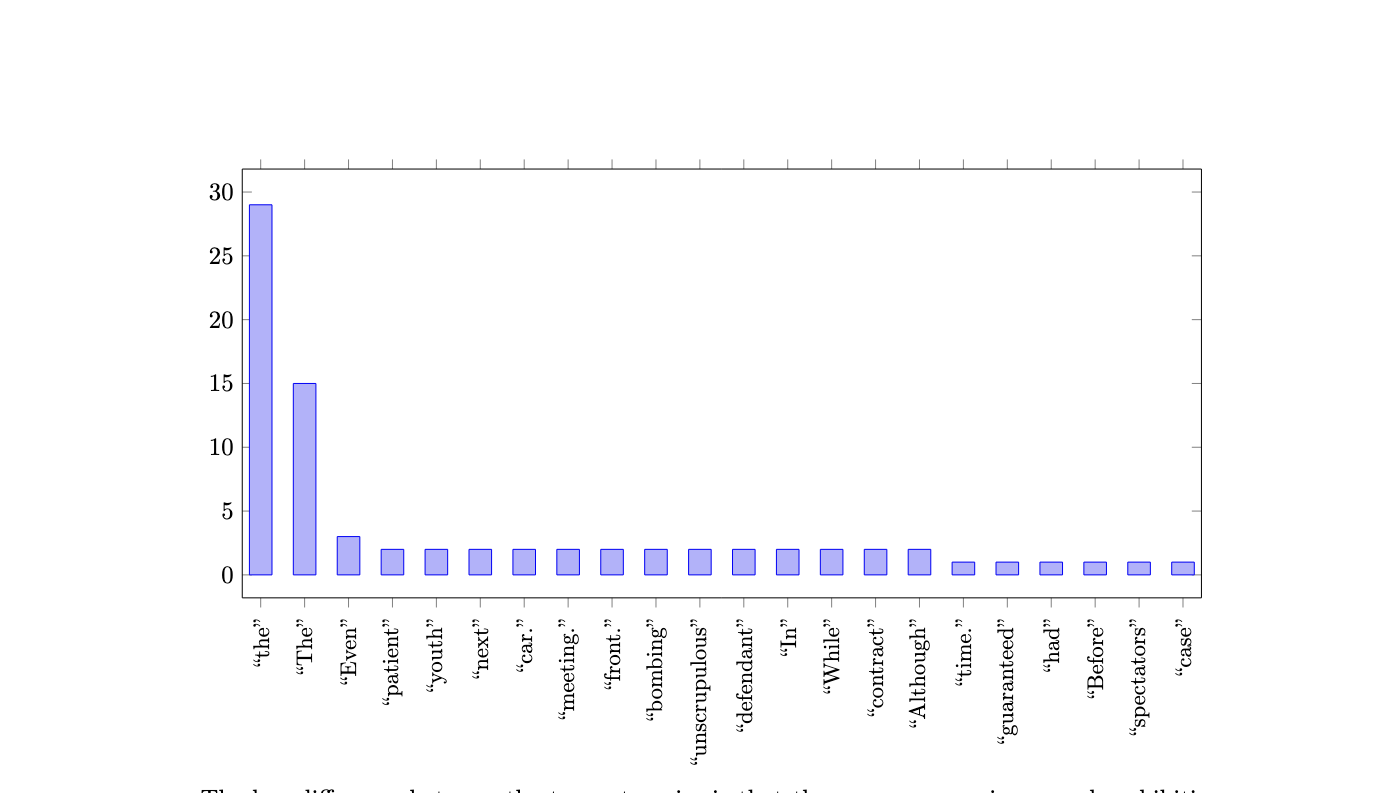}
    \caption{Plot depicting the number of times that each corresponding token is the most turbulent token in its respective sentence.}
    \label{fig:frequent_most_turbulent}
\end{figure}
\begin{figure}[h]
    \centering
    \includegraphics[width=0.8\linewidth]{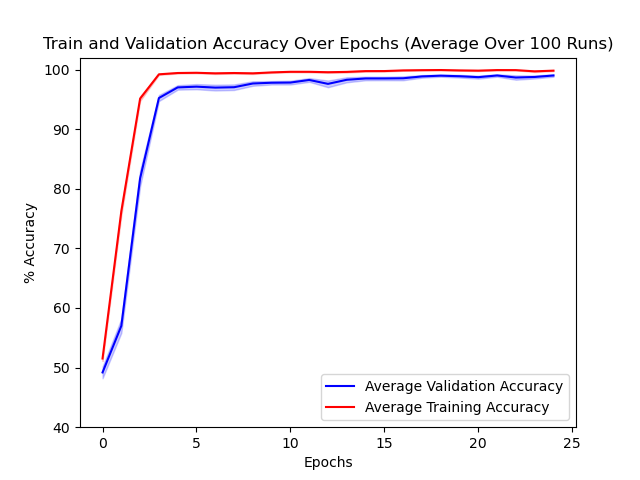}
    \caption{Training and Validation accuracy curves for the garden path variety classifier.}
    \label{fig:unshuffled}
\end{figure}
\begin{figure}
    \centering
    \includegraphics[width=0.9\linewidth]{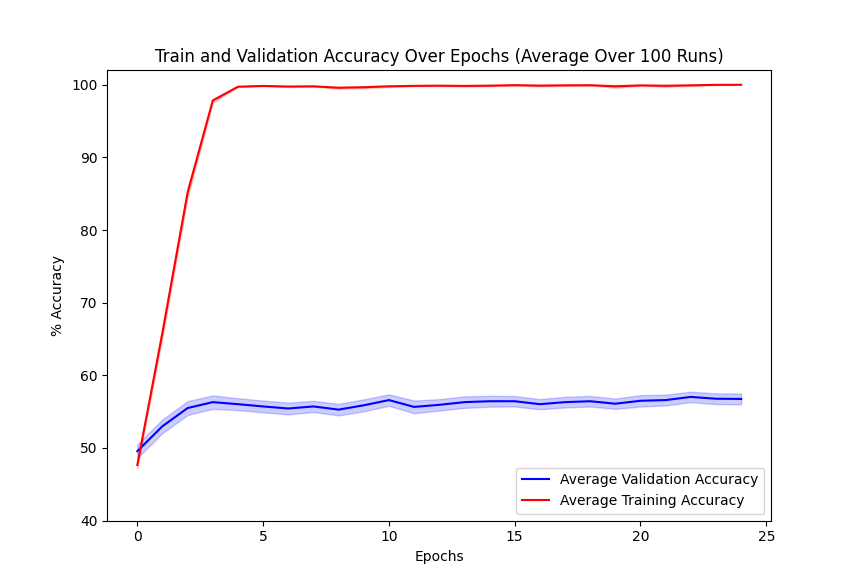}
    \caption{Training and validation accuracy curves of the classifier model trained on shuffled data.}
    \label{fig:shuffled}
\end{figure}
\begin{figure}[h]
    \centering
    \includegraphics[width=0.9\linewidth]{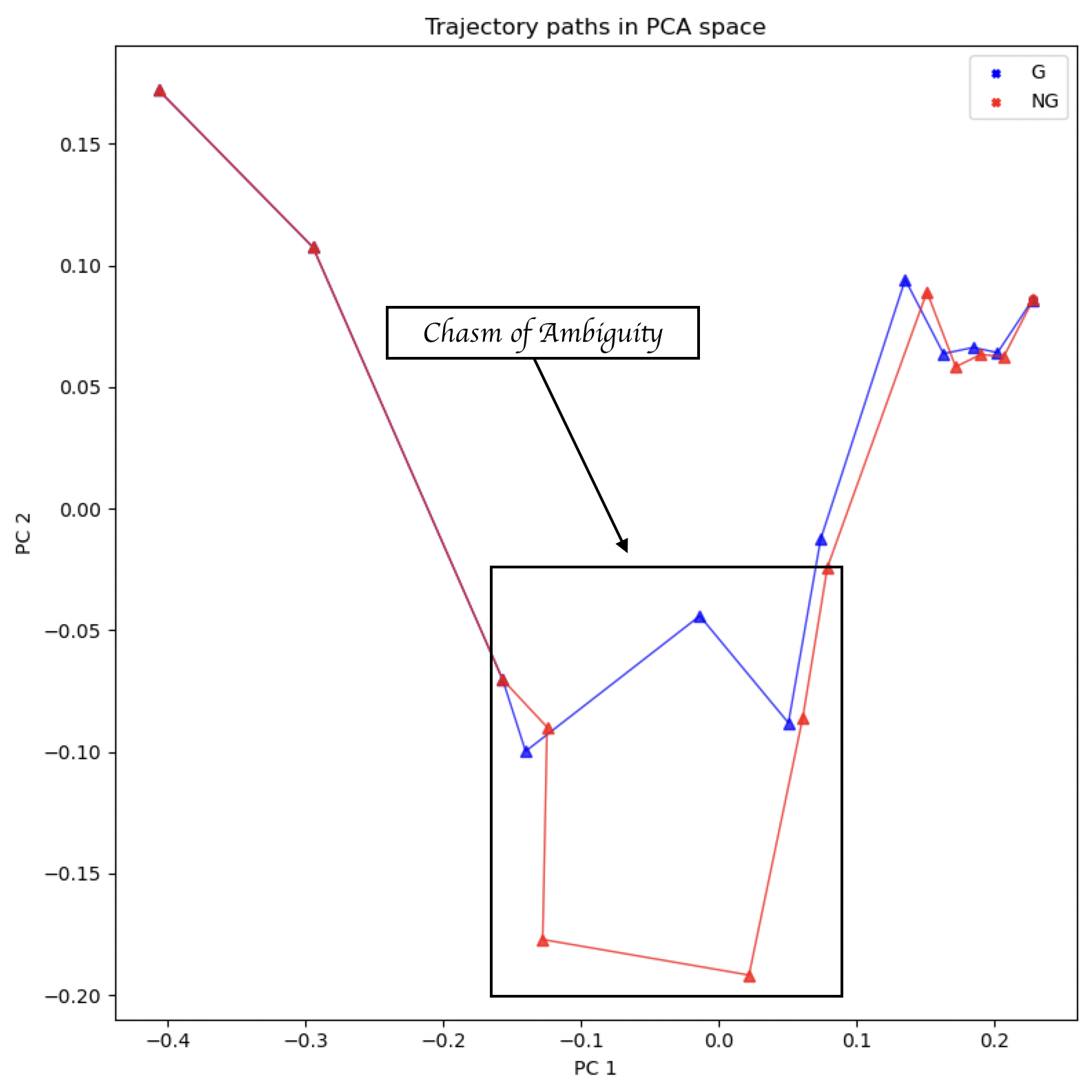}
    \caption{PCA plot showing the characteristic gap between parses of a garden path sentence and the unambiguous equivalent. }
    \label{fig:chasm_with_no_art}
\end{figure}
\begin{figure}[h]
    \centering
    \includegraphics[width=0.45\linewidth]{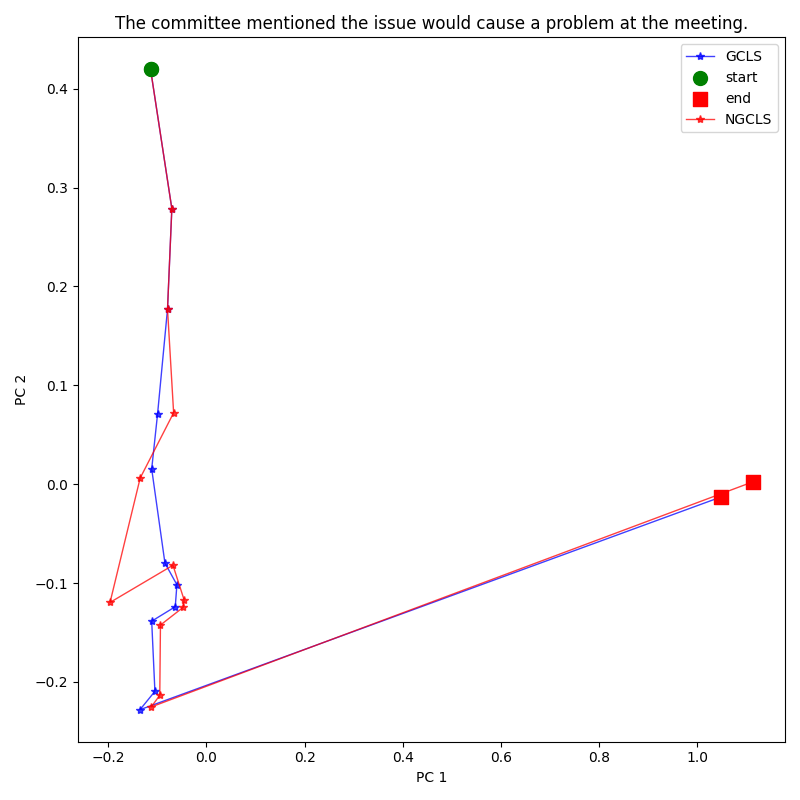}
    \includegraphics[width=0.45\linewidth]{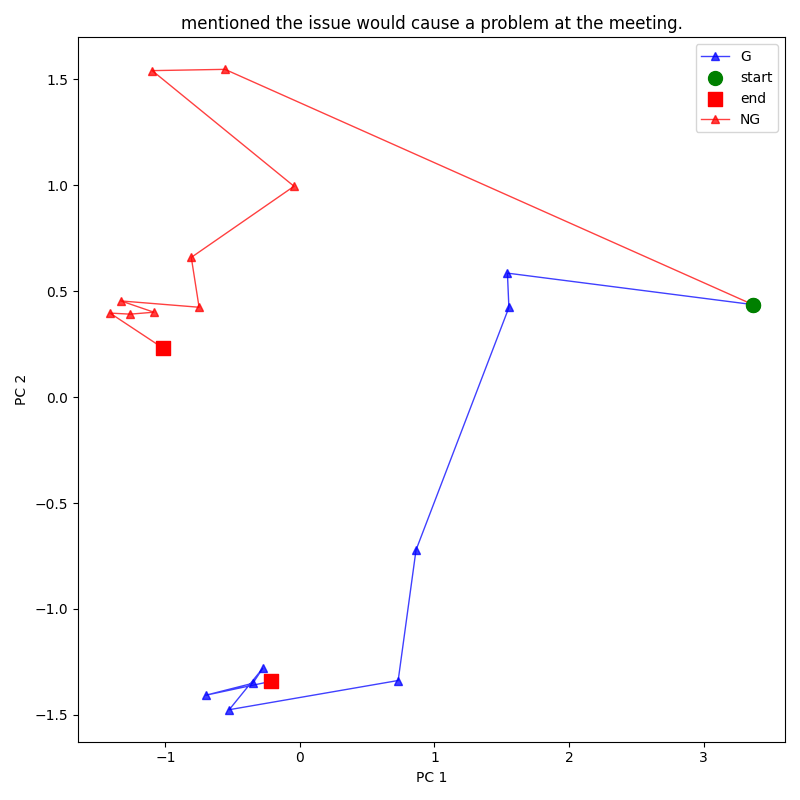}
    \caption{PCA plot CLS token trajectory and ambiguous token trajectory. The left plot shows the CLS token trajectories and the right plot shows the ambiguous token trajectories.}
    \label{fig:placeholder}
\end{figure}
\begin{figure}
    \centering
    \includegraphics[width=0.45\linewidth]{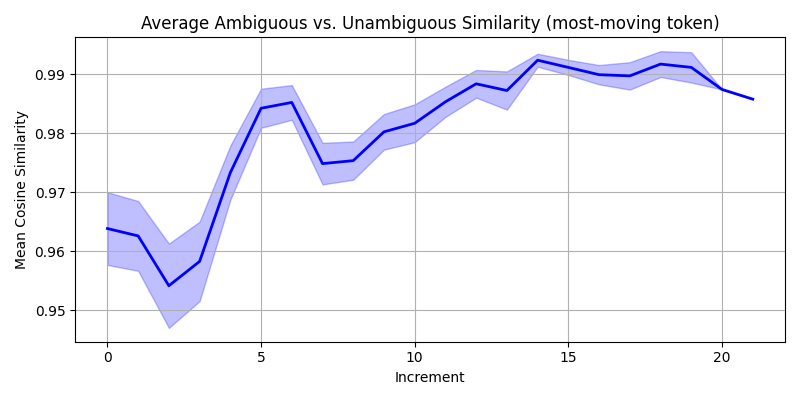}
    \includegraphics[width=0.45\linewidth]{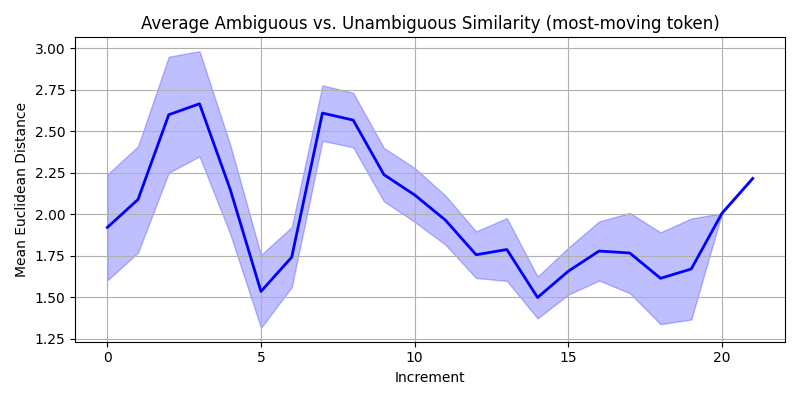}
    \label{fig:most_moving_token}
    \caption{Average cosine similarity and euclidean distance of the token that exhibits the most displacement over the trajectory of garden path sentences.}
\end{figure}
\begin{figure}[h]
    \centering
    \includegraphics[width=0.9\linewidth]{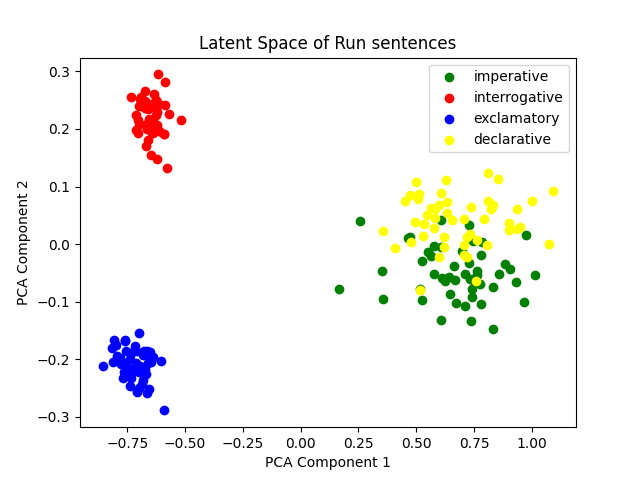}
    \caption{Principle Component Analysis of the latent space representation of the CLS token of sentences with the word run, grouped by sentence type.}
    \label{fig:run_latent}
\end{figure}
\begin{figure}[h]
    \centering
    \includegraphics[width=0.45\linewidth]{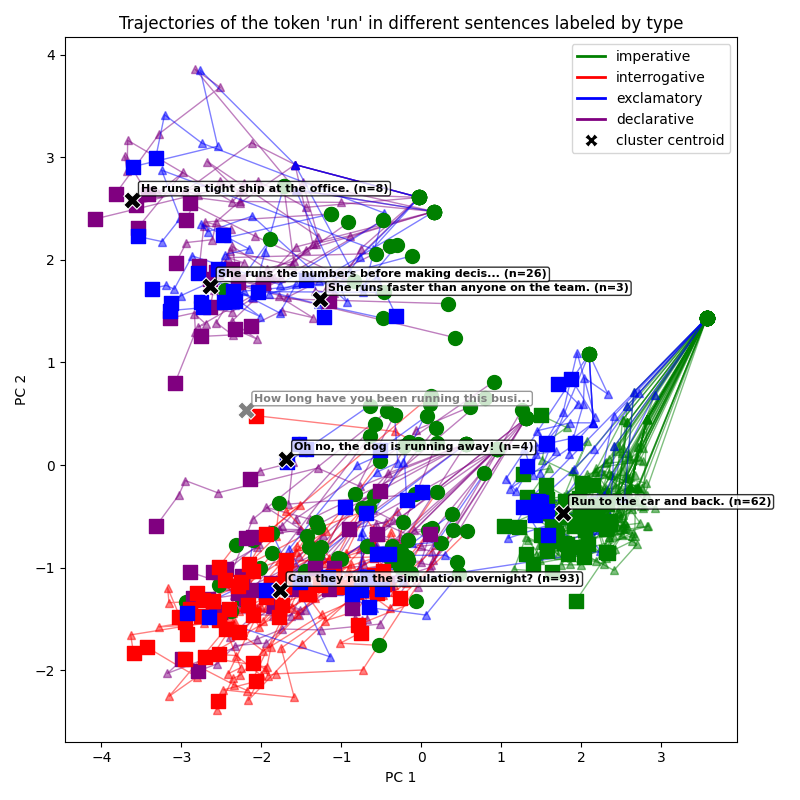}
    \includegraphics[width=0.45\linewidth]{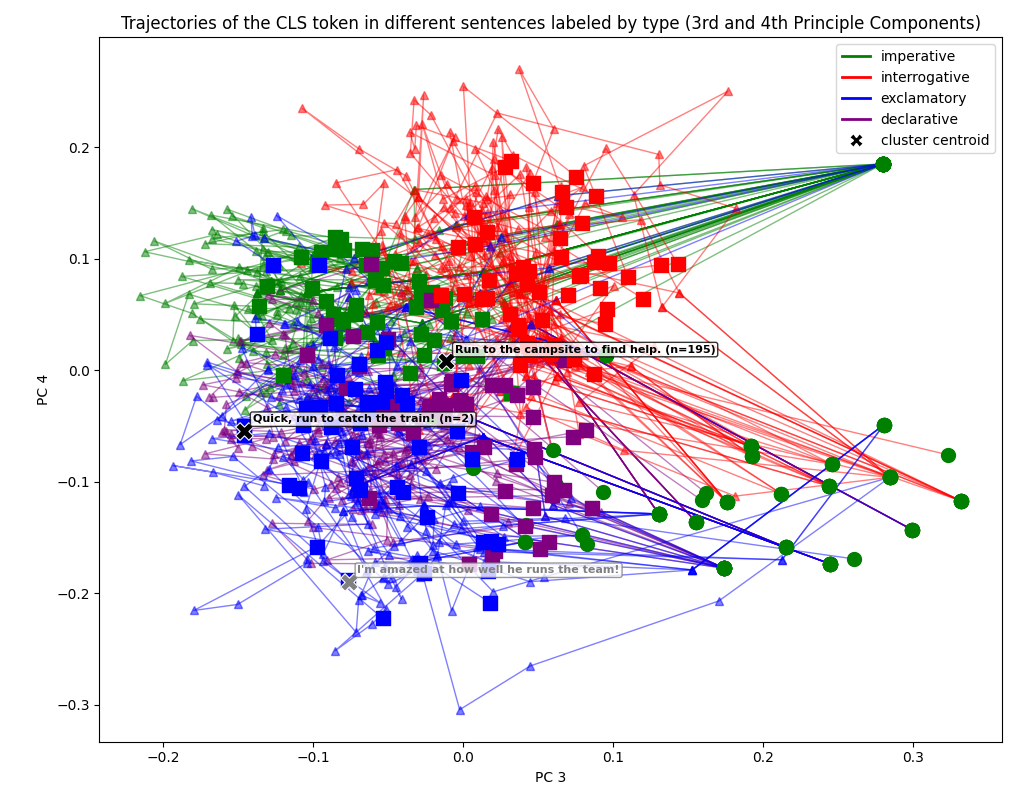}
    \caption{Principle Component Analysis of the latent space representation of the CLS token of sentences with the word run, grouped by sentence type, with labelled cluster centroids.}
    \label{fig:latent_traj}
\end{figure}
\begin{figure}
    \centering
    \includegraphics[width=0.3\linewidth]{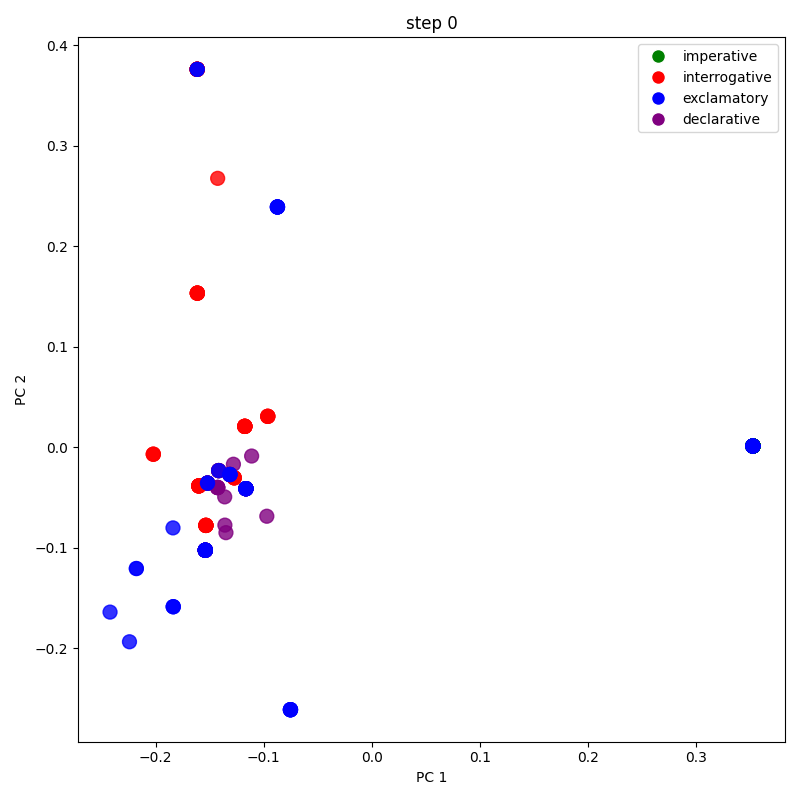}
    \includegraphics[width=0.3\linewidth]{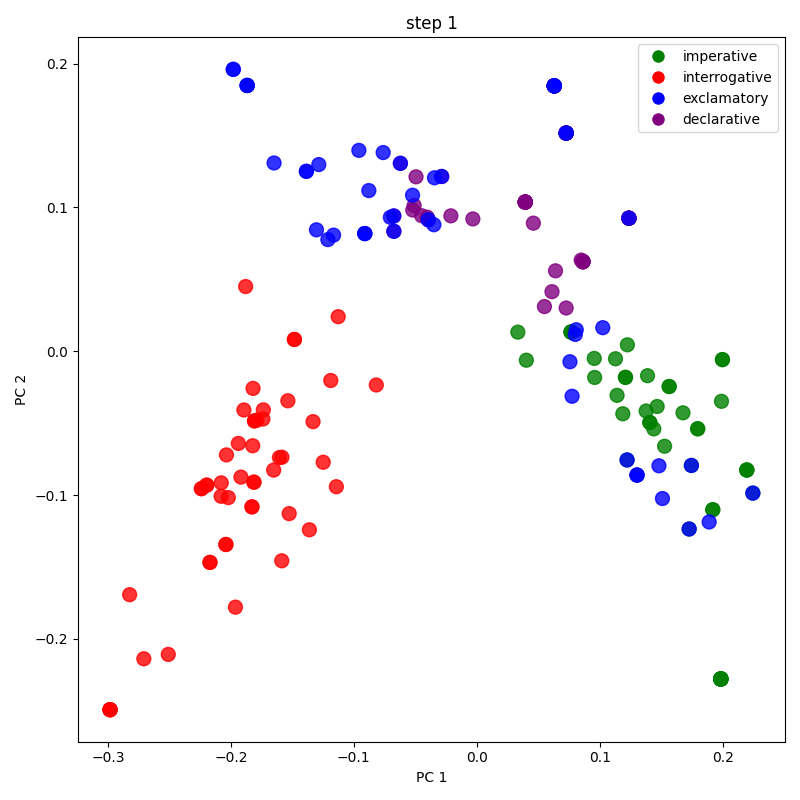}
    \includegraphics[width=0.3\linewidth]{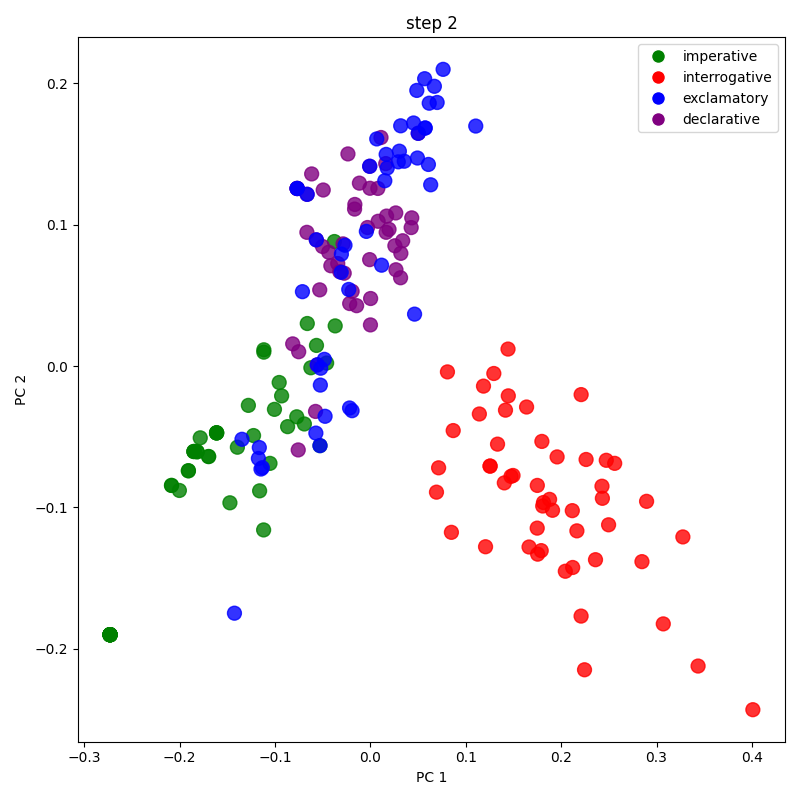}
    \includegraphics[width=0.3\linewidth]{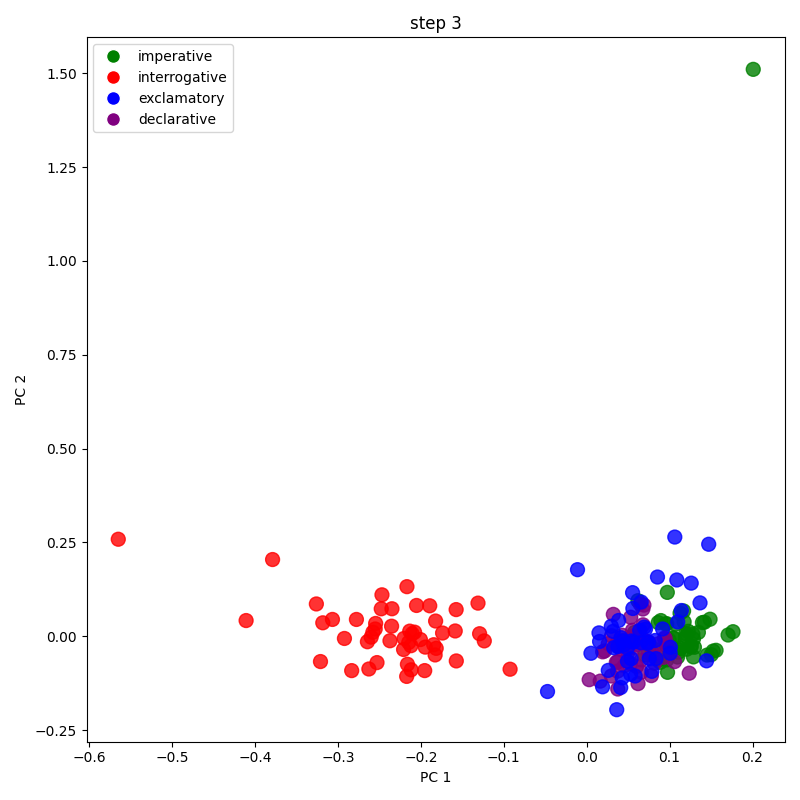}
    \includegraphics[width=0.3\linewidth]{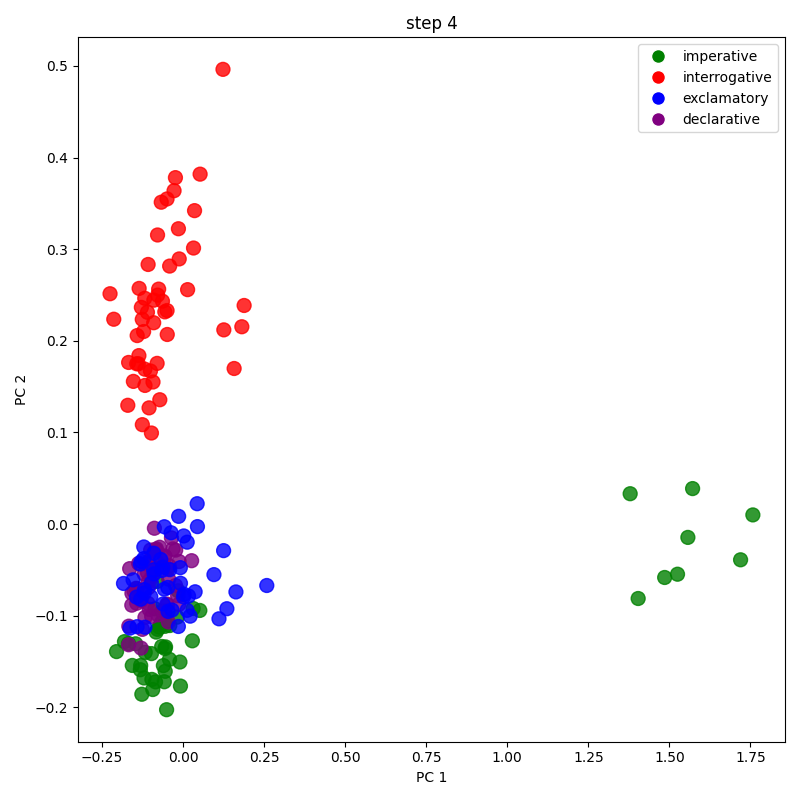}
    \includegraphics[width=0.3\linewidth]{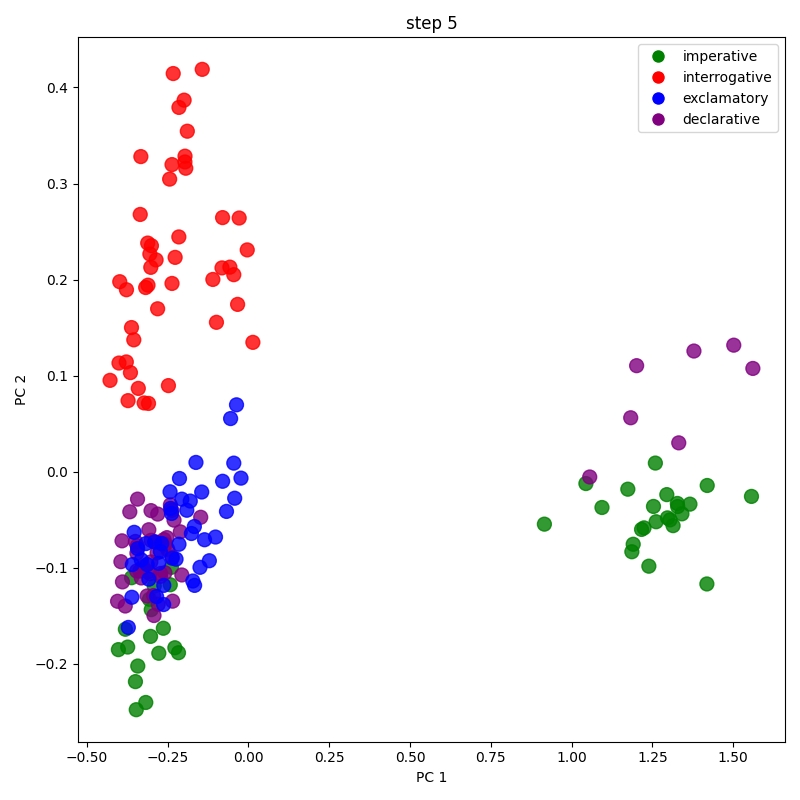}
    \includegraphics[width=0.3\linewidth]{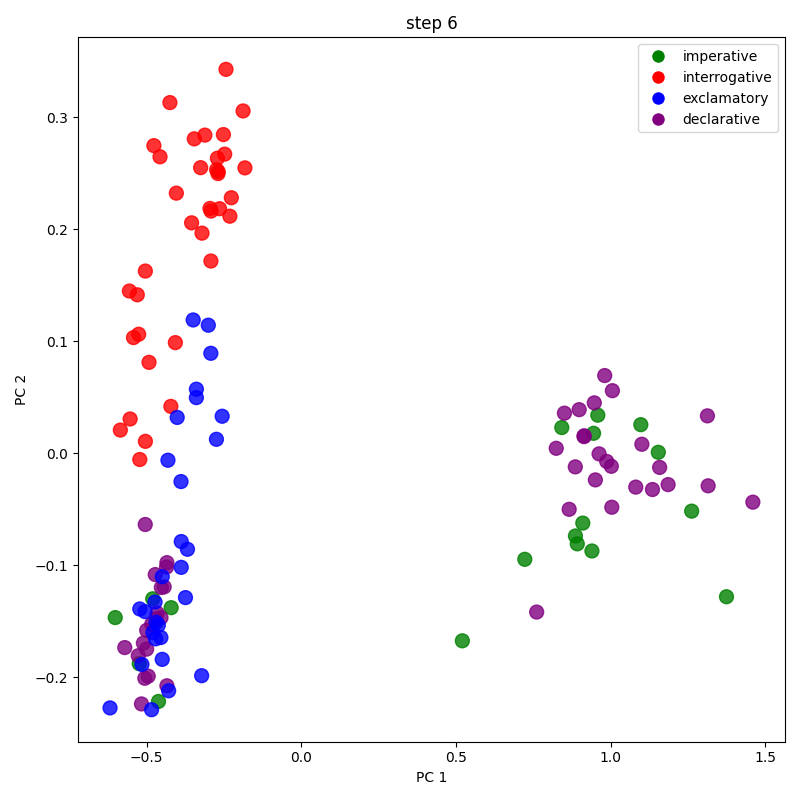}
    \includegraphics[width=0.3\linewidth]{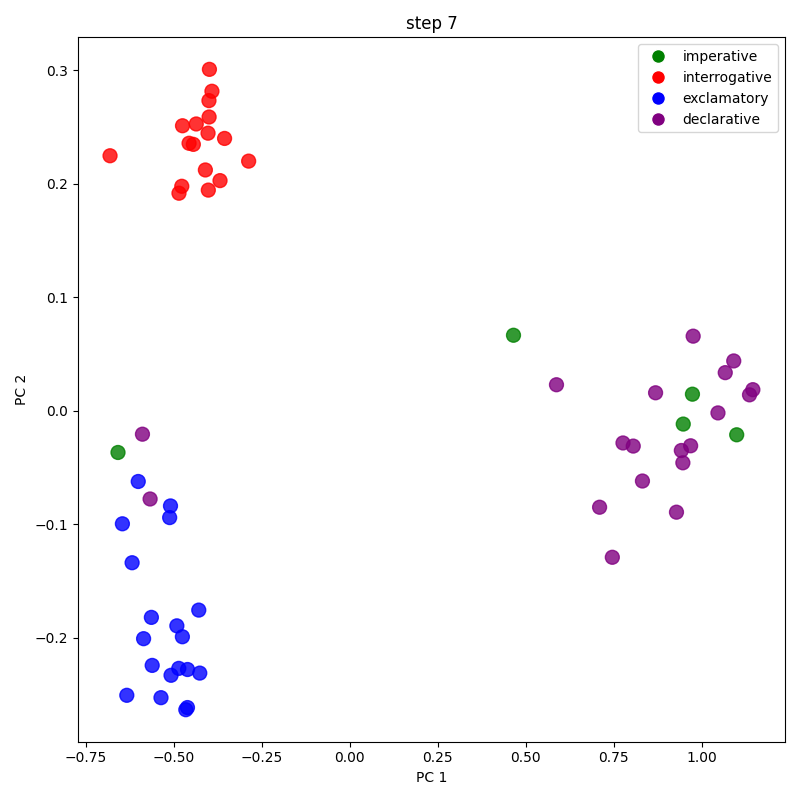}
    \caption{Sequential PCA plots of CLS token embedding after adding 1 token to context per step showing emergence of organization in latent space.}
    \label{fig:emergence_run_sentences}
\end{figure}
\begin{figure}
    \centering
    \includegraphics[width=0.5\linewidth]{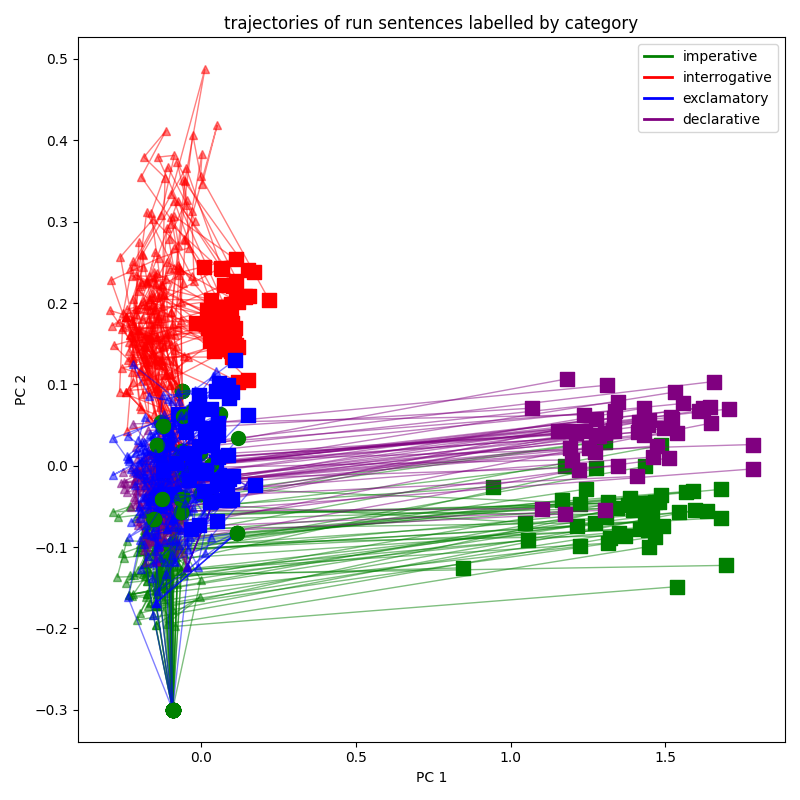}
    \caption{Principle Component Analysis of the latent space representation of the incremental trajectory of sentences containing the token ``run''.}
    \label{fig:run_sentences_with_traj}
\end{figure}
\begin{figure}
    \centering
    \includegraphics[width=0.9\linewidth]{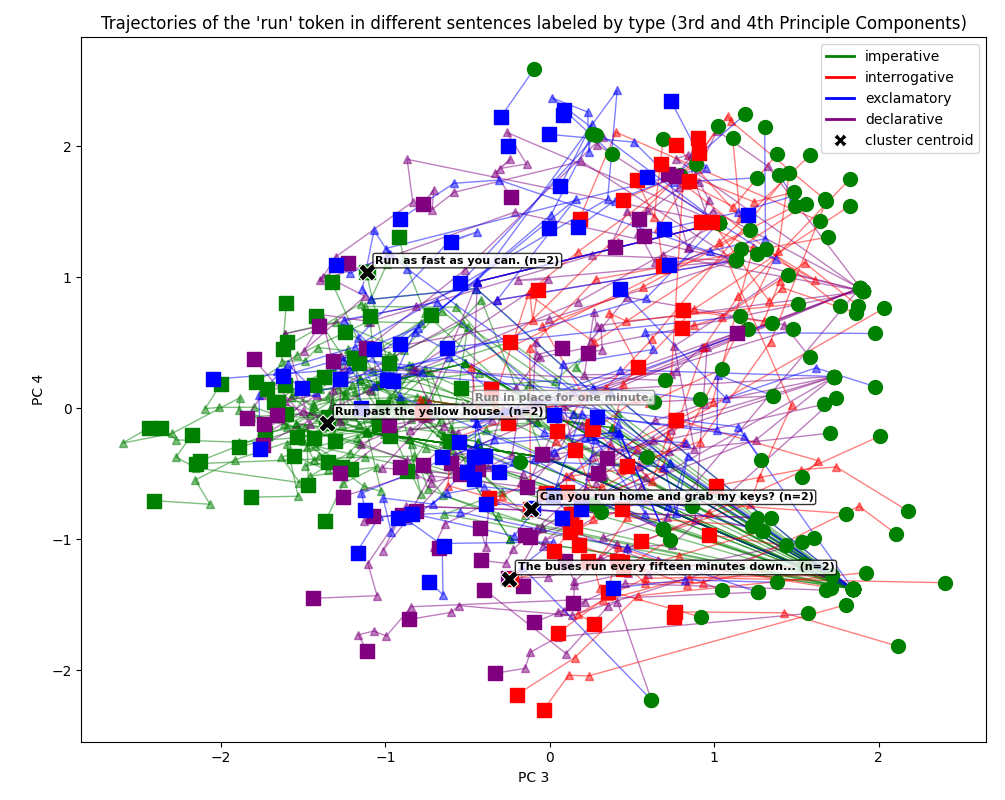}
    \caption{Principle Component Analysis of the latent space representation of the target token (``run'') of sentences, grouped by sentence type.}
\end{figure}
\begin{figure}
    \centering
    \includegraphics[width=0.5\linewidth]{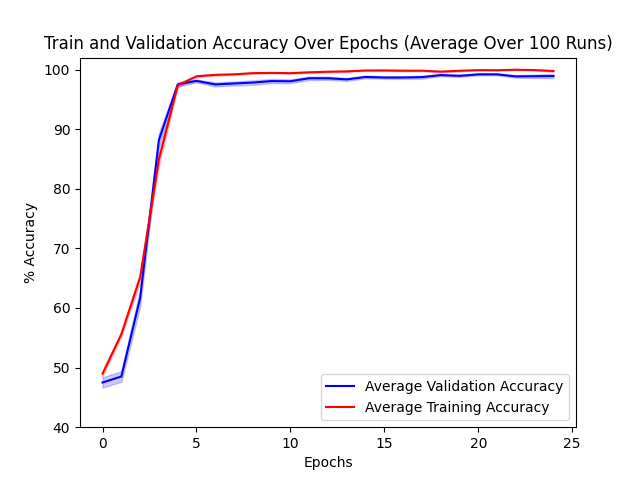}
    \caption{Validation and training accuracy curves for classifier trained on the trajectory of the first token in each sentence.}
    \label{fig:first_token}
\end{figure}
\begin{figure}
    \centering
    \includegraphics[width=0.5\linewidth]{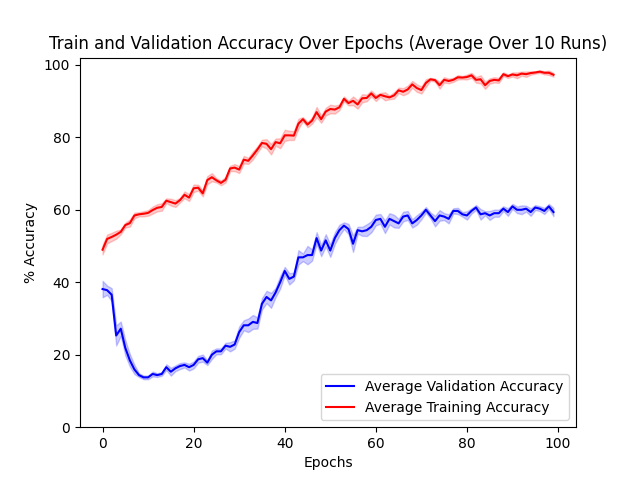}
    \caption{Validation and training accuracy curves for fully-connected feed-forward classifier trained on the difference vector between the penultimate and last token in each sentence.}
    \label{fig:last_token}
\end{figure}
\begin{figure}
    \centering
    \includegraphics[width=0.5\linewidth]{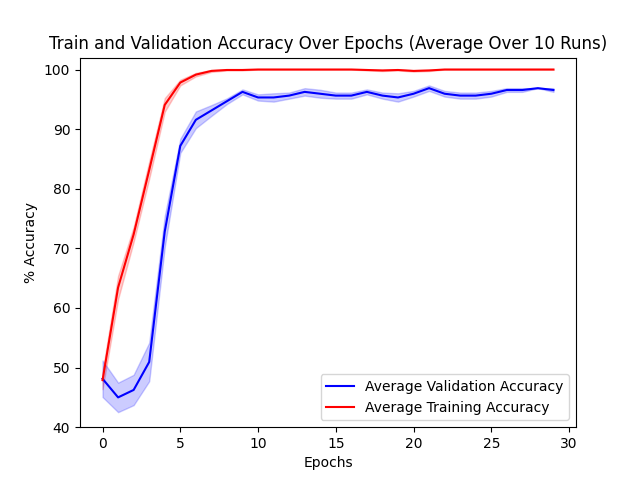}
    \caption{Validation and training accuracy curves for classifier trained on the trajectory of the first token in the critical region of each sentence.}
    \label{fig:divergent_hinge}
\end{figure}

\end{appendices}



\end{document}